\documentclass[10pt,twocolumn,letterpaper]{article}

\usepackage[pagenumbers]{cvpr} 

\usepackage{graphicx}
\usepackage{amsmath}
\usepackage{amssymb}
\usepackage{booktabs}
\usepackage{enumitem}
\usepackage{makecell}
\usepackage{multicol}
\usepackage{multirow}
\usepackage{soul}
\usepackage{float}

\usepackage{algorithm}
\usepackage{algorithmic}

\usepackage[dvipsnames]{xcolor}

\def\darkgreen#1{\textcolor[RGB]{30,150,30}{\textbf{#1}}}
\def\darkred#1{\textcolor[RGB]{180,20,20}{\textbf{#1}}}

\definecolor{cvprblue}{rgb}{0.21,0.49,0.74}
\usepackage[pagebackref,breaklinks,colorlinks,citecolor=cvprblue]{hyperref}

\usepackage[capitalize]{cleveref}
\crefname{section}{Sec.}{Secs.}
\Crefname{section}{Section}{Sections}
\Crefname{table}{Table}{Tables}
\crefname{table}{Tab.}{Tabs.}

\makeatletter 
\def\@captype{figure} 
\makeatother

\newcommand{\tabincell}[2]{\begin{tabular}{@{}#1@{}}#2\end{tabular}}

\def\paperID{8103} 
\def\confName{CVPR}
\def\confYear{2024}

\title{Adversarial Score Distillation: When score distillation meets GAN}

\author{Min Wei$^{1}$\footnotemark[1]~~~Jingkai Zhou$^{2}$\footnotemark[1]~~~Junyao Sun$^{3}$~~~Xuesong Zhang$^{1}$\\ \small {$^1$Beijing University of Posts and Telecommunications ~~~$^2$Independent Researcher~~~ $^3$South China University of Technology}\\ \small \{mw, xuesong\_zhang\}@bupt.edu.cn ~~~ fs.jingkaizhou@gmail.com ~~~ 201810108390@mail.scut.edu.cn}

\begin{document}
\maketitle

\renewcommand{\thefootnote}{\fnsymbol{footnote}}
\footnotetext[1]{Equal contribution} 
\footnotetext{Corresponding authors: {Jingkai Zhou}, {Xuesong Zhang}}

\begin{abstract}
Existing score distillation methods are sensitive to classifier-free guidance (CFG) scale, manifested as over-smoothness or instability at small CFG scales, while over-saturation at large ones. To explain and analyze these issues, we revisit the derivation of Score Distillation Sampling (SDS) and decipher existing score distillation with the Wasserstein Generative Adversarial Network (WGAN) paradigm. With the WGAN paradigm, we find that existing score distillation either employs a fixed sub-optimal discriminator or conducts incomplete discriminator optimization, resulting in the scale-sensitive issue. We propose the Adversarial Score Distillation (ASD), which maintains an optimizable discriminator and updates it using the complete optimization objective. Experiments show that the proposed ASD performs favorably in 2D distillation and text-to-3D tasks against existing methods. Furthermore, to explore the generalization ability of our paradigm, we extend ASD to the image editing task, which achieves competitive results. The project page and code are at \href{https://2y7c3.github.io/ASD/asd.html}{this link}.
\end{abstract}    
\section{Introduction}
\label{sec:intro}
Score distillation is a rapidly growing technique attracting a lot of attention~\cite{sds, sjc, vsd, dds, diff_instruct, csd, collaborative_score_distillation}. It transfers knowledge from a pretrained 2D diffusion model to downstream tasks, such as image editing~\cite{dds, collaborative_score_distillation}, video editing~\cite{collaborative_score_distillation}, diffusion model distillation~\cite{diff_instruct, add}, and text-to-3D~\cite{sds, magic3d, sjc, vsd, latent_nerf, perp_neg, fantasia3d, csd, mvdreamer, points23d,it3d}, with no downstream data required. 

\begin{figure}[t]
    \centering 
    \small
    \addtolength{\tabcolsep}{-5.5pt}
    \begin{tabular}{cccc}
    \includegraphics[width=.245\linewidth]{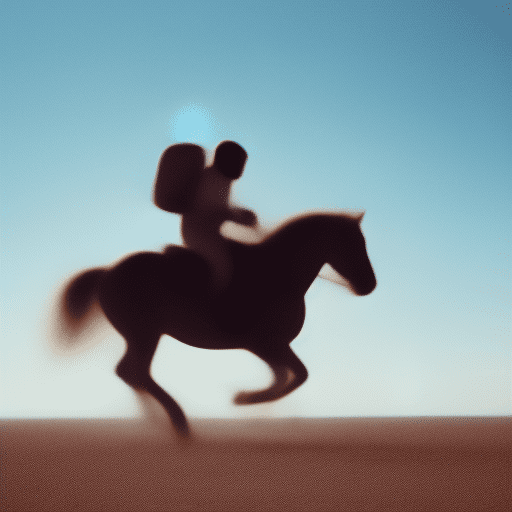}&
    \includegraphics[width=.245\linewidth]{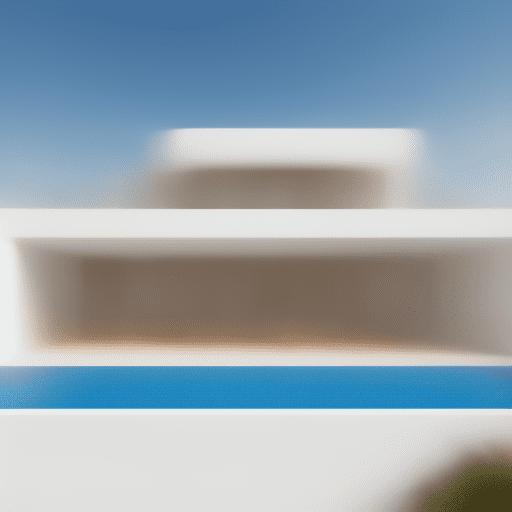}&
    \includegraphics[width=.245\linewidth]{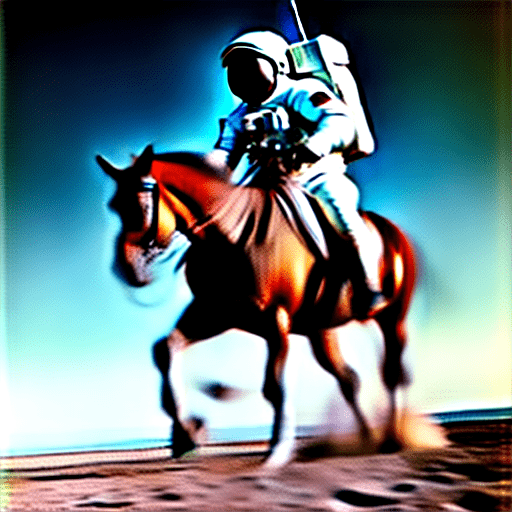}&
    \includegraphics[width=.245\linewidth]{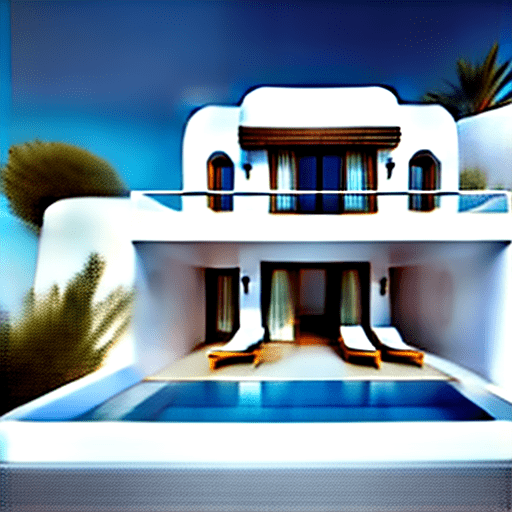}\\
    \multicolumn{2}{c}{SDS (CFG = 7.5)}&
    \multicolumn{2}{c}{SDS (CFG = 100)}\\
    \multicolumn{4}{c}{\vspace{2mm}\normalsize{(a) SDS is sensitive to the CFG scale.}}\\
    \multicolumn{4}{c}{Optimization Steps\vspace{-2mm}}\\
    \multicolumn{4}{c}{\hspace{2mm}\vspace{-2mm}\includegraphics[width=.8\linewidth]{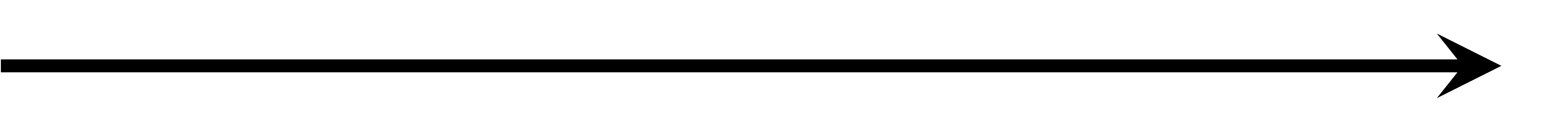}}\\
    \includegraphics[width=.245\linewidth]{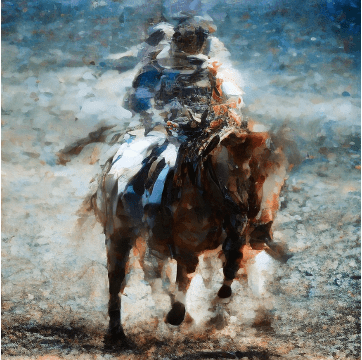}&
    \includegraphics[width=.245\linewidth]{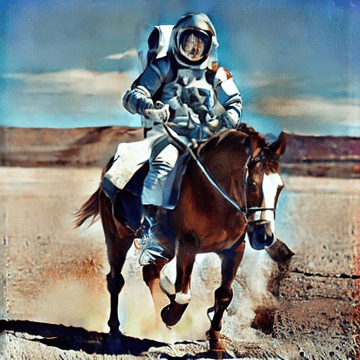}&
    \includegraphics[width=.245\linewidth]{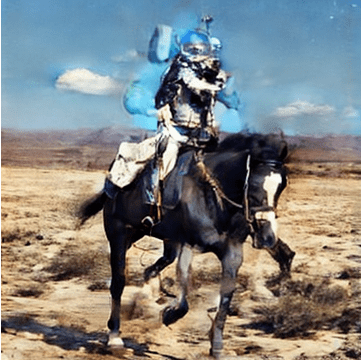}&
    \includegraphics[width=.245\linewidth]{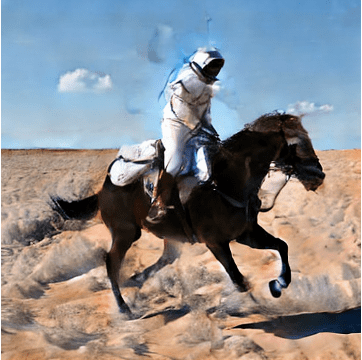}\\
    \includegraphics[width=.245\linewidth]{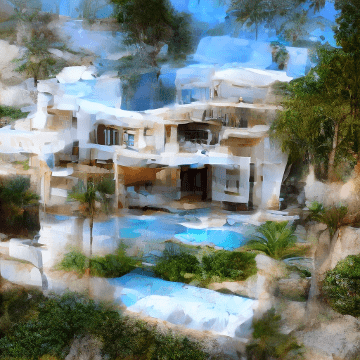}&
    \includegraphics[width=.245\linewidth]{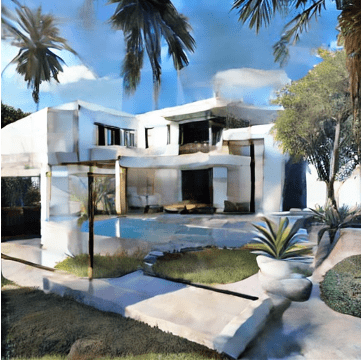}&
    \includegraphics[width=.245\linewidth]{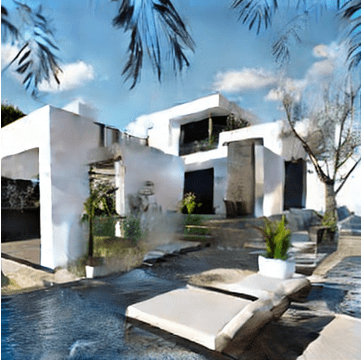}&
    \includegraphics[width=.245\linewidth]{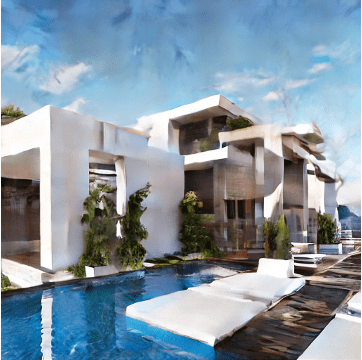}\\
    \multicolumn{4}{c}{\normalsize{(b) Unstable distillation progress of VSD.}}
    \end{tabular}
\caption{2D score distillation examples with the prompts ``a photograph of an astronaut riding a horse'' and ``exterior frontal perspective shot of resort villa inspired by Mykonos architecture''. SDS is very sensitive to the CFG scales while VSD exhibits fluctuation of generated contents during distillation at small CFG scales.}
\label{fig:sds_vsd_issue}
\end{figure}

However, as a milestone in score distillation methods, Score Distillation Sampling (SDS)~\cite{sds} is very sensitive to the classifier-free guidance (CFG)~\cite{cfg} scale. Figure~\ref{fig:sds_vsd_issue} takes 2D score distillation as an example. When setting a small CFG scale, SDS tends to obtain over-smoothing results, whereas when setting a large one, the generated images become over-saturated. Recently, Variational Score Distillation (VSD)~\cite{vsd} alleviates the over-smoothing problem at small CFG scales, but we observe that VSD is a bit unstable during the distillation progress at small CFG scales, as shown in Figure~\ref{fig:sds_vsd_issue}. These phenomena make us conjecture that there are still some unrevealed methodological issues behind SDS and VSD.

We start from revisiting the derivation of SDS. 
In~\cite{sds}, the gradient of SDS is derived from the L2 loss of the diffusion model. This derivation holds only when the vanilla-predicted noise is used in the gradient. However, in practice, the SDS gradient exploits the CFG noise rather than the vanilla-predicted one. This little trick changes the whole thing. It means that, in practice, the real SDS loss is no longer the L2 loss of the diffusion model, but some other loss associated with the implicit classifier or discriminator.

\begin{figure*}[t]
    \centering 
    \small
    \addtolength{\tabcolsep}{-6pt}
    \begin{tabular}{cccc}
    \includegraphics[width=.249\linewidth]{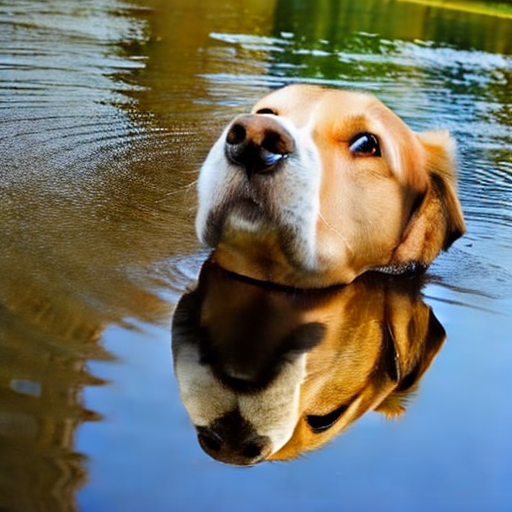}&
    \includegraphics[width=.249\linewidth]{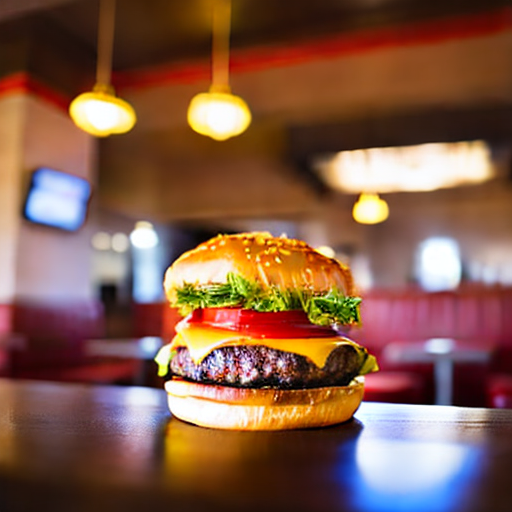}&
    \includegraphics[width=.249\linewidth]{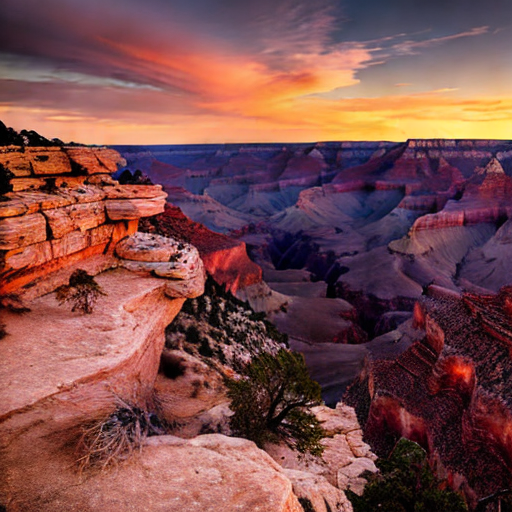}&
    \includegraphics[width=.249\linewidth]{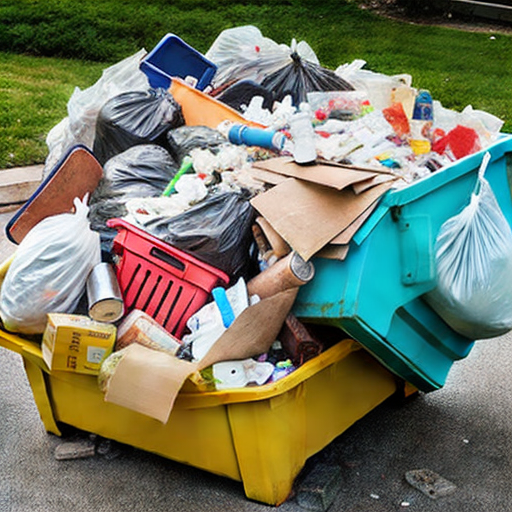}\\
    \tabincell{c}{\textit{a dog with its reflection below}} &
    \tabincell{c}{\textit{a DSLR photo of a hamburger}\\\textit{inside a restaurant}} &
    \tabincell{c}{\textit{a professional photo of a sunset}\\\textit{behind the Grand Canyon}} &
    \tabincell{c}{\textit{a dumpster full of trash}}\vspace{1mm}\\
    \multicolumn{4}{c}{\vspace{2mm}\large{(a) ASD can generate photorealistic images through 2D score distillation.}}\\
    \includegraphics[width=.249\linewidth,trim=3 3 3 3, clip]{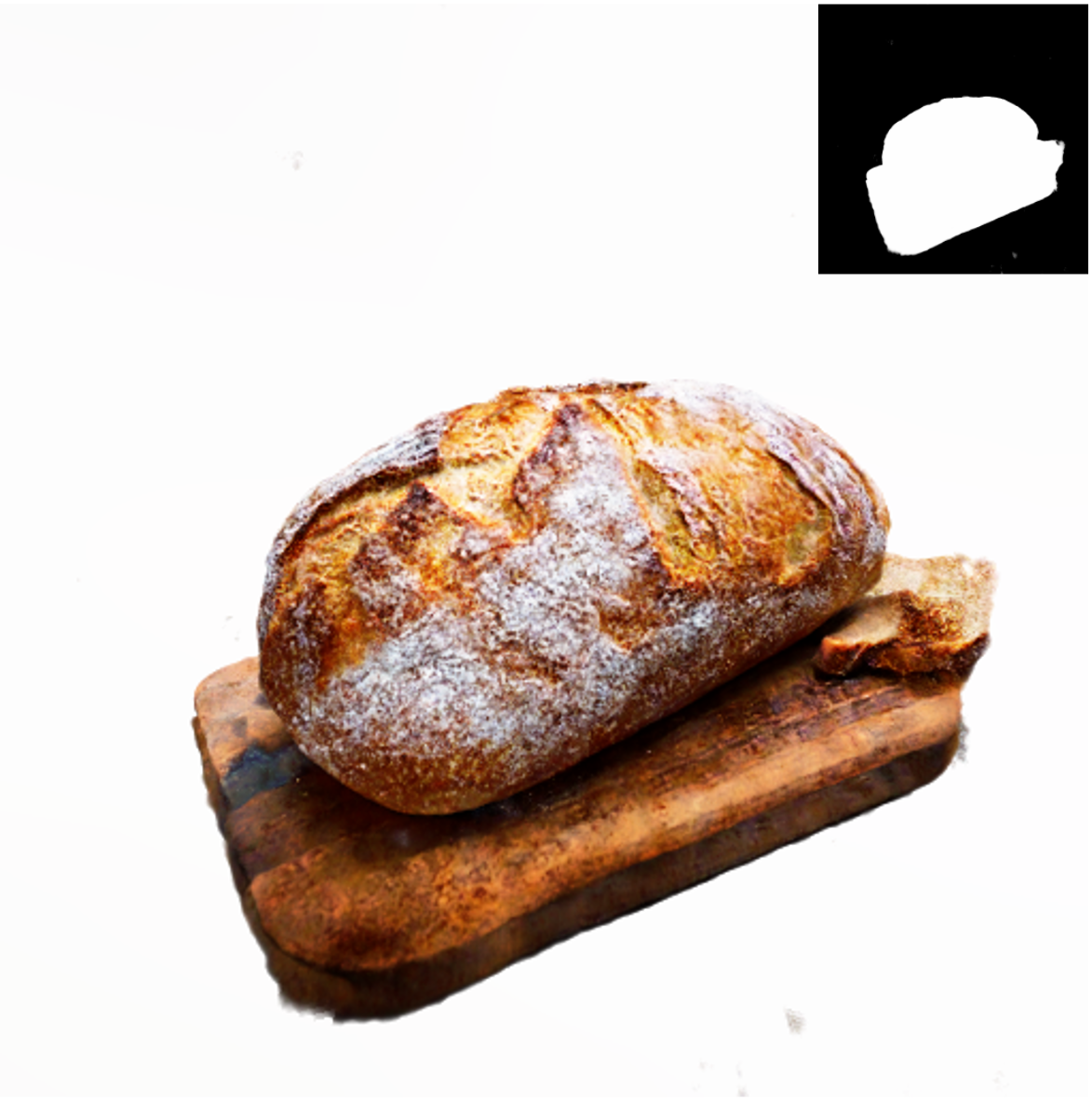}&
    \includegraphics[width=.249\linewidth,trim=3 3 3 3, clip]{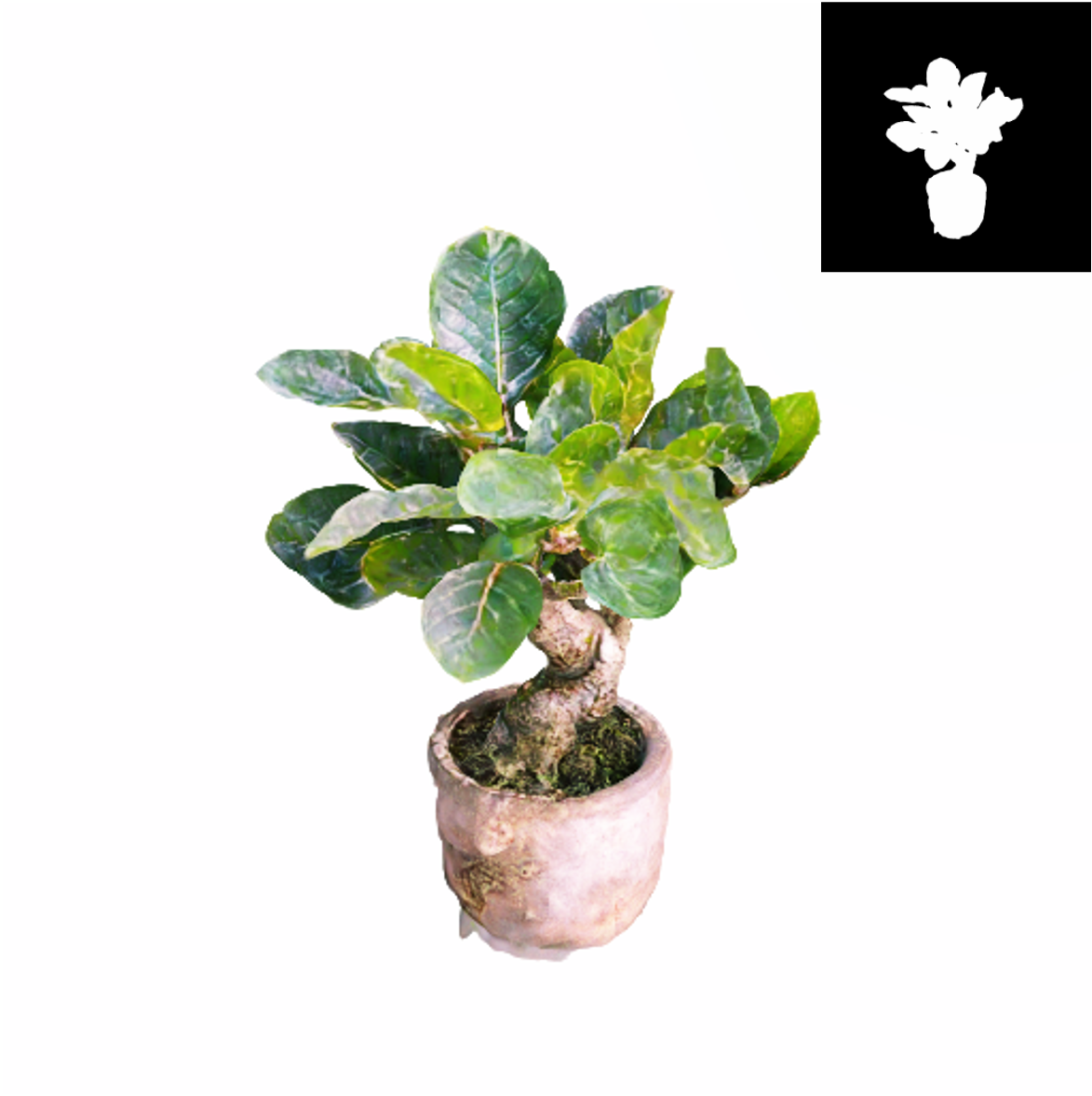}&
    \includegraphics[width=.249\linewidth,trim=3 3 3 3, clip]{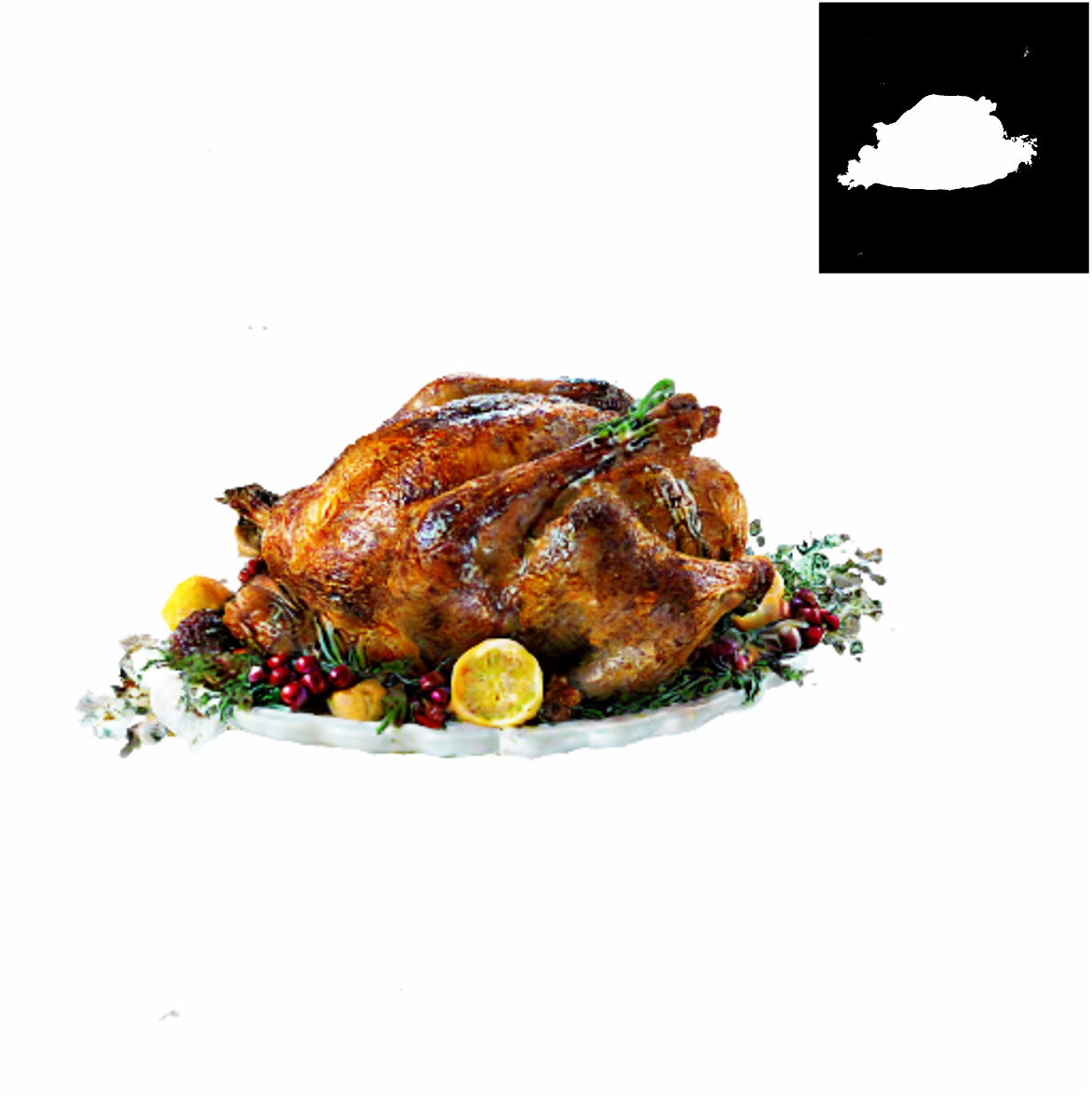}&
    \includegraphics[width=.249\linewidth,trim=3 3 3 3, clip]{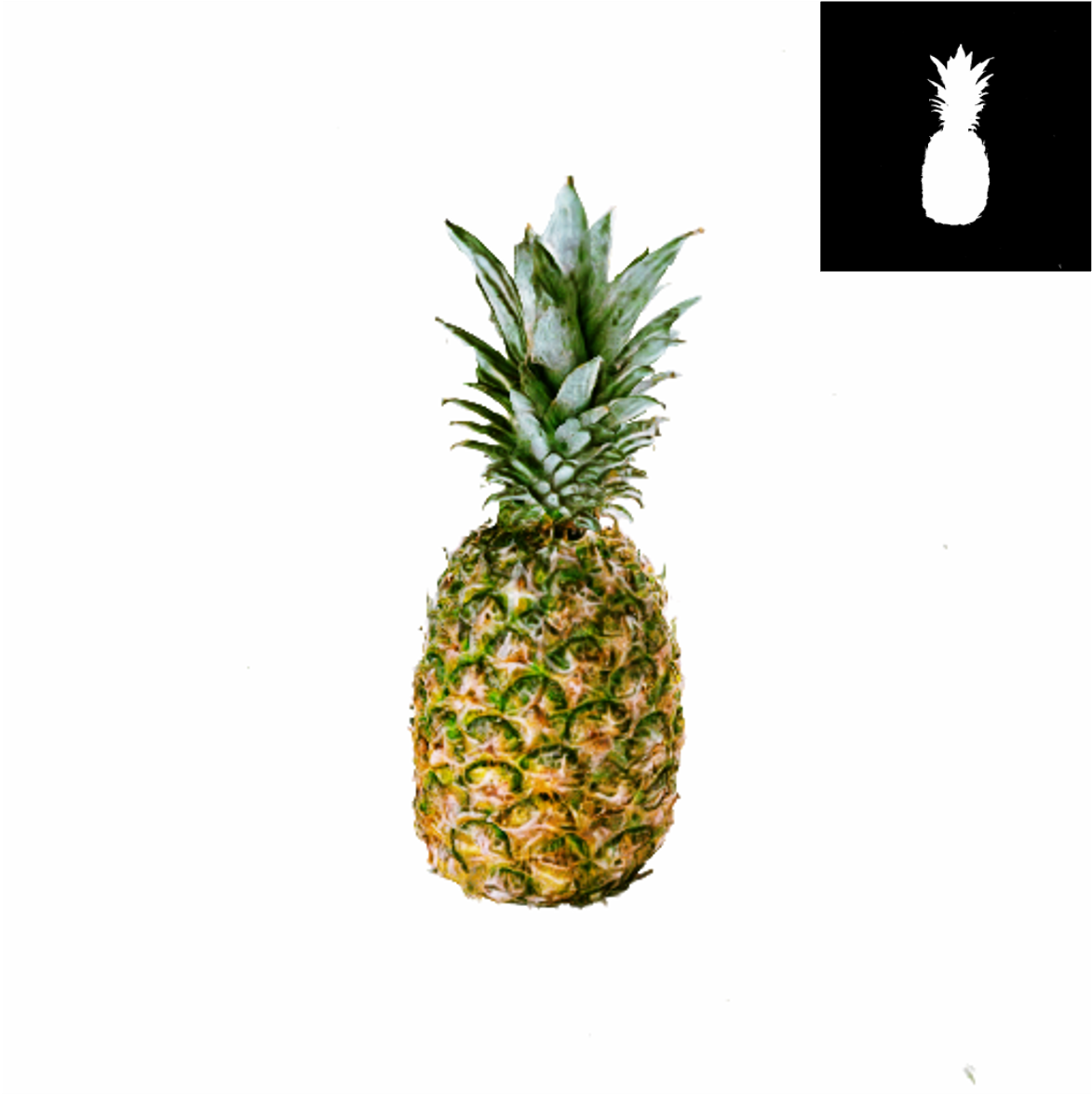}\\
    \tabincell{c}{\textit{a freshly baked loaf of sourdough}\\ \textit{bread on a cutting board}} &
    \tabincell{c}{\textit{a ficus planted in a pot}} &
    \tabincell{c}{\textit{a roast turkey on a platter}} &
    \tabincell{c}{\textit{a pineapple}}\vspace{1mm}\\
    \multicolumn{4}{c}{\vspace{2mm}\large{(b) ASD can generate high quality 3D NeRFs from scratch (only stage 1 of VSD~\cite{vsd}).}}\\
    \includegraphics[width=.249\linewidth]{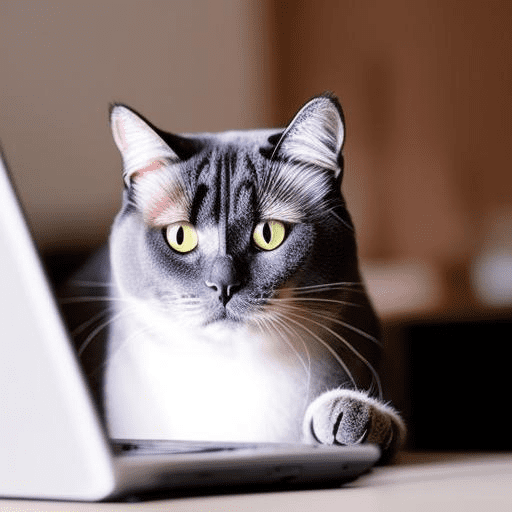}&
    \includegraphics[width=.249\linewidth]{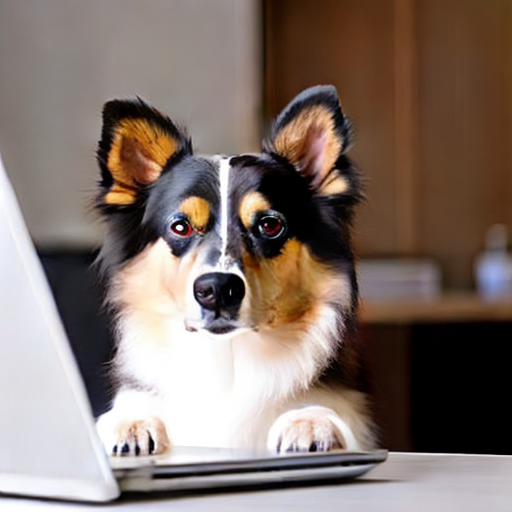}&
    \includegraphics[width=.249\linewidth]{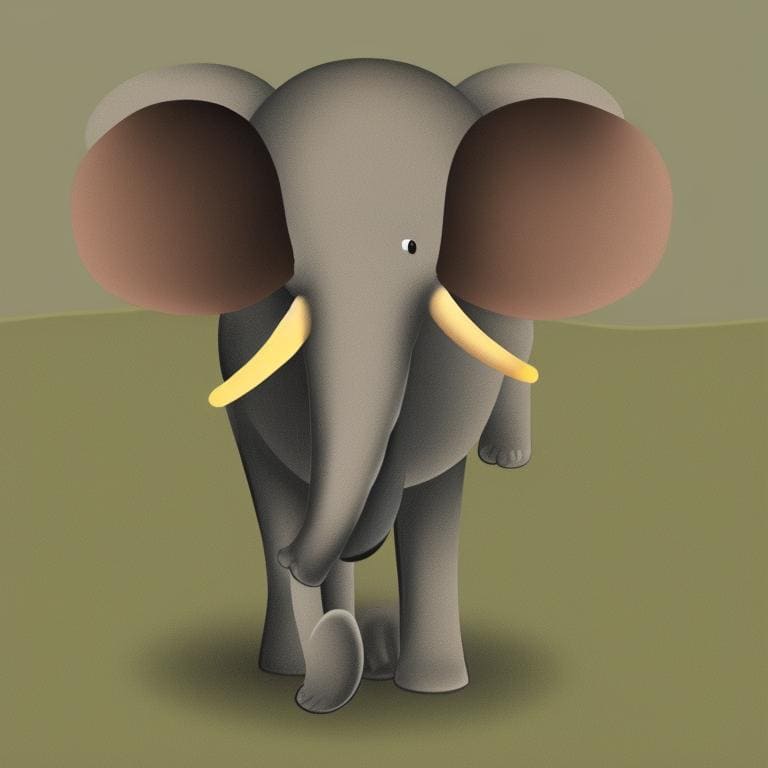}&
    \includegraphics[width=.249\linewidth]{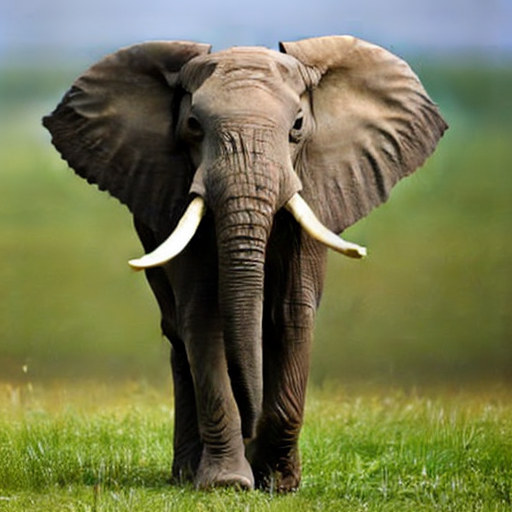}\\
    \tabincell{c}{\textit{a \darkred{cat} is posing next to}\\ \textit{a laptop computer}} &
    \tabincell{c}{\textit{a \darkgreen{dog} is posing next to}\\ \textit{a laptop computer}} &
    \tabincell{c}{\textit{A \darkred{cartoon} elephant}} &
    \tabincell{c}{\textit{A \darkgreen{\st{cartoon}} elephant}}\vspace{1mm}\\
    \multicolumn{4}{c}{\large{(c) ASD can extend to image editing with caption modification.}}
    \end{tabular}
\caption{Examples generated by ASD in 2D distillation, image editing, and text-to-3D tasks. Stable Diffusion~\cite{sd_2_1_base, sd_2_1} is used as the pretrained diffusion model. For more results please refer to our project page.}
\label{fig:the_first_fig}
\vspace{-4mm}
\end{figure*}

To figure out where the real SDS gradient comes from, we connect SDS to Generative Adversarial Networks (GANs)~\cite{gan, wgan, wgan_div}. By carefully designing the discriminator using the diffusion model, we can naturally derive the SDS gradient from the generator loss in the Wasserstein GAN (WGAN)~\cite{wgan, wgan_div}. This means that SDS inherently comes from WGAN. However, when we employ the WGAN paradigm to explain and analyze SDS, we find that the discriminator optimization in WGAN is ignored by SDS. More specifically, SDS only optimizes the generator loss in WGAN by using a fixed sub-optimal discriminator implemented with added noise. Similarly, VSD can also be represented with the WGAN paradigm, in which case the discriminator is optimizable but the optimization objective is incomplete compared to the WGAN discriminator loss, making the distillation progress unstable.

In this work, we propose Adversarial Score Distillation (ASD) based on the WGAN paradigm. ASD maintains an optimizable discriminator and optimizes it using the complete WGAN discriminator loss, thereby improving distillation stability and quality. Specifically, the discriminator can be implemented using the combination of diffusion models, where textual-inversion embedding~\cite{text_inervsion} or LoRA~\cite{lora} make it optimizable. During discriminator optimization, we propose two kinds of optimization objectives derived from the complete WGAN discriminator loss. One exploits both real sample distribution and a pretrained diffusion model and can therefore be applied to the image-conditioned task, such as image editing. The other only requires a pretrained diffusion model, which is suitable for text-conditioned cases. Experiments on 2D distillation and text-to-3D tasks demonstrate the superior performance of ASD compared to existing score distillation methods. See Figure~\ref{fig:the_first_fig} for some examples. In addition, we extend ASD to image editing via discriminator transformations and show that the Delta Denoising Score (DDS)~\cite{dds} is a special case in our paradigm. Figure~\ref{fig:the_first_fig} and experiments show some competitive editing results. This demonstrates that our paradigm can be generalized to more GAN paradigms for various tasks.

Beyond methodological improvements, the more important significance of bridging score distillation and GANs is to enable GAN paradigms to exploit powerful diffusion models in the form of score distillation. More specifically, we can extend a general diffusion model to various downstream tasks (\eg, text-to-3D, image editing, diffusion distillation) by designing GAN paradigms, which eliminates the need to redesign specific diffusion models or collect large amounts of finetuning samples. 

In short, we make the following contributions:
\begin{itemize}[itemsep=0mm, leftmargin=*]
    \item Bridged score distillation and WGAN to explain the methodological issues of existing score distillation, which also enables the pretrained model to extend to various downstream tasks by designing GAN paradigms.
    \item Proposed ASD based on the WGAN paradigm that employs the complete WGAN discriminator loss, resulting in better distillation stability and quality.
    \item Comprehensive experiments show that the proposed ASD performs favorably against existing methods in 2D distillation, text-to-3D, and image editing tasks.
\end{itemize}

\section{Related Work}
\noindent \textbf{Score Distillation.} Score distillation, first proposed by \cite{sds} and \cite{sjc}, is a method to optimize the parameter space via distilling a pretrained diffusion model. In text-to-3D generation, Score Distillation Sampling (SDS) shows great potential for optimizing 3D representations based on differentiable rendering. However, SDS often suffers from problems such as over-saturation, over-smoothing, and low diversity. The recent work Variational Score Distillation (VSD)~\cite{vsd} has made tremendous progress in solving these problems, which optimizes a 3D distribution by employing the Wasserstein gradient flow. However, VSD is a bit unstable during the optimization progress at small CFG scales, which will be discussed later. Score distillation can also be used in image editing~\cite{dds, styleganfusion}.  Delta Denoising Score (DDS)~\cite{dds} proposes to modify the image in a zero-shot way by only editing its caption. Moreover, StyleGAN-Fusion~\cite{styleganfusion} and Diff-Instruct~\cite{diff_instruct} apply score distillation to knowledge distillation between generators in a data-free manner. Our work will trace the origin of SDS and connect it with GAN to reveal its methodological issues.

\noindent \textbf{Generative Adversarial Networks.} GANs~\cite{gan, wgan, wgan_div, principle, improved_wgan, Mao_2017_ICCV,gansurvey} have achieved great success in generating realistic and clear images, such as image in-painting~\cite{dolhansky2018eye, iizuka2017globally}, image manipulation applications~\cite{zhu2016generative, nam2018text, brock2016neural, he2019attgan, bau2018gan}, image-to-image translation~\cite{isola2017image, wang2018perceptual, zhu2017unpaired, kim2017learning, yi2017dualgan, choi2018stargan, liu2017unsupervised, huang2018multimodal, lee2018diverse}. 
Among these 2D GANs, Wasserstein GAN (WGAN)~\cite{wgan} is a remarkable study that minimizes the Wasserstein distance instead of the Jensen-Shannon divergence to better handle the case when the generated sample distribution is far away from the real distribution.  
In 3D modeling, a lot of work on 3D GANs~\cite{3dgan,pmlr-v78-smith17a, pmlr-v80-achlioptas18a, Yang_2017_ICCV} has been proposed. However, the generation of 3D representations from low-dimensional latent spaces needs expensive high-quality 3D data for training. GAN2Shape~\cite{pan2021do} demonstrates that conventional 2D GANs, when exclusively trained on images, contain rich 3D knowledge and can faithfully reconstruct intricate 3D shapes without 3D annotations. In this work, we only use the form of WGAN to explain and analyze score distillation methods, such as SDS and VSD.

\section{Preliminaries}

\noindent \textbf{Score Distillation Sampling.} Score Distillation Sampling (SDS)~\cite{sds} proposes to use 2D diffusion priors to optimize the parameters $\theta$ of 3D representations. It first employs a differentiable renderer $g(\theta,c)$ parameterized by $\theta$ to generate an image $x_0^g = g(\theta, c)$ at a random view $c$, then adds noise to this image $x_0^g$ to obtain $x_t^g$, and uses a pretrained diffusion model to predict noise conditioned on a given text $y$. In the SDS paper~\cite{sds}, the gradient is directly derived from the L2 loss of the diffusion model, which can be written as
\begin{align}
    &\nabla_\theta \mathcal{L}_{\text{diff}} = \nabla_\theta \mathbb{E}_{t, \epsilon}[\omega(t)\lVert {\epsilon}_{x_t^g; y, t} - \epsilon \rVert_2^2] \nonumber\\
    \label{eq:sds}
    = &\mathbb{E}_{t, \epsilon}[\omega(t)({\epsilon}_{x_t^g; y, t} - \epsilon)\frac{\partial {\epsilon}_{x_t^g; y, t}}{\partial x_t^g}\frac{\partial x_t^g}{\partial x_0^g}\frac{\partial x_0^g}{\partial \theta}]\\
    \approx & \mathbb{E}_{t, \epsilon}[\omega(t)(\epsilon_{x_t^g; y, t} - \epsilon)\frac{\partial x_0^g}{\partial \theta}] = \nabla_\theta \mathcal{L}_{\text{SDS}}\nonumber
\end{align}
where $\omega(t)$ is a weight based on the time $t$, and $\epsilon_{x_t^g; y, t}$ is the vanilla-predicted noise with input $x_t^g$ conditioned on $y$ and $t$. However, in practice, SDS uses the classifier-free guidance (CFG)~\cite{cfg} noise, that is
\begin{equation}
\tilde{\epsilon}_{x_t^g; y, t} = \epsilon_{x_t^g; t} + \lambda(\epsilon_{x_t^g; y, t} -\epsilon_{x_t^g; t})
\label{eq:cfg}
\end{equation}
where $\epsilon_{x_t^g; t}$ is the vanilla-predicted noise with input $x_t^g$ conditioned on $t$, and $\lambda$ is the CFG scale. Replacing $\epsilon_{x_t^g; y, t}$ with $\tilde{\epsilon}_{x_t^g; y, t}$ makes the derivation of Eq.~\ref{eq:sds} no longer valid. In other words, $\nabla_\theta \mathcal{L}_{\text{SDS}}$ should be written as
\begin{equation}
    \small
    \nabla_\theta \mathcal{L}_{\text{SDS}} = \mathbb{E}_{t, \epsilon}[\omega(t)(\epsilon_{x_t^g; t} + \underbrace{\lambda(\epsilon_{x_t^g; y, t} -\epsilon_{x_t^g; t})}_{\text{grad from implicit clssifier}} - \epsilon)\frac{\partial x_0}{\partial \theta}]
    \label{eq:grad_sds}
\end{equation}
which cannot be derived from $\mathcal{L}_{\text{diff}}$, but rather from another loss related to the implicit classifier or discriminator.

\vspace{1mm}
\noindent \textbf{Generative Adversarial Networks.}
Generative Adversarial Networks (GANs), first introduced in ~\cite{gan}, can be formulated by a zero-sum competition between a discriminator network $D$ and a generator network $G$. The original form of GAN~\cite{gan} essentially minimizes the Jensen-Shannon (JS) divergence between the distributions of real samples and generated samples, which gets a gradient close to zero when two distributions are very different. The Wasserstein GAN (WGAN)~\cite{wgan, wgan_div} reformulates the original GAN with the Wasserstein distance to handle the zero-gradient issue, which can be written as 
\begin{equation}
     \small
     \min_G\max_D\mathbb{E}_{\mu_{r}}[D(x^r)] - \mathbb{E}_{\mu_g}[D(x^g)] - \underbrace{\tau \mathbb{E}_{\tilde{\mu}}[\lVert \nabla_{\tilde{x}} D(\tilde{x}) \rVert^\text{P}}_{\text{penalty term}}]\label{eq:wgan}
\end{equation}
where $x^r$ comes from the distribution $\mu_{r}$ of real samples, $x^g$ comes from the distribution $\mu_g$ of generated samples, $\tilde{x}$ comes from a mixture distribution $\tilde{\mu}$ of first two, $\lVert \cdot \rVert^\text{P}$ denotes p-norm, and $\tau \geq 0$ is the penalty weight. 
\section{Adversarial Score Distillation}
\label{sec:ASD}
One goal of this work is to unravel the methodological issues behind SDS by revisiting and discovering the origin of the SDS gradient. As the gradient used in practice should originate from the loss associated with the implicit classifier or discriminator, it is natural to connect SDS with GAN. After carefully building a discriminator using the diffusion model, we can derive the SDS gradient directly from the generator loss in the WGAN.

We define a discriminator in the form as
\begin{align}
    D(x_t; y) &= \log \frac{p(y|x_t)}{p(\phi|x_t)}
    \label{eq:discriminator}
\end{align}
where $p(y|x_t)$ denotes the probability of the given noisy sample $x_t = \alpha_t x_0 + \sigma_t\epsilon$ to be categorized as the samples described by prompt $y$, $p(\phi|x_t)$ represents the probability of $x_t$ to be categorized as the samples described by prompt $\phi$. Here, we consider $\phi$ as the prompt for fake samples. When $x_t$ is close to samples described by prompt $y$, this $D(x_t; y)$ will give high confidence. Otherwise, it will get low confidence when $x_t$ is close to samples described by prompt $\phi$. Based on WGAN~\cite{wgan, wgan_div}, we have the generator loss as
\begin{align}
    \mathcal{L}_G &= \mathbb{E}_{t,\epsilon}[-D(x_t^g; y)] = \mathbb{E}_{t,\epsilon}[-\text{log}\frac{p(y|x_t^g)}{p(\phi|x_t^g)}]\label{eq:g_loss}\\
    &\leq \mathbb{E}_{t,\epsilon}[-\text{log}\frac{p(y|x_t^g)^\lambda}{p(\phi|x_t^g)}] := \mathcal{L}_G'~~~~~~~~s.t.~~\lambda \ge 1 \nonumber
\end{align}
where $x_t^g = \alpha_t g(\theta, c) + \sigma_t\epsilon$ denotes the generated sample with noise.
According to the Bayesian rule, we can do the transformation $\nabla_{x_t} \text{log}p(y|x_t) = -1/\sigma_t(\epsilon_{x_t; y, t} - \epsilon_{x_t, t})$ similar to CFG~\cite{cfg}. Thus, we have
\begin{align}
    &\nabla_\theta \mathcal{L}^{'}_G = \nabla_\theta \mathbb{E}_{t,\epsilon}[\text{log}p(\phi|x_t^g) - \lambda\text{log}{p(y|x_t^g)}]\label{eq:grad_g}\\
    &= \mathbb{E}_{t, \epsilon}[\omega(t)(\epsilon_{x_t^g; t} + \lambda (\epsilon_{x_t^g; y, t} -\epsilon_{x_t^g; t}) - \epsilon_{x_t^g; \phi, t})\frac{\partial x_0^g}{\partial \theta}]\nonumber
\end{align}
where $\omega(t)$ is a weight based on the time $t$. Compare the SDS gradient in Eq.~\eqref{eq:grad_sds} with the Eq.~\eqref{eq:grad_g}, the only difference is that the SDS gradient uses the added noise $\epsilon$ to approximate the noise $\epsilon_{x_t ^ g; \phi, t}$. This approximation may not be optimal, because $\epsilon_{x_t^g; \phi, t}$ should come from the diffusion model, so that $\epsilon_{x_t^g; \phi, t} - \epsilon_{x_t^g; t}$ approximates $-\sigma_t\nabla_{x_t} \text{log}p(\phi|x_t)$. From another intuitive perspective,  $\epsilon_{x_t^g; \phi, t}$ contains the inductive bias of the pretrained diffusion model while $\epsilon$ does not. Recently, VSD~\cite{vsd} uses the LoRA branch to predict $\epsilon_{x_t^g; \phi, t}$ which also contains the inductive bias of the pretrained diffusion model, thereby achieving better quality.

Let us go further into discriminator optimization. We have the discriminator loss, based on Eq.~\eqref{eq:wgan} as
\begin{align}
     &\small{\mathcal{L}_D = \mathbb{E}_{t, \epsilon}[D(x_t^g; y) - D(x_t^r; y) + \tau\lVert \nabla_{\tilde{x}_t} D(\tilde{x}_t)\rVert_2^2]} \label{eq:d_loss} \\
     &\small{= \mathbb{E}_{t,\epsilon}[\text{log}\frac{p(y|x_t^g)}{p(\phi|x_t^g)} - \text{log}\frac{p(y|x_t^r)}{p(\phi|x_t^r)}+ \tau\lVert \nabla_{\tilde{x}_t} \log \frac{p(y|\tilde{x}_t)}{p(\phi|\tilde{x}_t)}\rVert_2^2]} \nonumber 
\end{align}
where $x_t^r = \alpha_t x_0^r + \sigma_t \epsilon$ is the noisy real sample based on the real sample $x_0^r$ and the time $t$. In Eq.~\eqref{eq:d_loss}, $p(y|x_t^r)$ and
$p(y|x_t^g)$ are based on the given pretrained diffusion model, so 
they are not optimizable. $\phi$ in $p(\phi|x_t^g)$ and $p(\phi|x_t^r)$ is optimizable, which can be implemented using an textual-inversion embedding~\cite{text_inervsion} or LoRA~\cite{lora}. Thus, $\mathcal{L}_D$ is a loss function only about $\phi$, its gradient can be derived as
\begin{align}
    &\small{\nabla_\phi \mathcal{L}_D = \nabla_\phi \mathbb{E}_{t,\epsilon}[\text{log}\frac{p(\phi|x_t^r)}{p(\phi|x_t^g)} + \tau\lVert \nabla_{\tilde{x}_t} \log \frac{p(y|\tilde{x}_t)}{p(\phi|\tilde{x}_t)}\rVert_2^2]}\nonumber\\
    &\small{= \nabla_\phi \mathbb{E}_{t,\epsilon}[\text{log}\frac{p(x_t^r|\phi)}{p(x_t^g|\phi)} + \tau\lVert \nabla_{\tilde{x}_t} \log \frac{p(\tilde{x}_t|y)}{p(\tilde{x}_t|\phi)}\rVert_2^2]}\label{eq:grad_d}\\
    &\small{\approx \nabla_\phi \mathbb{E}_{t,\epsilon}[\lVert\epsilon_{x_t^g; \phi, t} - \epsilon\rVert_2^2 - \lVert \epsilon_{x_t^r; \phi, t} - \epsilon\rVert_2^2} \nonumber\\
    &\small{\quad + \underbrace{\eta \lVert \epsilon_{x_t^r; y, t} - \epsilon_{x_t^r;\phi, t}\rVert_2^2 + \gamma\lVert \epsilon_{x_t^g; y, t} - \epsilon_{x_t^g;\phi, t}\rVert_2^2}_{\text{penalty term}}]}\nonumber
\end{align}
where $\tilde{x}$ comes from a mixture distribution $\tilde{\mu}$, so it can be replaced by a combination of $x_t^r$ and $x_t^g$. $\eta$ and $\gamma$ are adjustable hyperparameters, to keep the penalty term positive, they are subject to $\gamma \geq - \eta \lVert \epsilon_{x_t^r; y, t} - \epsilon_{x_t^r;\phi, t}\rVert_2^2 /  \lVert \epsilon_{x_t^g; y, t} - \epsilon_{x_t^g;\phi, t}\rVert_2^2$. In practice, we find $\eta = 1/2$, $\gamma \in [-1, 0)$ works for most cases.  See supplementary materials for the derivation details of Eq.~\eqref{eq:grad_d}. In SDS~\cite{sds}, this discriminator optimization is completely ignored, as there is no optimizable $\phi$. VSD~\cite{vsd} uses the LoRA or simple UNet branch to implement $\phi$, and updates it using $\nabla_\phi \lVert\epsilon_{x_t^g; \phi, t} - \epsilon\rVert_2^2$, which is just a part of Eq.~\eqref{eq:grad_d}.

\begin{figure}[t]
    \centering 
    \small
    \includegraphics[width=.98\linewidth]{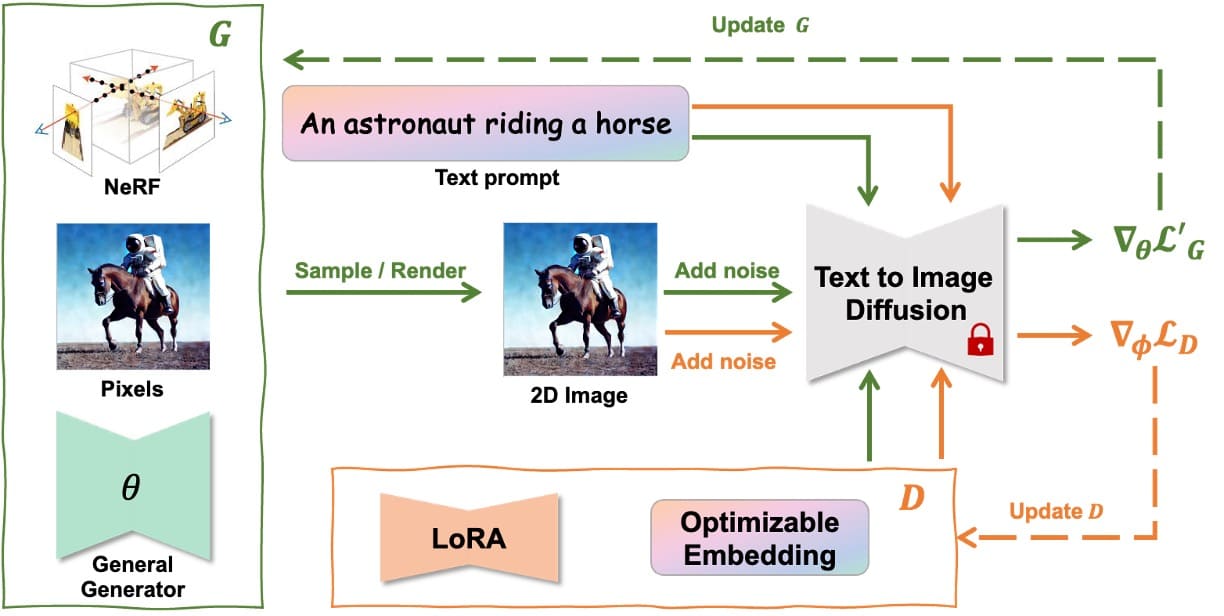}
\caption{Workflow of ASD. Green lines show the pipeline of generator optimization. Orange lines show the pipeline of discriminator optimization. The avatar of NeRF is adapted from~\cite{nerf}. See the supplementary for the algorithm description.}
\label{fig:workflow}
\end{figure}

\vspace{1mm}
We propose Adversarial Score Distillation (ASD) based on the WGAN paradigm. Figure~\ref{fig:workflow} illustrates the workflow of ASD. ASD uses Eq.~\eqref{eq:grad_g} to update the parameters $\theta$ of a generator. This generator could be a NeRF~\cite{nerf, instant_ngp} in the text-to-3d task, or a general generator similar to that in~\cite{diff_instruct}, or just image pixels in the 2D distillation and image editing tasks. ASD maintains an optimizable discriminator by implementing $\phi$ with an optimizable conditional embedding~\cite{text_inervsion} or LoRA~\cite{lora}, and updates this discriminator using the complete $\mathcal{L}_D$. However, in Eq.~\eqref{eq:grad_d}, the noisy real sample $x_t^r$ is unavailable for text-conditioned cases. For these cases, we can use the upper bound of $\mathcal{L}_D$, which does not contain $x_t^r$ terms.

\vspace{1mm}
\noindent{\textbf{$\mathcal{L}'_D$ for text-conditioned distillation.}} When $\eta = {1}/{2}$, based on triangle and Cauchy–Schwarz inequality, we have $\frac{1}{2}\lVert \epsilon_{x_t^r; y, t} - \epsilon_{x_t^r;\phi, t}\rVert_2^2 \leq \lVert \epsilon_{x_t^r; y, t} - \epsilon\rVert_2^2 + \lVert\epsilon_{x_t^r;\phi, t} - \epsilon\rVert_2^2$. Thus, we can rewrite Eq.~\eqref{eq:grad_d} as
\begin{align}
    \nabla_\phi \mathcal{L}_D \leq 
    &\nabla_\phi \mathbb{E}_{t,\epsilon}[\lVert\epsilon_{x_t^g; \phi, t} - \epsilon\rVert_2^2 \nonumber\\
    & + \gamma\lVert \epsilon_{x_t^g; y, t} - \epsilon_{x_t^g;\phi, t}\rVert_2^2] := \nabla_\phi\mathcal{L}'_D
    \label{eq:loss_d_text}
\end{align}
For the case where real samples are unavailable, we use Eq.~\eqref{eq:loss_d_text} to update the prompt $\phi$ of ASD. For cases where real samples are available, like image editing or image / 3D-conditioned distillation, we can continue to use Eq.~\eqref{eq:grad_d}.

\begin{figure}[t]
    \centering 
    \small
    \addtolength{\tabcolsep}{-5.5pt}
    \begin{tabular}{cccc}
    \includegraphics[width=.245\linewidth]{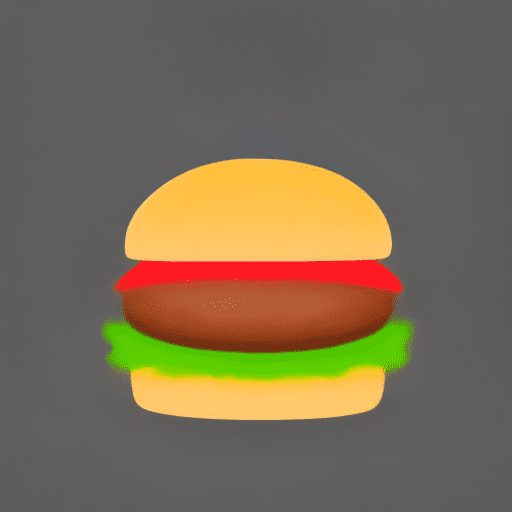}&
    \includegraphics[width=.245\linewidth]{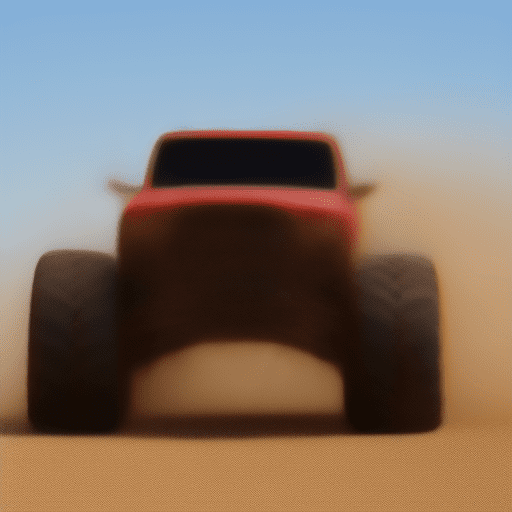}&
    \includegraphics[width=.245\linewidth]{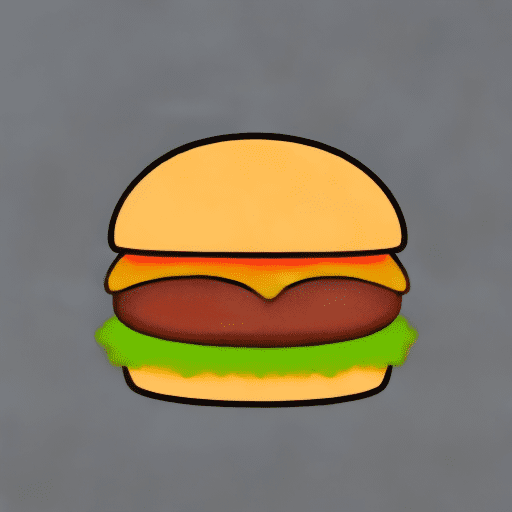}&
    \includegraphics[width=.245\linewidth]{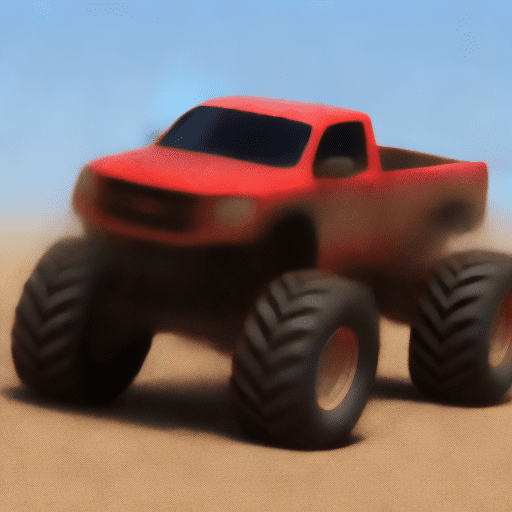}\\
    \multicolumn{2}{c}{SDS (CFG = 7.5)}&
    \multicolumn{2}{c}{VSD$^\dagger$ (CFG = 7.5)}
    \end{tabular}
\caption{2D score distillation results with the prompts ``hamburger'' and ``a monster truck''. VSD$^\dagger$ denotes that the LoRA branch is updated with 50 steps per iteration, resulting in over-smoothing images similar to SDS.}
\label{fig:multiple_step_vsd}
\end{figure}

\vspace{1mm}
\noindent{\textbf{$\mathcal{L}_G^{e}$ for image editing.}} We extend ASD to image editing by using two discriminators with source prompts $z$ and target prompts $y$. The new discriminator function is defined as $F(x_t; y, z) = D(x_t;y) - D(x_t;z)$, which gives high confidence for samples closing to the target prompt $y$ but away from the source prompts $z$. Then, the generator loss will be
\begin{align}
    \small{
    \nabla_{\theta} \mathcal{L}_G^{e}}&\small{=\nabla_\theta \mathbb{E}_{t,\epsilon}[\text{log}p(\phi_y|x_t^g) - \lambda\text{log}{p(y|x_t^g)}}\nonumber\\
    &\small{\quad - \text{log}p(\phi_z|x_t^g) + \lambda\text{log}{p(z|x_t^g)}]}\label{eq:loss_g_edit}\\
    &\small{= \mathbb{E}_{t, \epsilon}[\omega(t)\Big(\lambda (\epsilon_{x_t^g; y, t} -\epsilon_{x_t^g; t}) - (\epsilon_{x_t^g; \phi_y, t} - \epsilon_{x_t^g; t})}\nonumber\\
    &\small{\quad - \lambda (\epsilon_{x_t^g; z, t} -\epsilon_{x_t^g; t}) + (\epsilon_{x_t^g; \phi_z, t} - \epsilon_{x_t^g; t})\Big)\frac{\partial x_0^g}{\partial \theta}]}\nonumber
\end{align}
For discriminator optimization, we can use either Eq.~\eqref{eq:grad_d} or Eq.~\eqref{eq:loss_d_text} in the image editing task.
We find that DDS~\cite{dds} is a special case of our paradigm. Specifically, DDS employs two fixed discriminators (similar to SDS), and feeds the noisy source image into the second discriminator $D(x_t;z)$. 

\section{Comparison and Discussion}
\noindent\textbf{Score Distillation Sampling.} In practice, the SDS gradient comes from the generator loss in WGAN instead of the L2 diffusion loss. It maintains a fixed sub-optimal discriminator by using $\epsilon$ to approximate $\epsilon_{x_t^g; \phi, t}$, which ignores the inductive bias of the pretrained diffusion model. In contrast, ASD uses an optimizable conditional embedding or LoRA to implement $\phi$, thus keeping the inductive bias in $\epsilon_{x_t^g; \phi, t}$.

\noindent\textbf{Variational Score Distillation.} 
From the methodology perspective, we employ the WGAN paradigm to explain and analyze VSD instead of the Wasserstein gradient flow, which is more straightforward. Based on the Wasserstein gradient flow~\cite{vsd}, for each iteration, VSD needs to train the score function of the particle distribution (\ie, the LoRA branch) until convergence. However, some studies~\cite{wgan_fail, how_w} point out that WGAN may fail when approximating the very accurate Wasserstein distance, which means that updating the LoRA branch with too many steps per iteration may degrade the results. Figure~\ref{fig:multiple_step_vsd} shows the results if we update the LoRA branch with 50 steps per iteration. In addition, the discriminator optimization of VSD is just a part of the WGAN discriminator loss, which can be considered as a special case of ASD when setting $\gamma$ in Eq.~\eqref{eq:loss_d_text} to 0.

\begin{figure}[ht]
    \centering 
    \small
    \addtolength{\tabcolsep}{-5.5pt}
    \begin{tabular}{cccc}
    \includegraphics[width=.245\linewidth]{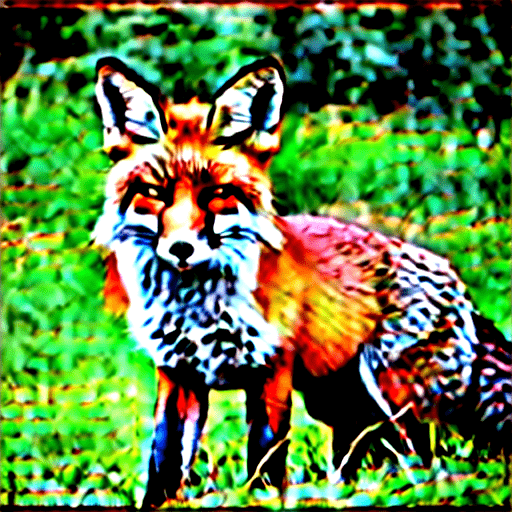}&
    \includegraphics[width=.245\linewidth]{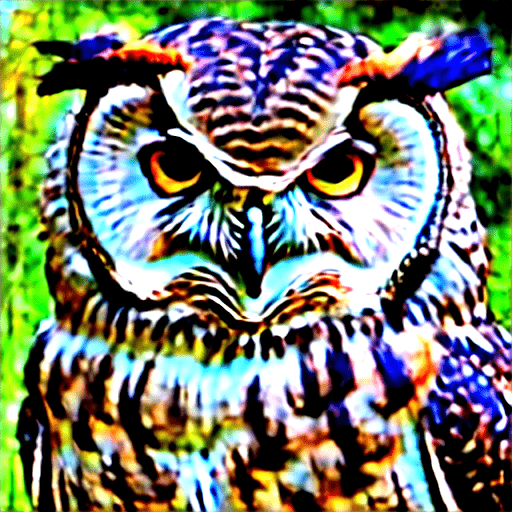}&
    \includegraphics[width=.245\linewidth]{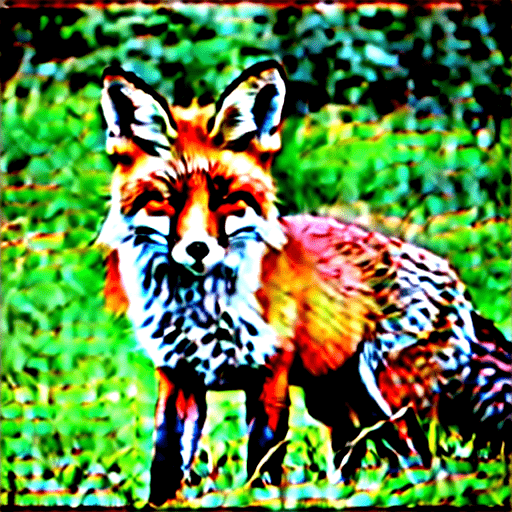}&
    \includegraphics[width=.245\linewidth]{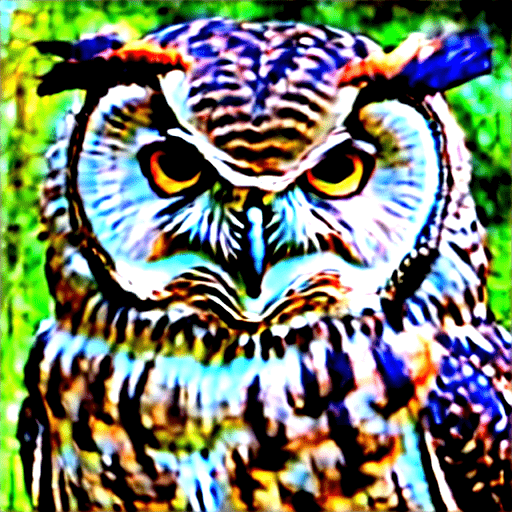}\\
    \multicolumn{2}{c}{SDS (CFG = 10k)}&
    \multicolumn{2}{c}{Only classifier term}\\
    \multicolumn{4}{c}{\vspace{2mm}\normalsize{(a) 2D score distillation results.}}\\
    \includegraphics[width=.245\linewidth]{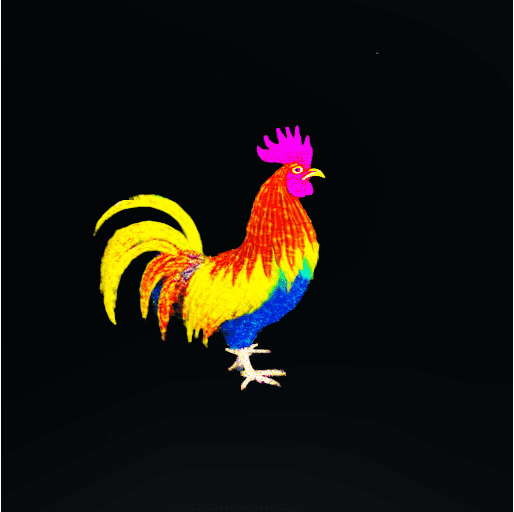}&
    \includegraphics[width=.245\linewidth]{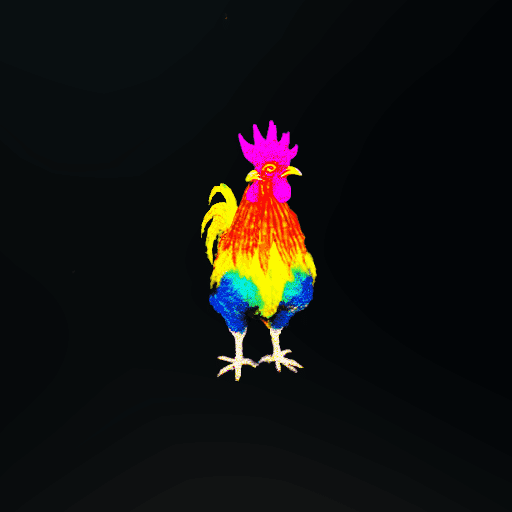}&
    \includegraphics[width=.245\linewidth]{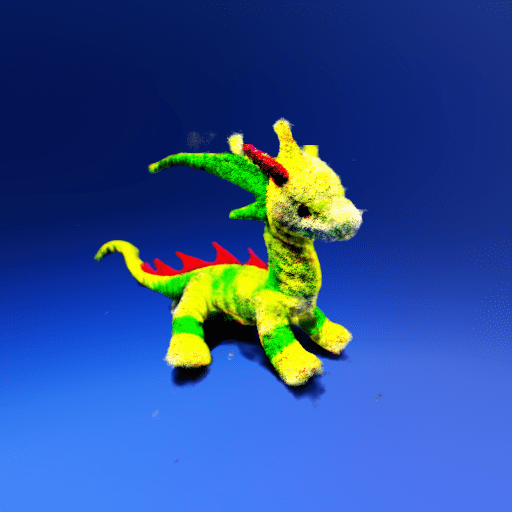}&
    \includegraphics[width=.245\linewidth]{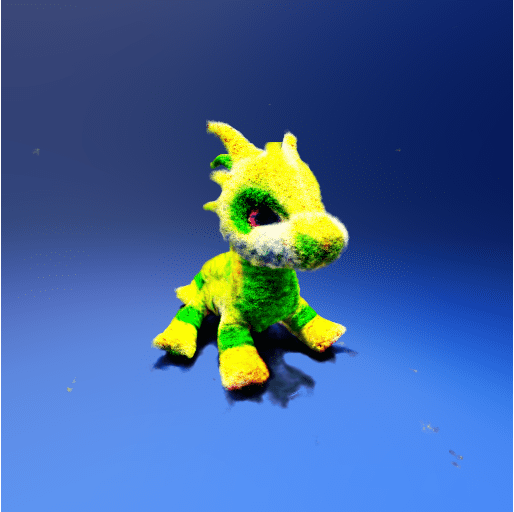}\\
    \multicolumn{4}{c}{\vspace{2mm}\normalsize{(b) 3D results by using the classifier term only.}}\\
    \multicolumn{4}{c}{Optimization Steps\vspace{-2mm}}\\
    \multicolumn{4}{c}{\hspace{2mm}\vspace{-2mm}\includegraphics[width=.8\linewidth]{figures/arrow.png}}\\
    \includegraphics[width=.245\linewidth]{figures/csd_fox.png}&
    \includegraphics[width=.245\linewidth]{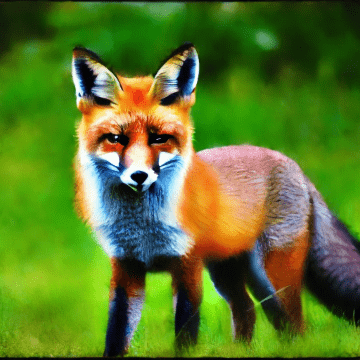}&
    \includegraphics[width=.245\linewidth]{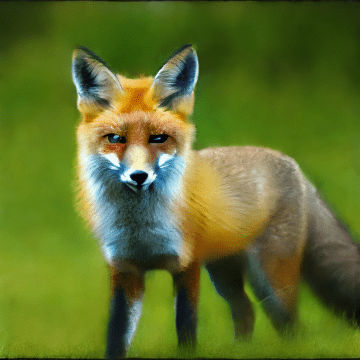}&
    \includegraphics[width=.245\linewidth]{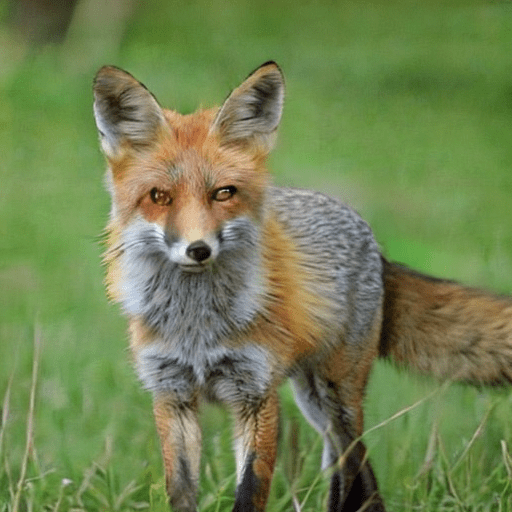}\\
    \includegraphics[width=.245\linewidth]{figures/csd_owl.png}&
    \includegraphics[width=.245\linewidth]{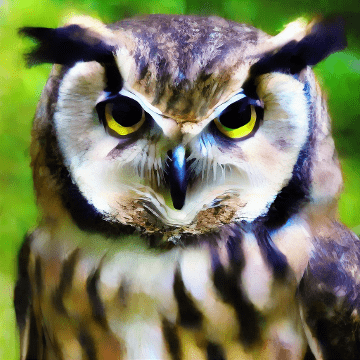}&
    \includegraphics[width=.245\linewidth]{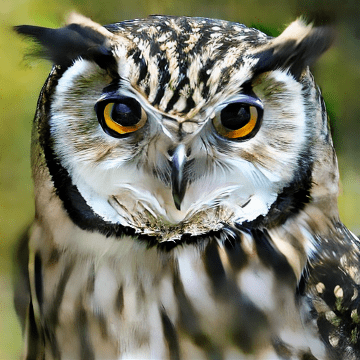}&
    \includegraphics[width=.245\linewidth]{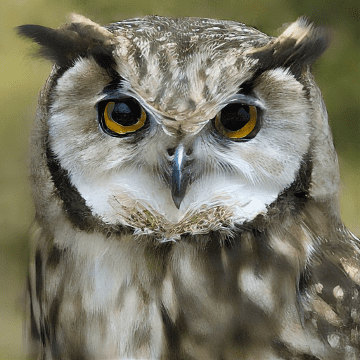}\\
    \multicolumn{4}{c}{\normalsize{(c) ASD recovers from over-saturated initialization.}}
    \end{tabular}
\caption{Score distillation results with the prompt ``a photograph of a fox'', ``a front view of an owl'' in 2D, and ``a colorful rooster'', ``a plush dragon toy'' in 3D. Only using the classifier term is equivalent to using SDS with a huge CFG scale, which tends to get over-saturated results in both 2D and 3D distillation. ASD can recover from over-saturated initialization.}
\label{fig:csd_vs_asd}
\vspace{-5mm}
\end{figure}

\noindent\textbf{Classifier Score Distillation.} Very recently, the concurrent work Classifier Score Distillation (CSD)~\cite{csd} highlights the importance of the classifier term in the SDS gradient. We observe that only using the classifier term is similar to using SDS with a very huge CFG scale (as the learning rate can be adjusted dynamically by the optimizer, the percentage of the gradient component is the key factor), thereby prone to over-saturation in both 2D and 3D distillation. As a very important trick, CSD proposes to use negative prompts to alleviate over-saturation, which is similar to the discriminator in Eq~\eqref{eq:discriminator} with fixed $\phi$. Thanks to the adversarial capabilities, ASD can recover from over-saturation even with the over-saturated initialization. See Figure~\ref{fig:csd_vs_asd} for visualization.

\noindent\textbf{Disscussion of $\gamma$.} When $\gamma = 0$, Eq~\eqref{eq:loss_d_text} becomes VSD, which can be considered as the WGAN~\cite{wgan,wgan_div} discriminator loss with fixed penalty term for Lipschitz constraint. For $\gamma < 0$, we decrease the weight of the penalty term, which can enhance the ability of the discriminator. For $\gamma > 0$, we 

\begin{figure}[H]
    \centering 
    \small
    \addtolength{\tabcolsep}{-5.5pt}
    \begin{tabular}{cccc}
    T2I & SDS@100 & VSD@7.5 & ASD@7.5\\
    \includegraphics[width=.245\linewidth]{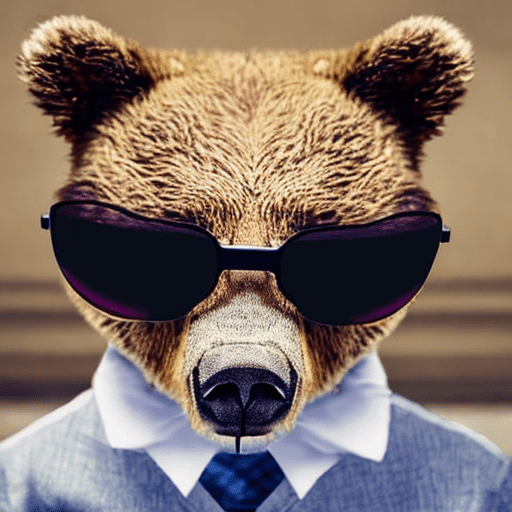}&
    \includegraphics[width=.245\linewidth]{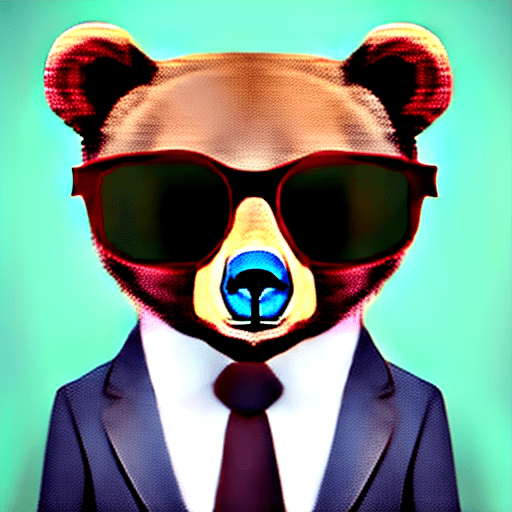}&
    \includegraphics[width=.245\linewidth]{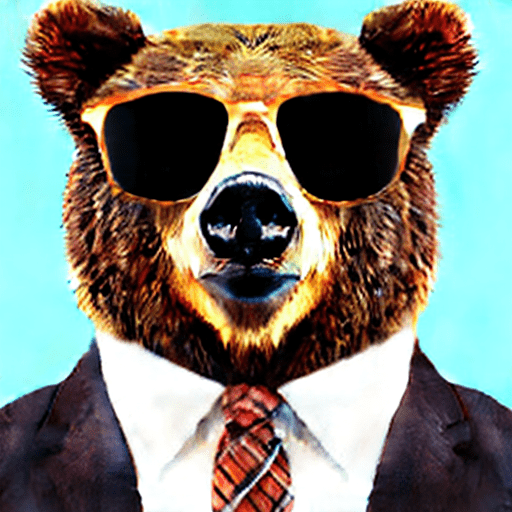}&
    \includegraphics[width=.245\linewidth]{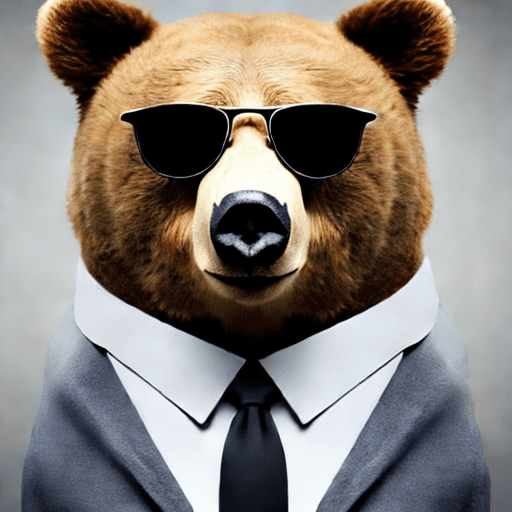}\\
    \multicolumn{4}{c}{\vspace{1.2mm}\textit{a bear wearing sunglasses and a tie looks very proud}}\\    
    \includegraphics[width=.245\linewidth]{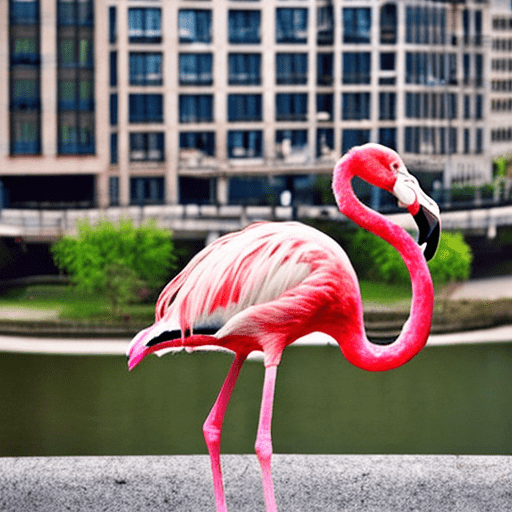}&
    \includegraphics[width=.245\linewidth]{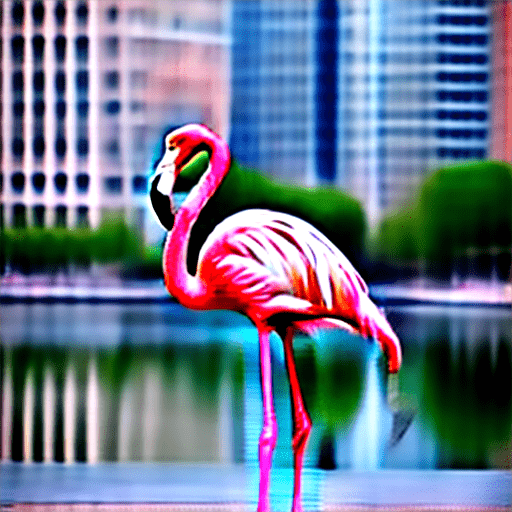}&
    \includegraphics[width=.245\linewidth]{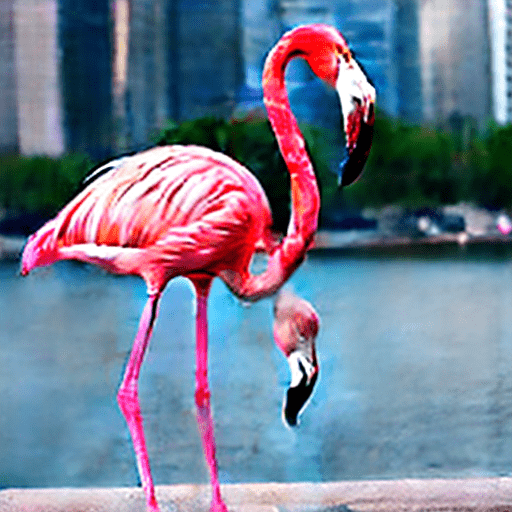}&
    \includegraphics[width=.245\linewidth]{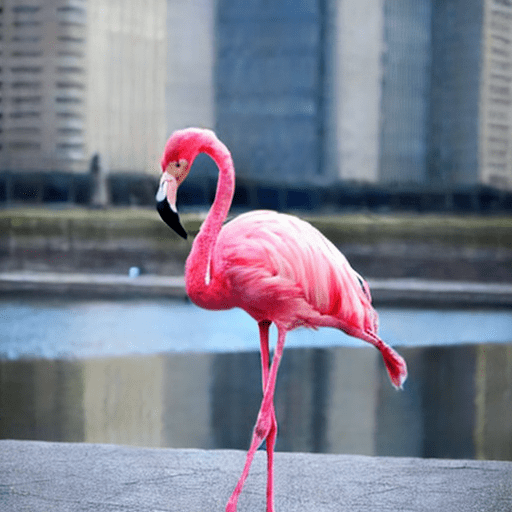}\\
    \multicolumn{4}{c}{\vspace{1.2mm}\textit{a flamingo in the city}}\\    
    \includegraphics[width=.245\linewidth]{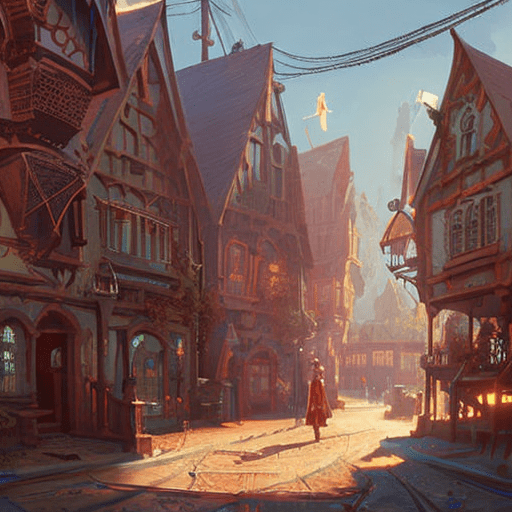}&
    \includegraphics[width=.245\linewidth]{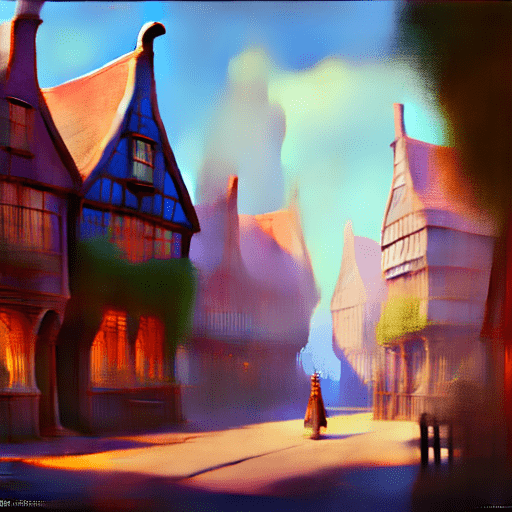}&
    \includegraphics[width=.245\linewidth]{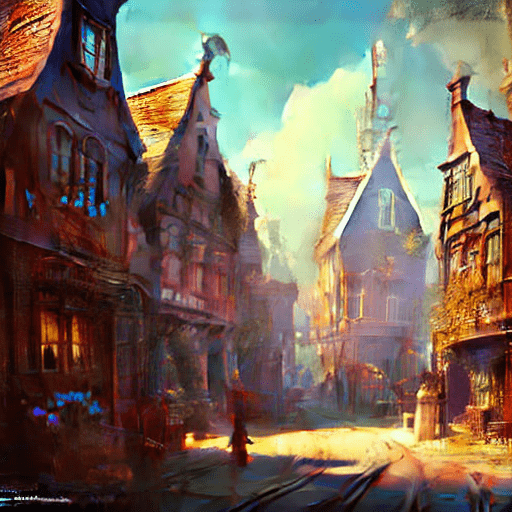}&
    \includegraphics[width=.245\linewidth]{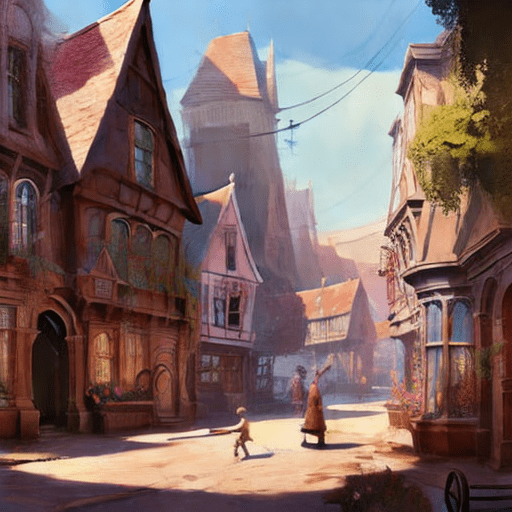}\\
    \multicolumn{4}{c}{\tabincell{c}{\textit{a lively magical town inspired by Victorian England} \\
                       \textit{and Amsterdam}}\vspace{1mm}}\\
    \multicolumn{4}{c}{\vspace{2mm}\normalsize{(a) Comparisons of 2D score distillation.}}
    \end{tabular}
    \begin{tabular}{ccc}
    SDS@100 & VSD@7.5 & ASD@7.5\\
    \includegraphics[width=.325\linewidth]{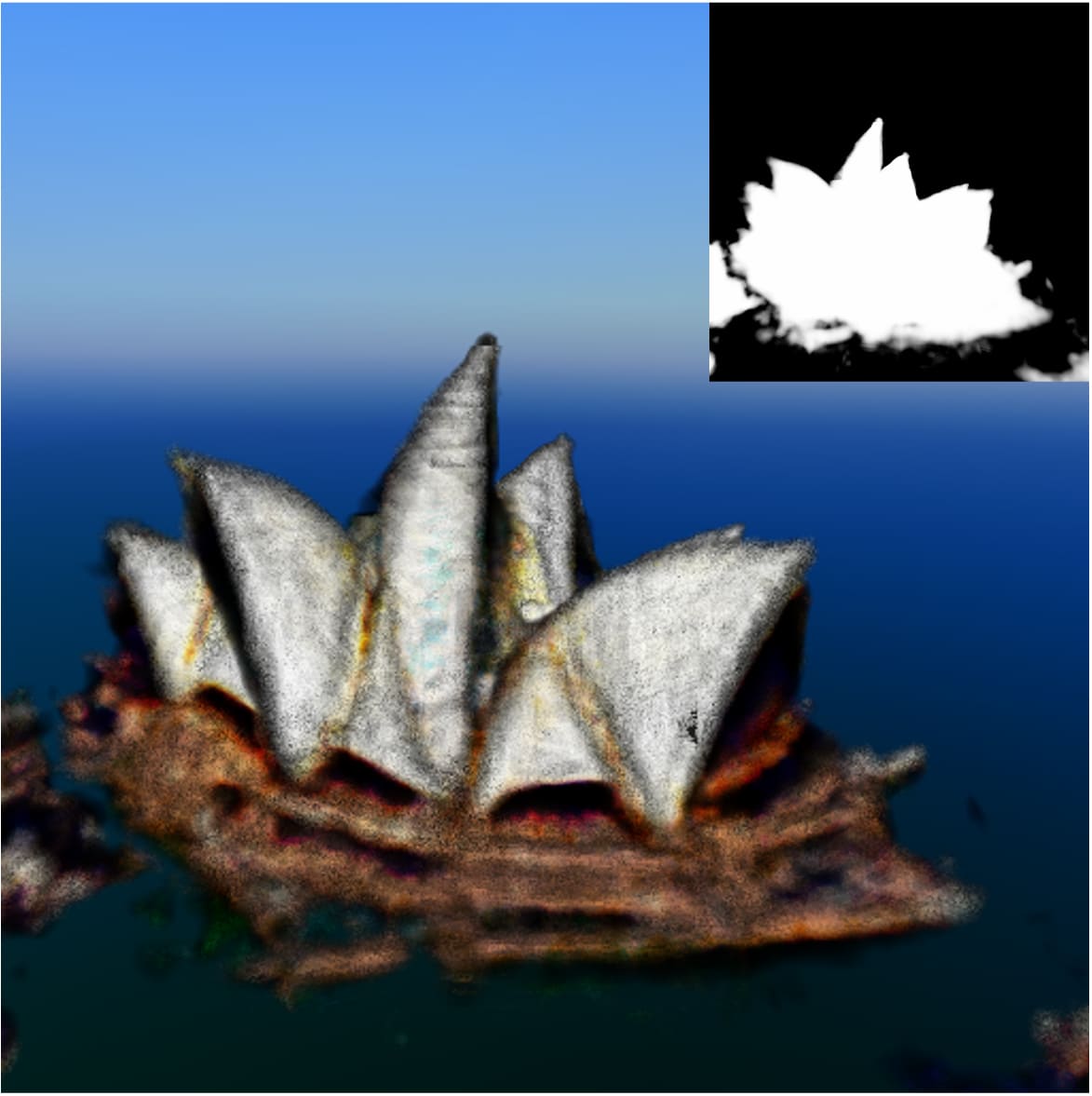}&
    \includegraphics[width=.325\linewidth]{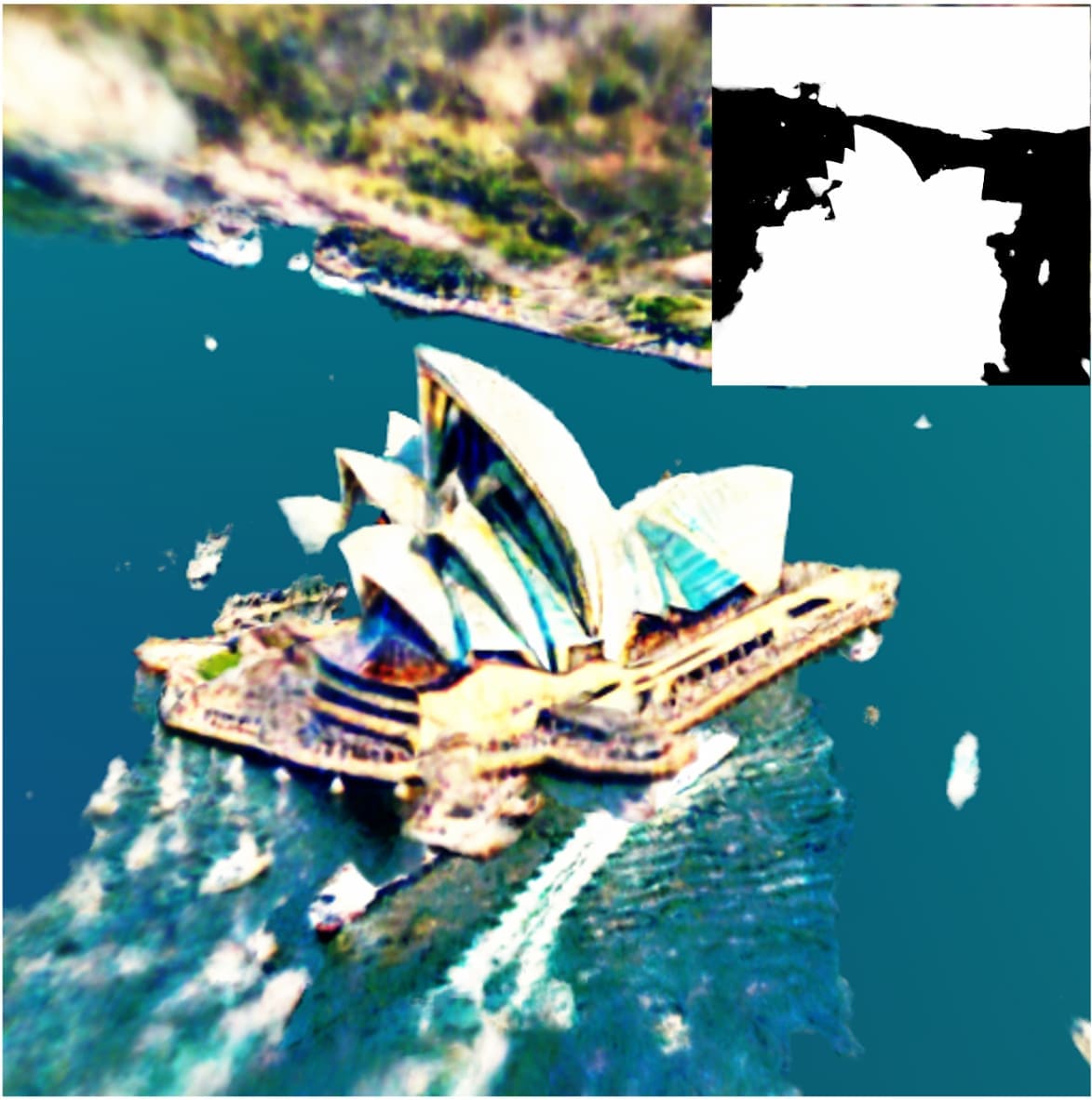}&
    \includegraphics[width=.325\linewidth]{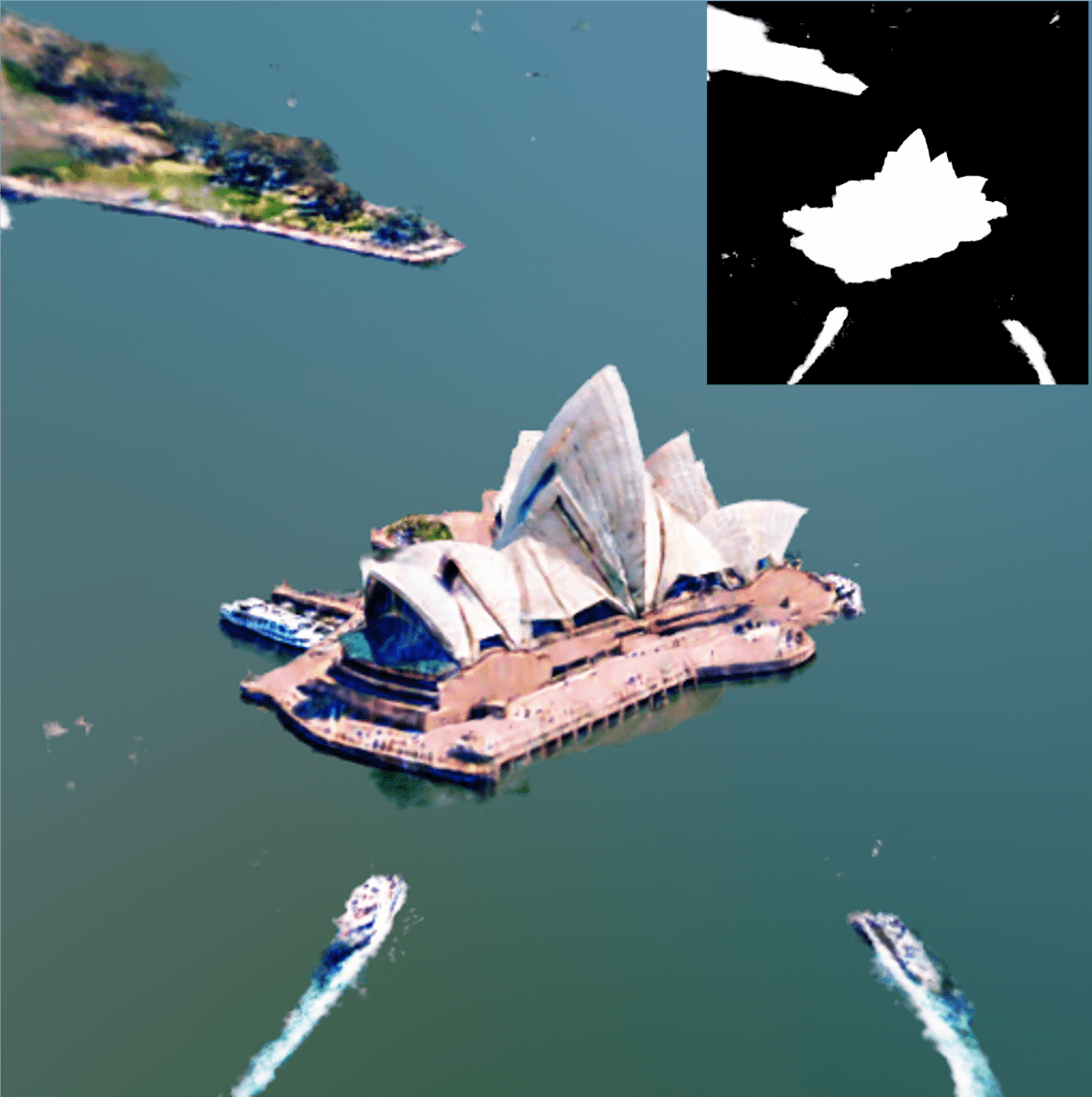}\\
    \multicolumn{3}{c}{\vspace{1.2mm}\textit{Sydney opera house, aerial view}}\\
    \includegraphics[width=.325\linewidth]{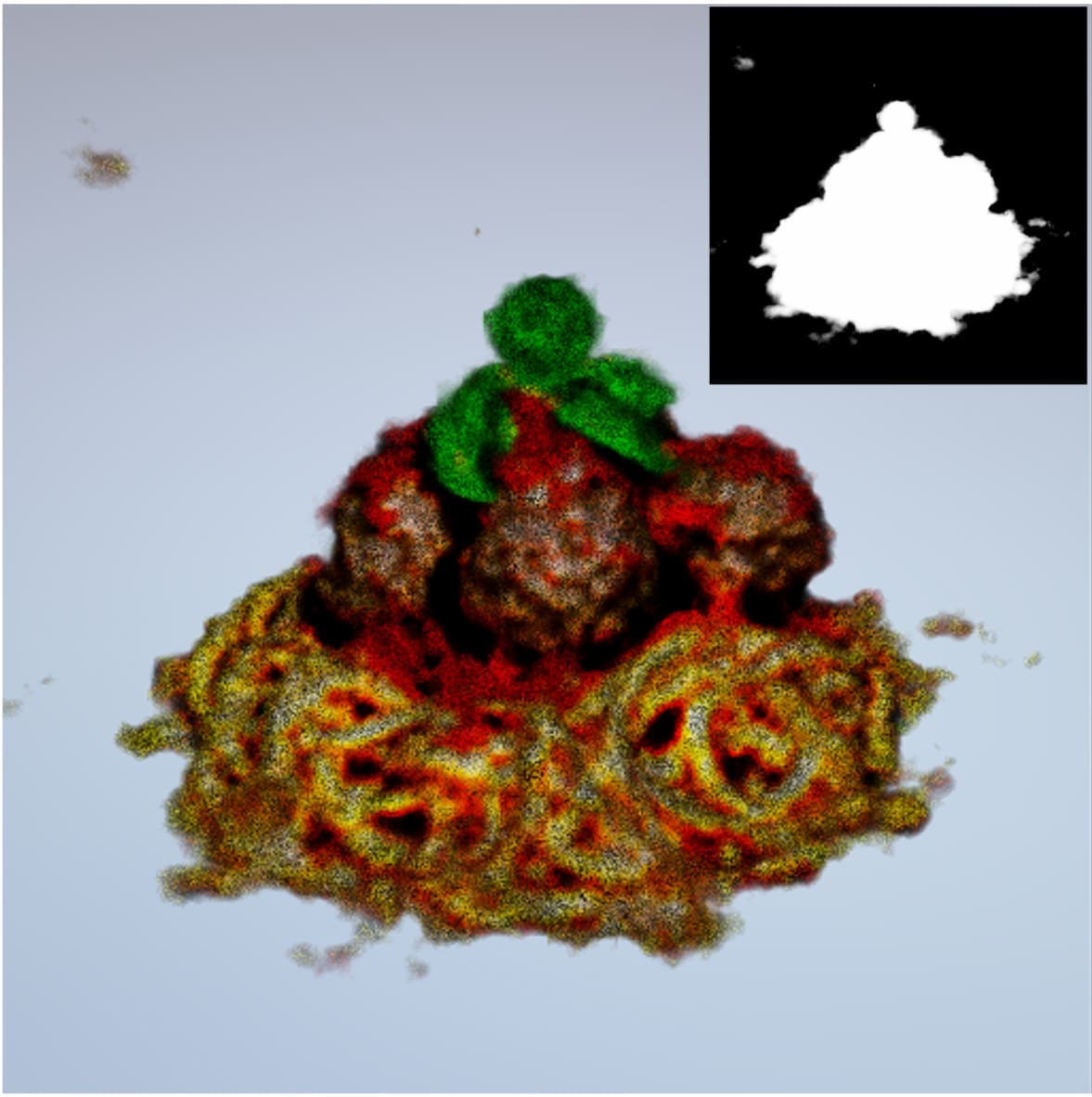}&
    \includegraphics[width=.325\linewidth]{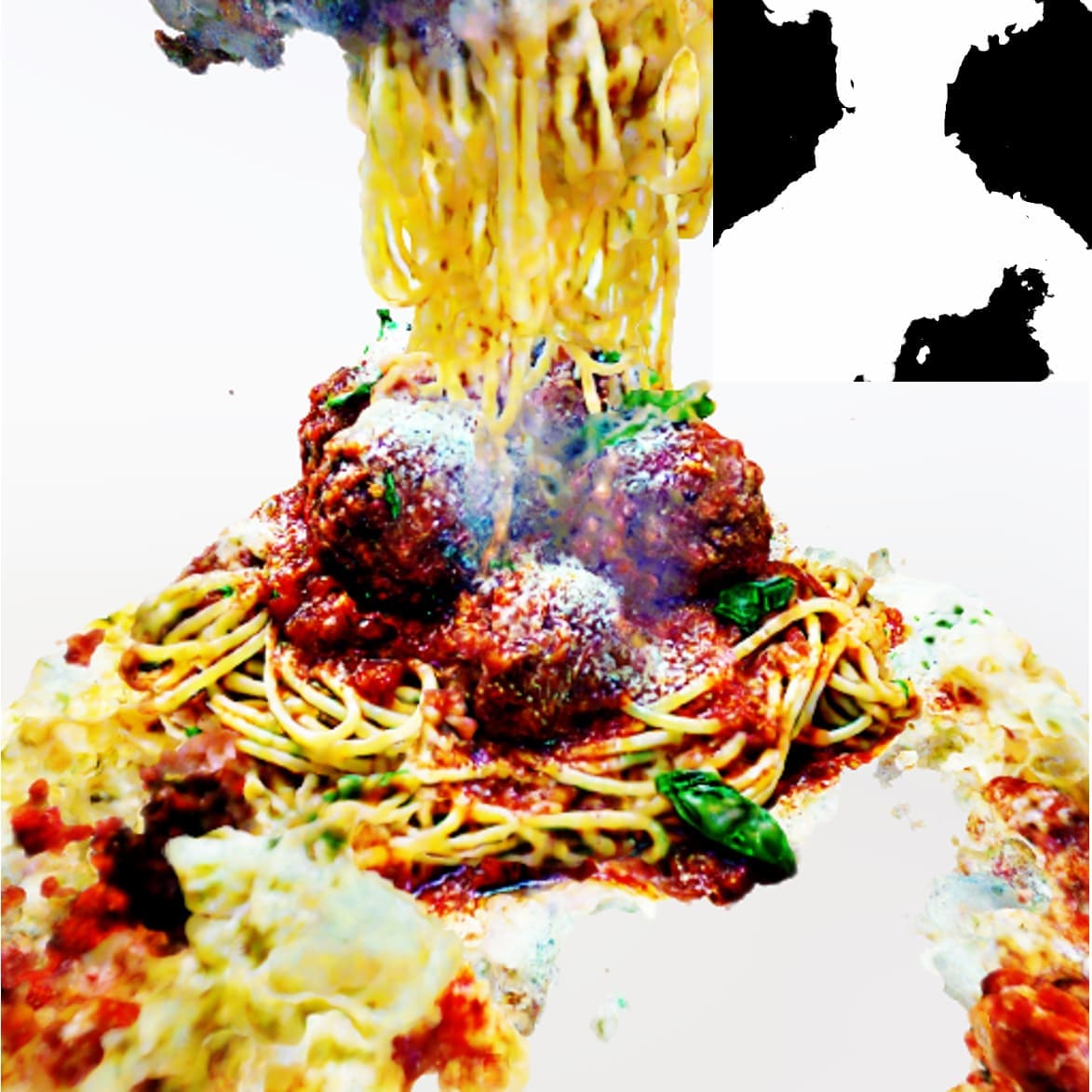}&
    \includegraphics[width=.325\linewidth]{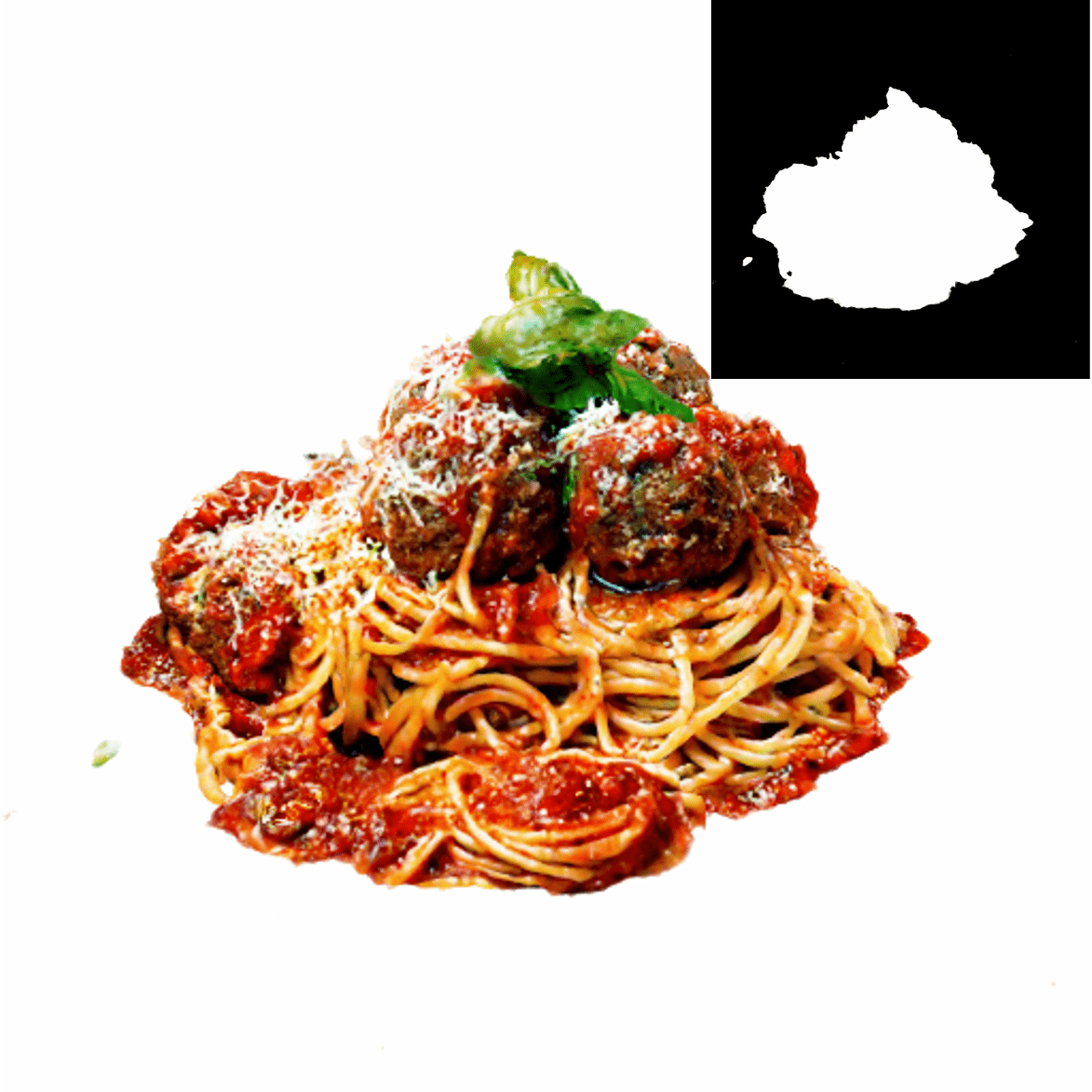}\\
    \multicolumn{3}{c}{\vspace{1.2mm}\textit{a steaming hot plate piled high with spaghetti and meatballs}}\\
    \includegraphics[width=.325\linewidth]{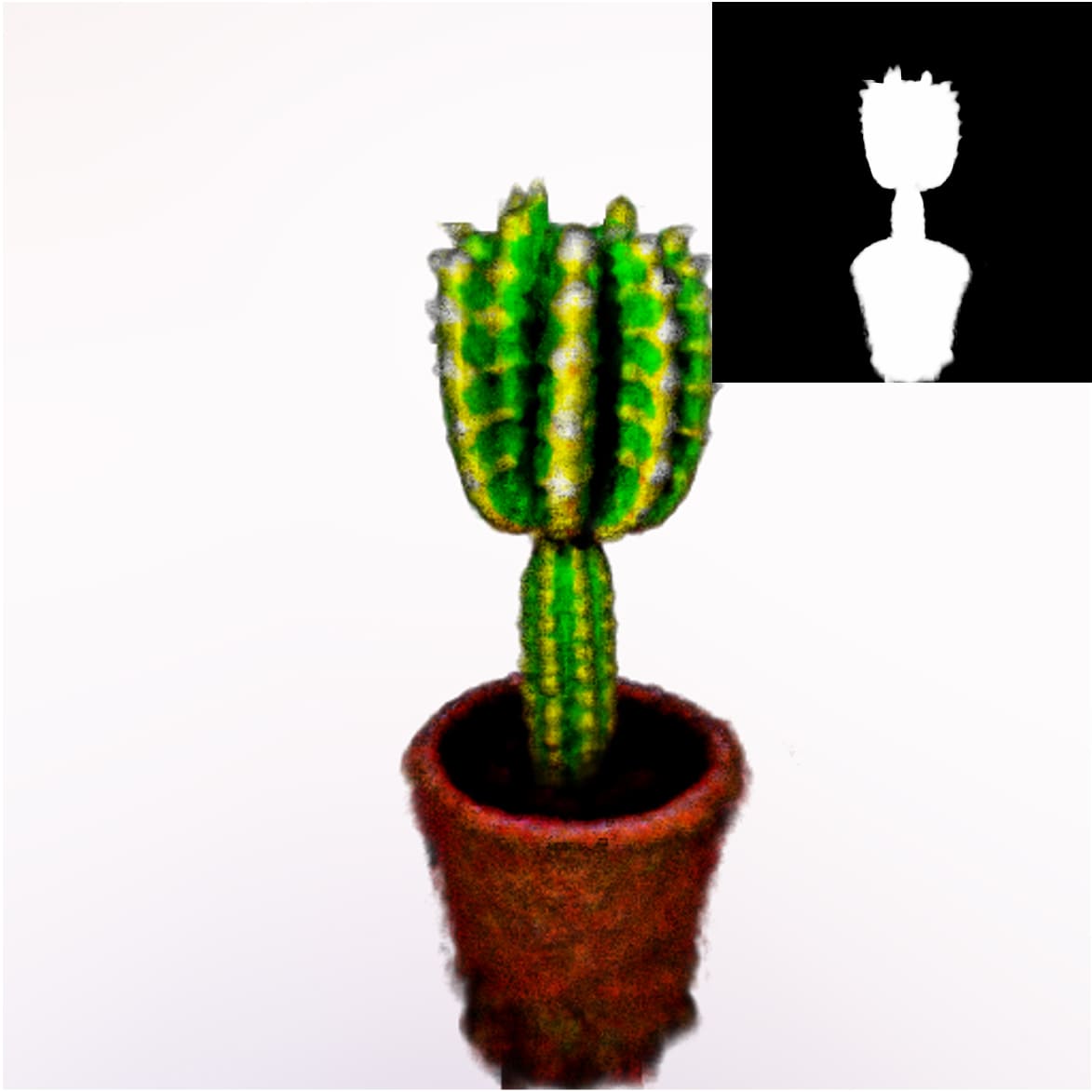}&
    \includegraphics[width=.325\linewidth]{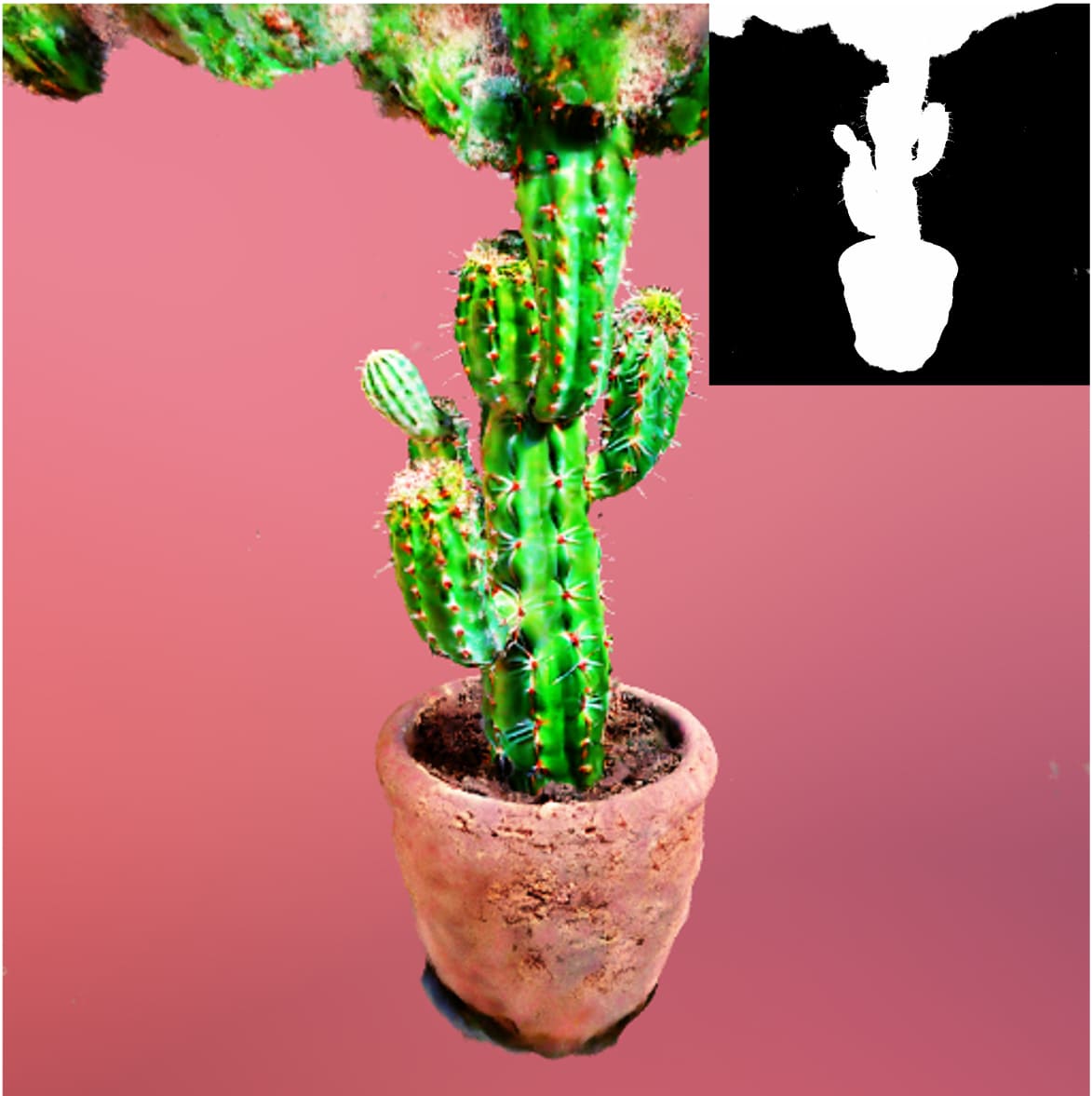}&
    \includegraphics[width=.325\linewidth]{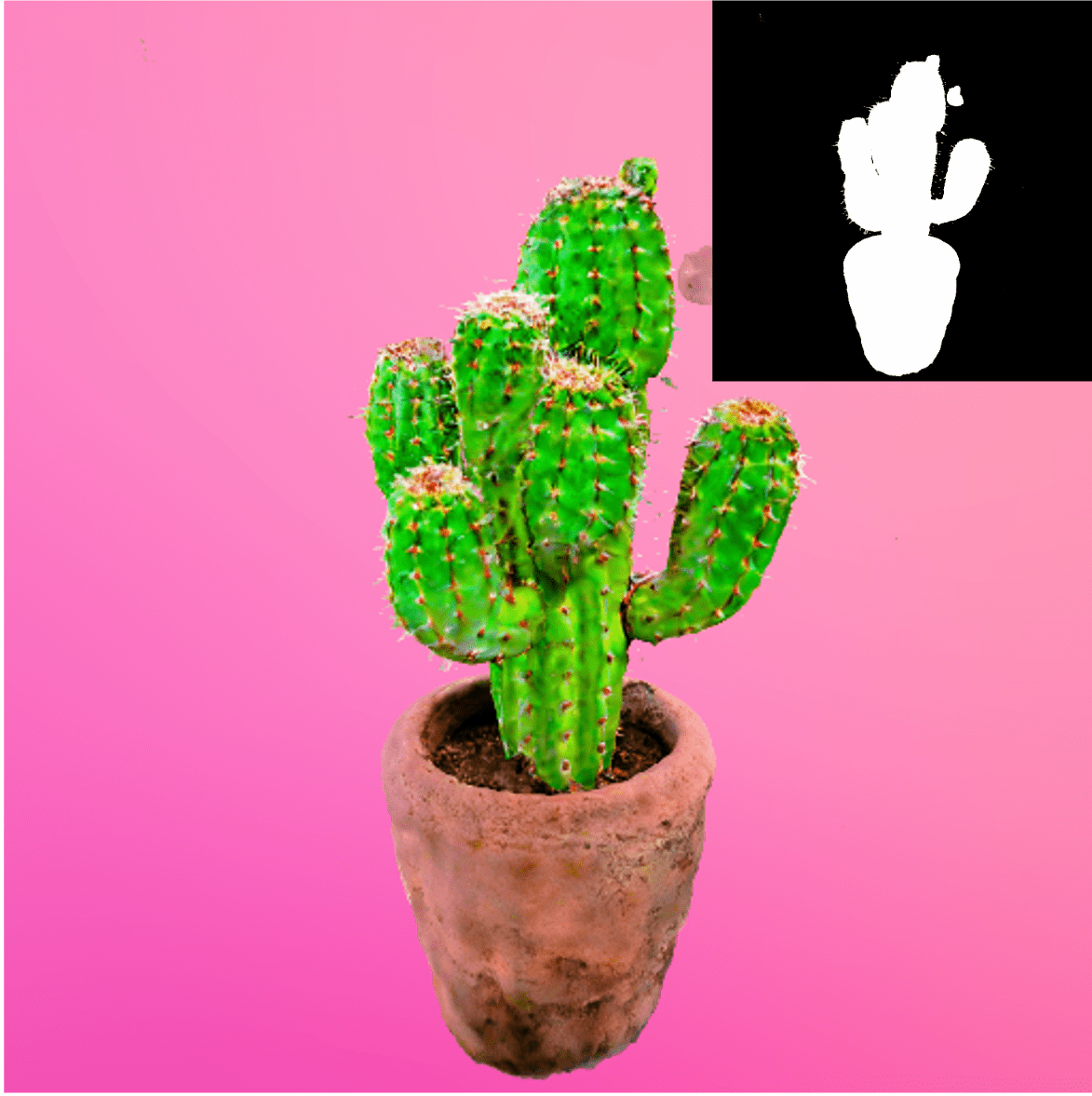}\\
    \multicolumn{3}{c}{\vspace{1mm}\textit{a small saguaro cactus planted in a clay pot}}\\
    \multicolumn{3}{c}{\normalsize{(b) Comparisons of text-conditioned 3D NeRF distillation.}}
    \end{tabular}
\caption{Quality comparisons of distillation results among SDS, VSD, and ASD. For the text-to-3D task, the effects of different representations vary widely, so we all compared the results from NeRF distillation for fair play. ``@7.5'' denotes ``at the CFG scale 7.5''. ``T2I'' denotes text-to-image results.}
\label{fig:quality}
\end{figure}

\noindent increase the weight of the penalty term. Specifically, when $\gamma \gg 0$, the penalty term will dominant Eq~\eqref{eq:loss_d_text}, which leads to $\epsilon_{x_t^g,y,t} \approx \epsilon_{x_t^g,\phi,t}$. Then the discriminator fails to recognize fake samples, and the gradient for generators~\eqref{eq:g_loss} mainly comes from the implicit classifier term. Only using the classifier term can be considered as a special case of our paradigm in this situation.

\begin{figure}[t]
    \centering 
    \small
    \addtolength{\tabcolsep}{-5.8pt}
    \begin{tabular}{ccccc}
    \multicolumn{5}{c}{Optimization Steps\vspace{-2mm}}\\
    \multicolumn{5}{c}{\hspace{2mm}\vspace{-2mm}\includegraphics[width=.8\linewidth]{figures/arrow.png}}\\
    \raisebox{0.2in}{\rotatebox{90}{VSD}}
    \includegraphics[width=.19\linewidth]{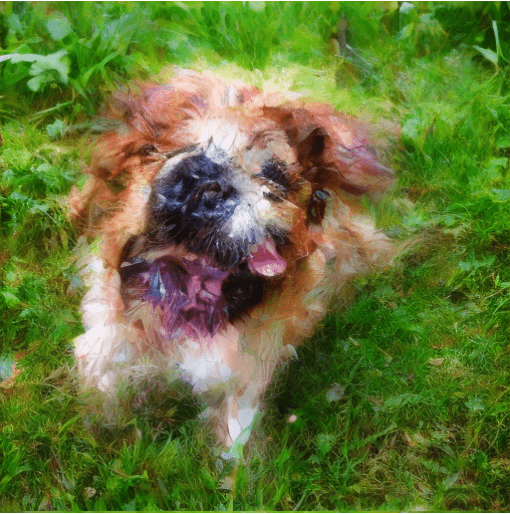}&
    \includegraphics[width=.19\linewidth]{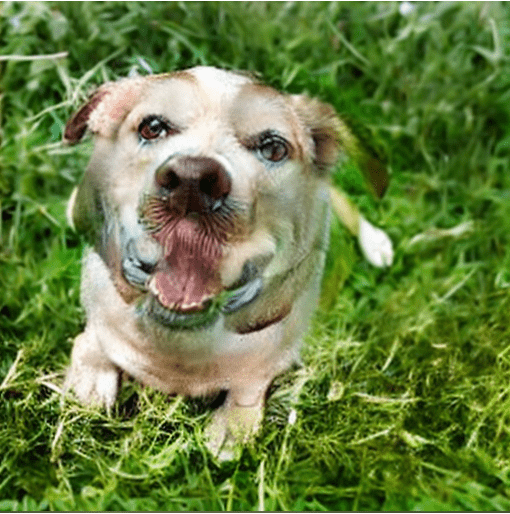}&
    \includegraphics[width=.19\linewidth]{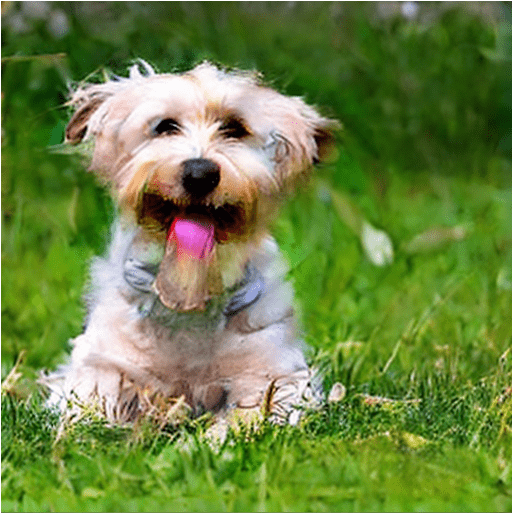}&
    \includegraphics[width=.19\linewidth]{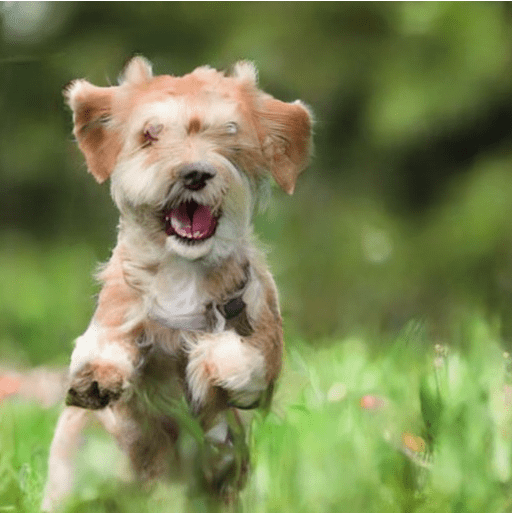}&
    \includegraphics[width=.19\linewidth]{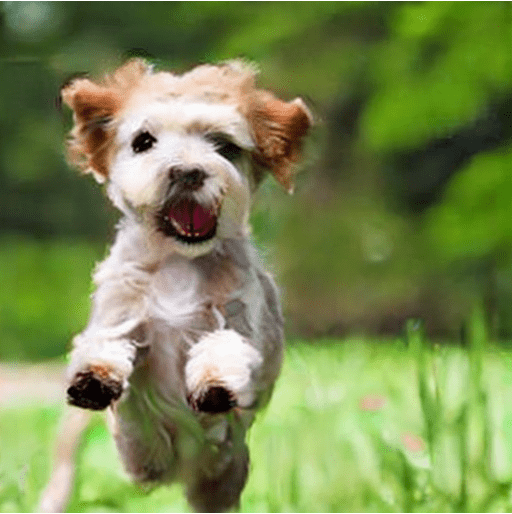}\\
    \raisebox{0.22in}{\rotatebox{90}{ASD}}
    \includegraphics[width=.19\linewidth]{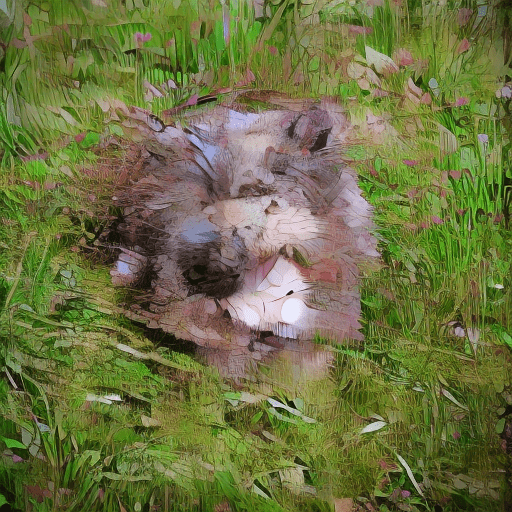}&
    \includegraphics[width=.19\linewidth]{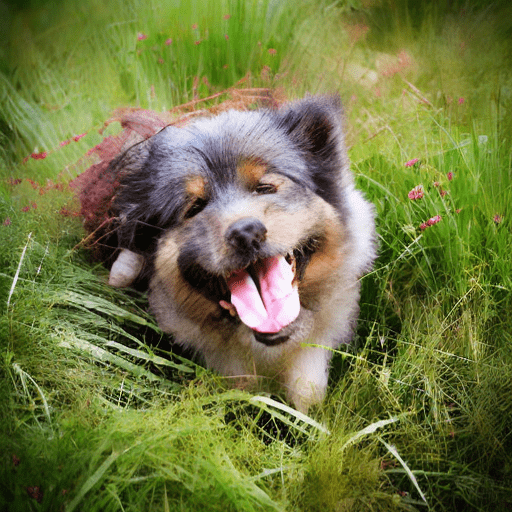}&
    \includegraphics[width=.19\linewidth]{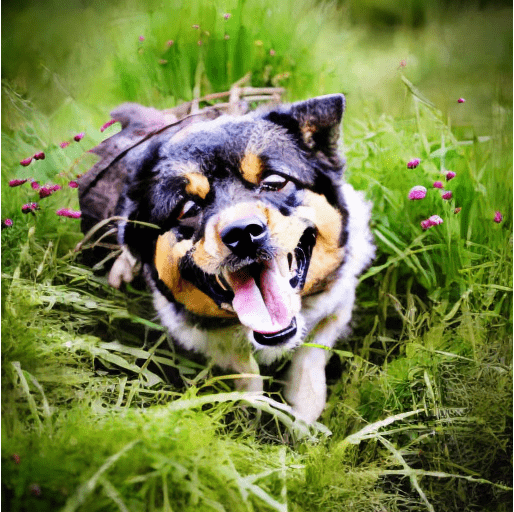}&
    \includegraphics[width=.19\linewidth]{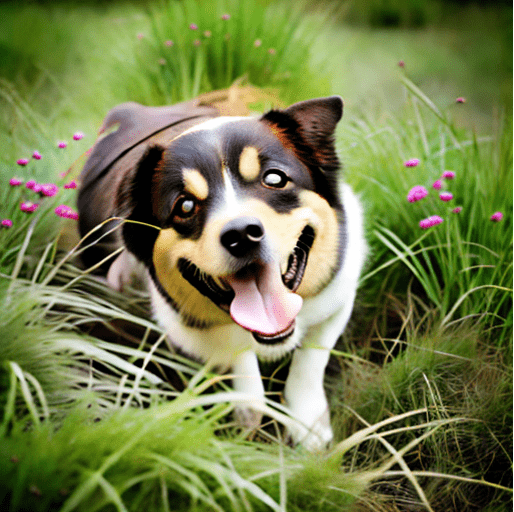}&
    \includegraphics[width=.19\linewidth]{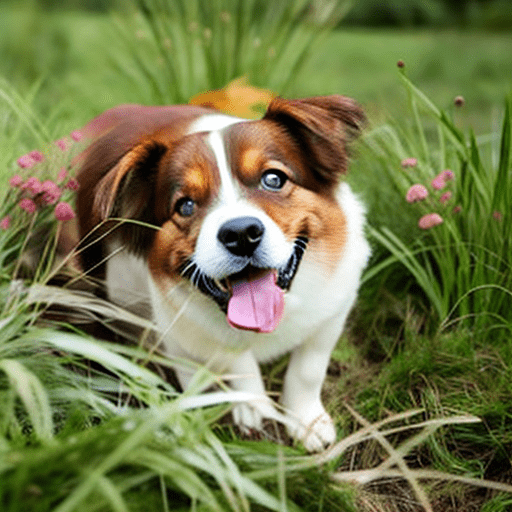}\\
    \multicolumn{5}{c}{\textit{a realistic happy dog playing in the grass}}\\
    \raisebox{0.2in}{\rotatebox{90}{VSD}}
    \includegraphics[width=.19\linewidth]{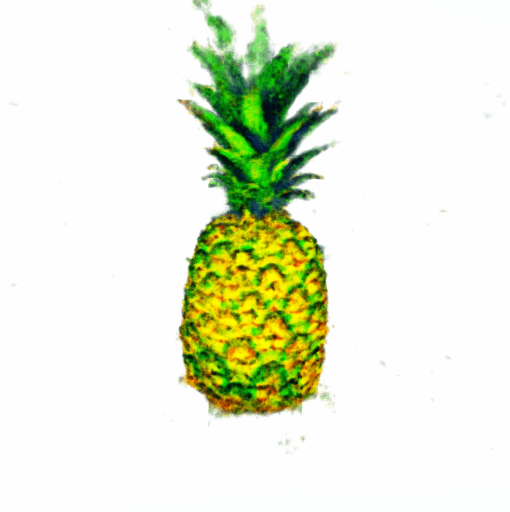}&
    \includegraphics[width=.19\linewidth]{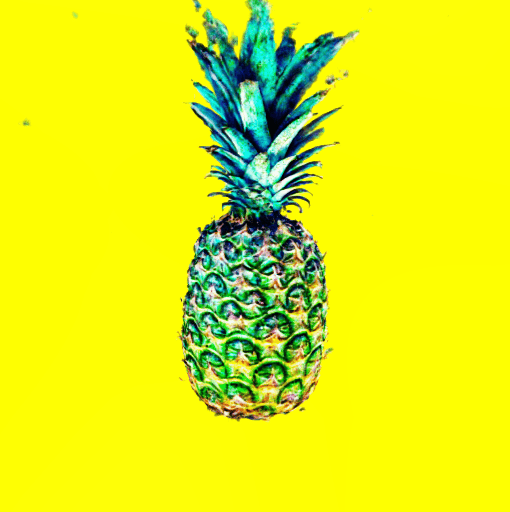}&
    \includegraphics[width=.19\linewidth]{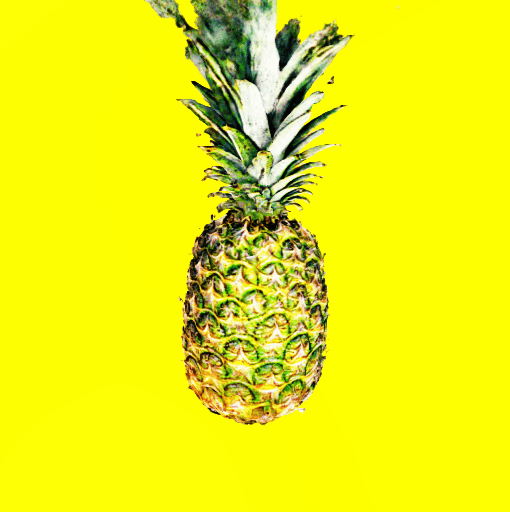}&
    \includegraphics[width=.19\linewidth]{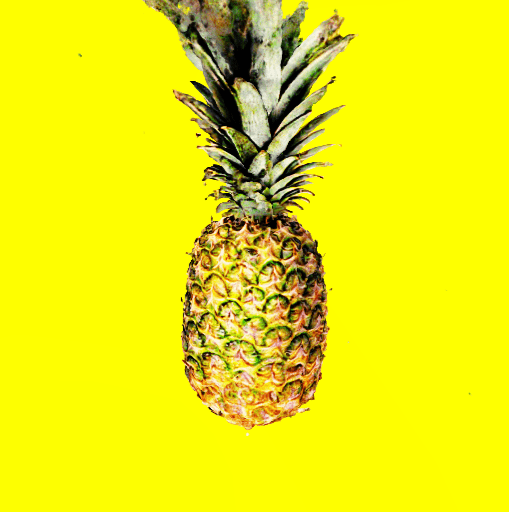}&
    \includegraphics[width=.19\linewidth]{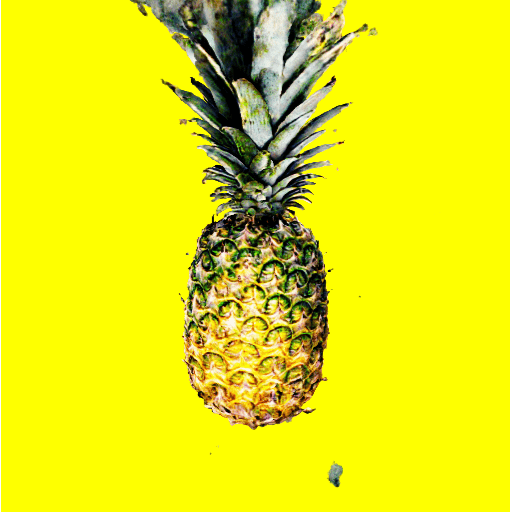}\\
    \raisebox{0.22in}{\rotatebox{90}{ASD}}
    \includegraphics[width=.19\linewidth]{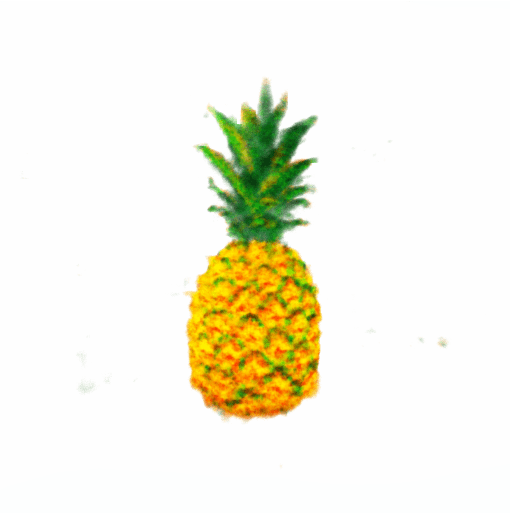}&
    \includegraphics[width=.19\linewidth]{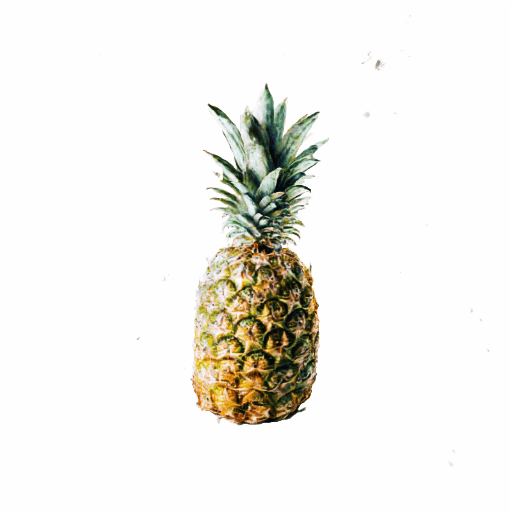}&
    \includegraphics[width=.19\linewidth]{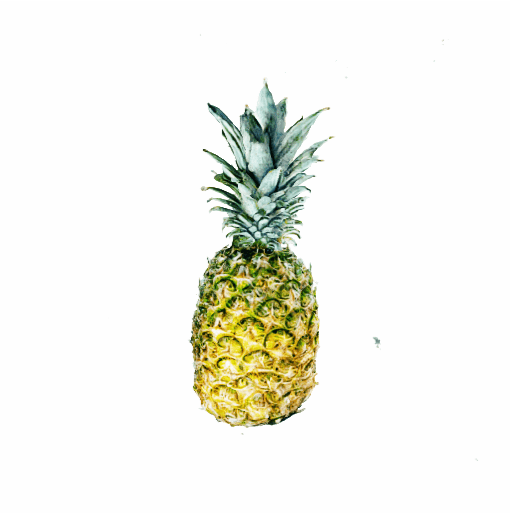}&
    \includegraphics[width=.19\linewidth]{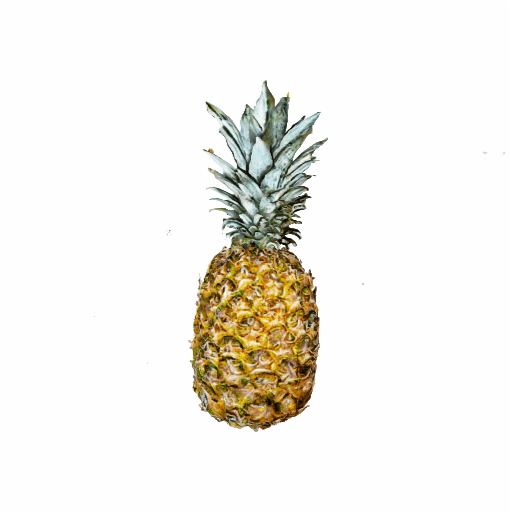}&
    \includegraphics[width=.19\linewidth]{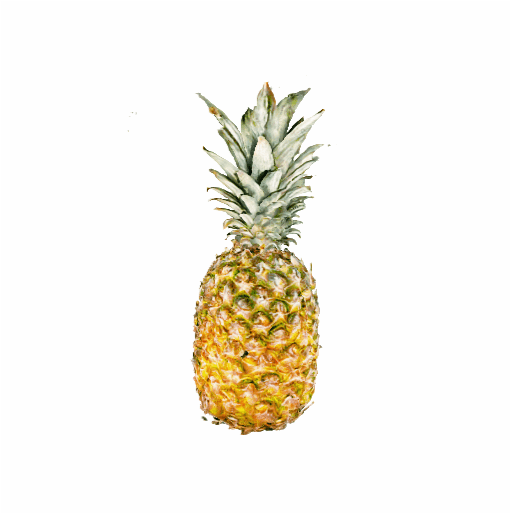}\\
    \multicolumn{5}{c}{\textit{a pineapple}}
    \end{tabular}
\caption{Stability comparisons of the distillation progress between VSD and SDS in both 2D and 3D. As can be seen, ASD constantly keeps the main object and image / NeRF structures unchanged.}
\label{fig:stability}
\vspace{-1mm}
\end{figure}

\section{Experiments}

\vspace{1mm}
\noindent{\textbf{Implementation Details.}}
We choose ProlificDreamer-2D~\cite{prolific_dreamer2d} as the code base for 2D score distillation, Threestiudio~\cite{threestudio2023} as the code base for the text-to-3D task, and DDS~\cite{dds} as the code base for image editing. We employ Stable Diffusion v2-1-base~\cite{sd_2_1_base} as the pretrained diffusion model except for the text-to-3D task, where the LoRA branch is initialized from Stable Diffusion v2-1~\cite{sd_2_1,salimans2022progressive}, following the configuration of VSD~\cite{vsd} in Threestudio.

We set $\lambda$ in Eq.~\eqref{eq:grad_g} to 7.5 by default and use Eq.~\eqref{eq:loss_d_text} for discriminator optimization in both 2D score distillation and text-to-3D tasks. We set $\gamma$ in Eq.~\eqref{eq:loss_d_text} to -1 for 2D score distillation, and set to -0.5 for the text-to-3D task. Although here we use fixed $\gamma$ for simplicity, we think that a dynamic $\gamma$ based on prompts and optimization steps may yield better results. In the text-to-3D task, all other configurations are inherited from the default configuration of VSD in Threestudio~\cite{threestudio2023}. 

\begin{figure}[t]
    \centering 
    \small
    \addtolength{\tabcolsep}{-5.5pt}
    \begin{tabular}{cccc}
    \includegraphics[width=.245\linewidth]{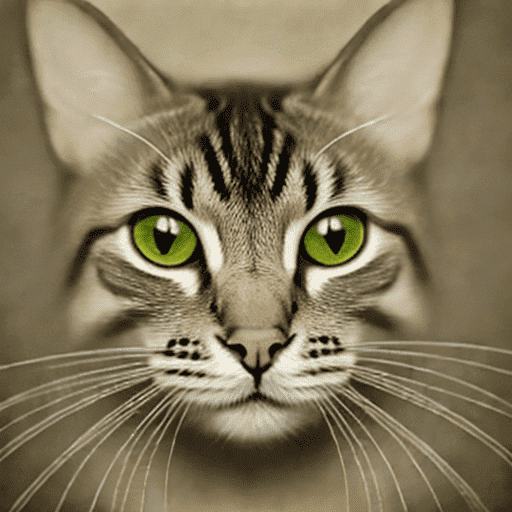}&
    \includegraphics[width=.245\linewidth]{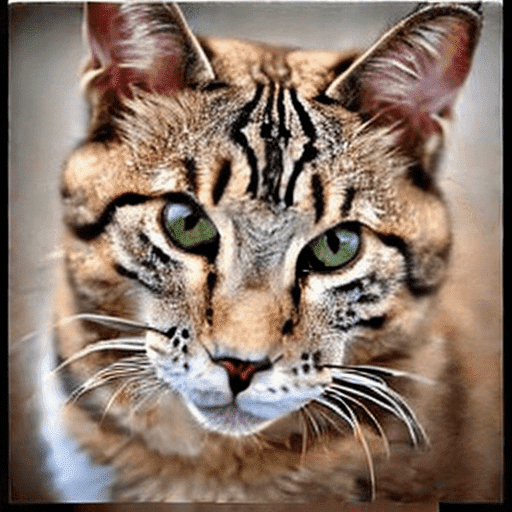}&
    \includegraphics[width=.245\linewidth]{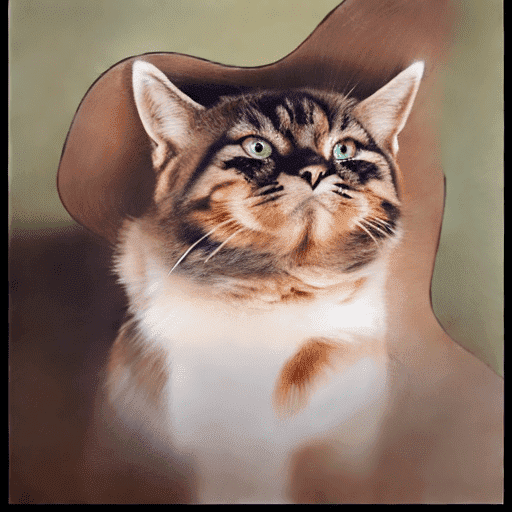}&
    \includegraphics[width=.245\linewidth]{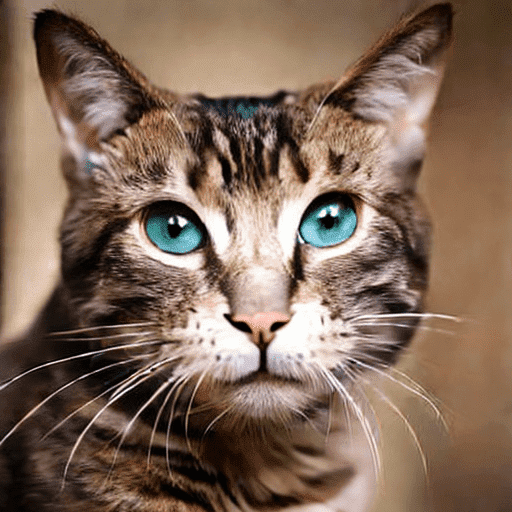}\\
    \multicolumn{4}{c}{\textit{a photograph of a lovely cat}}\\
    \includegraphics[width=.245\linewidth]{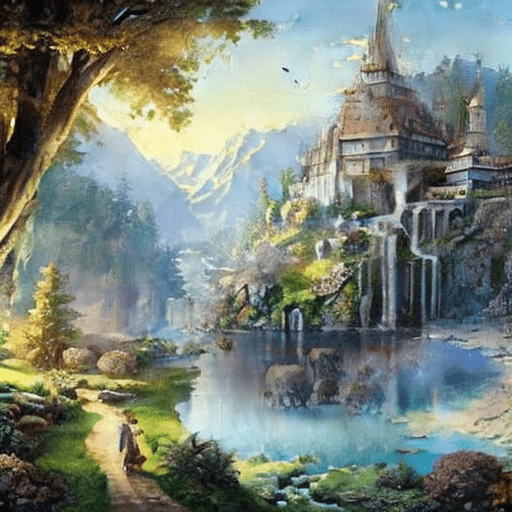}&
    \includegraphics[width=.245\linewidth]{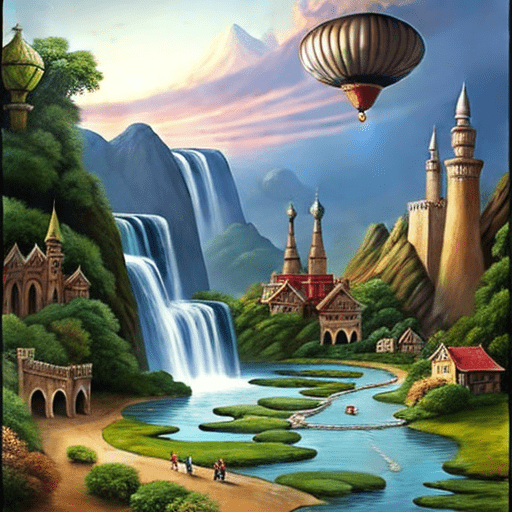}&
    \includegraphics[width=.245\linewidth]{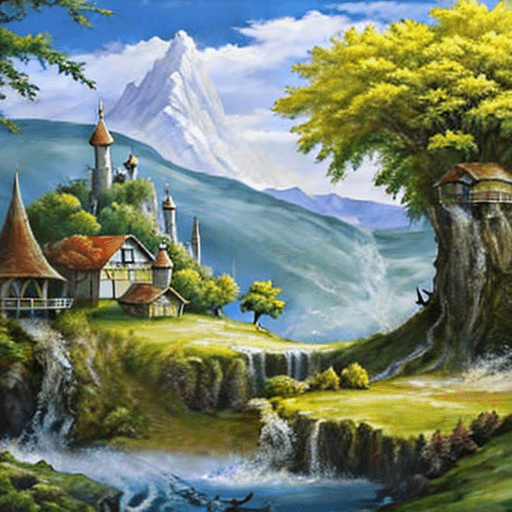}&
    \includegraphics[width=.245\linewidth]{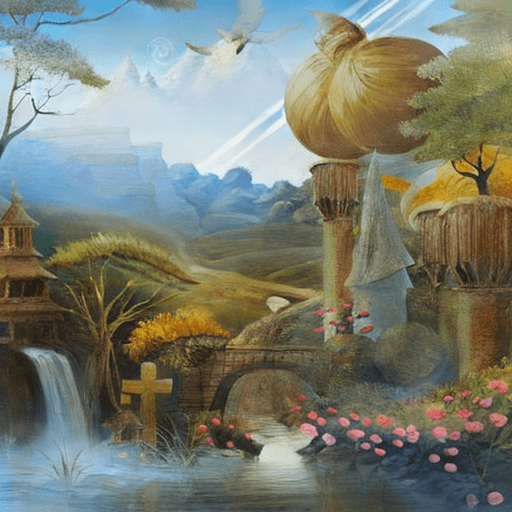}\\
    \multicolumn{4}{c}{\textit{a beautiful painting of fantasy land}}\\
    \includegraphics[width=.245\linewidth]{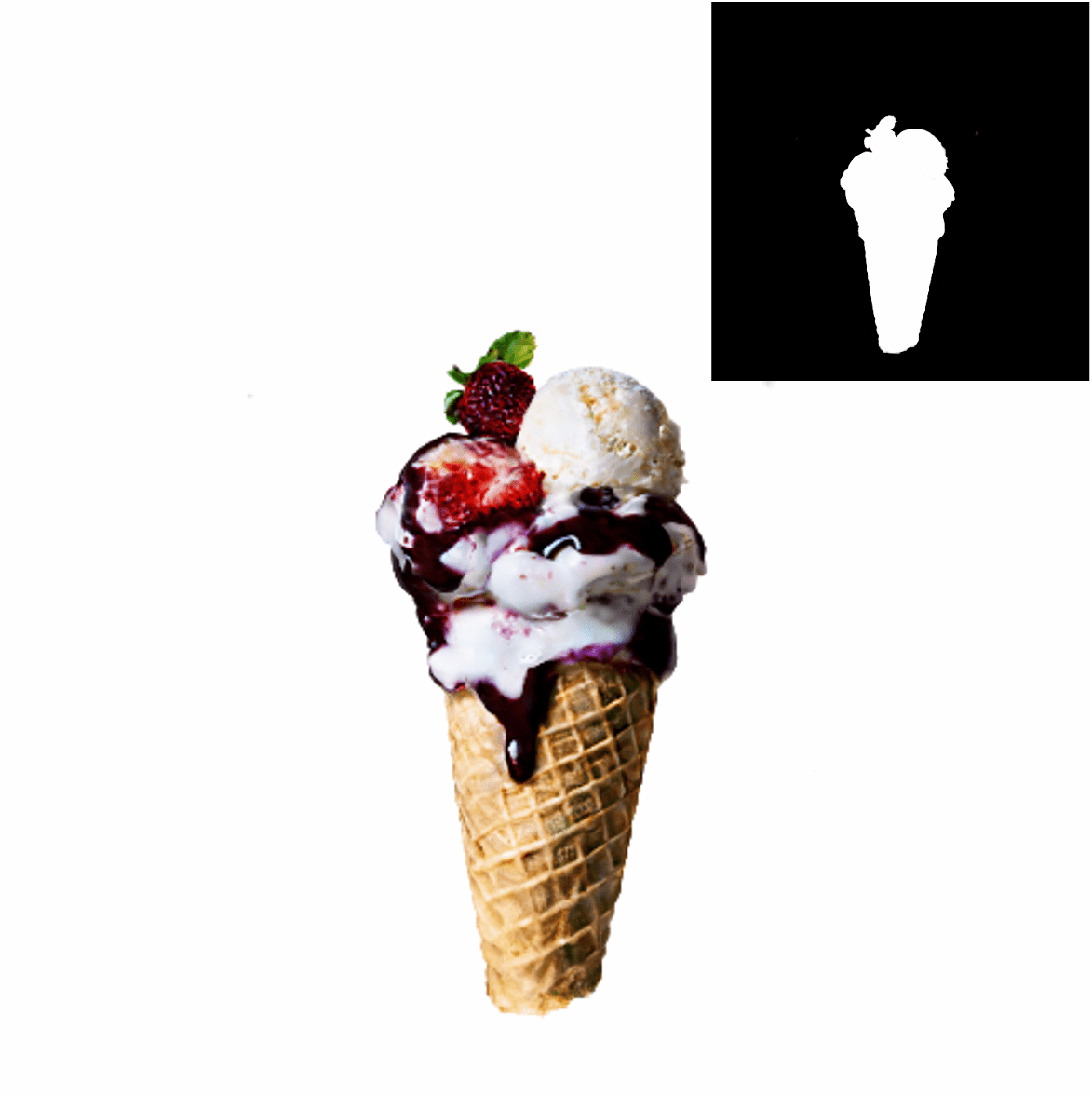}&
    \includegraphics[width=.245\linewidth]{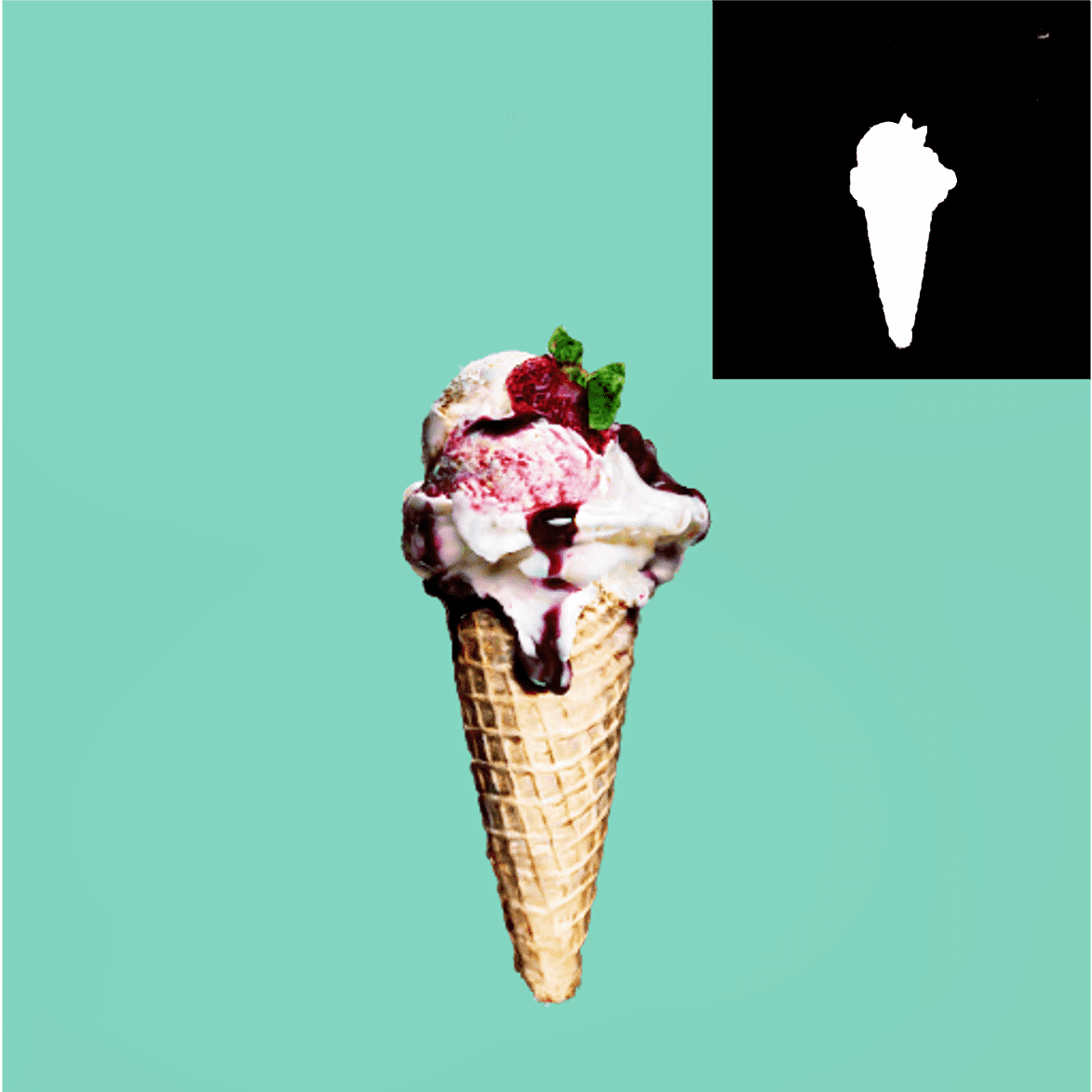}&
    \includegraphics[width=.245\linewidth]{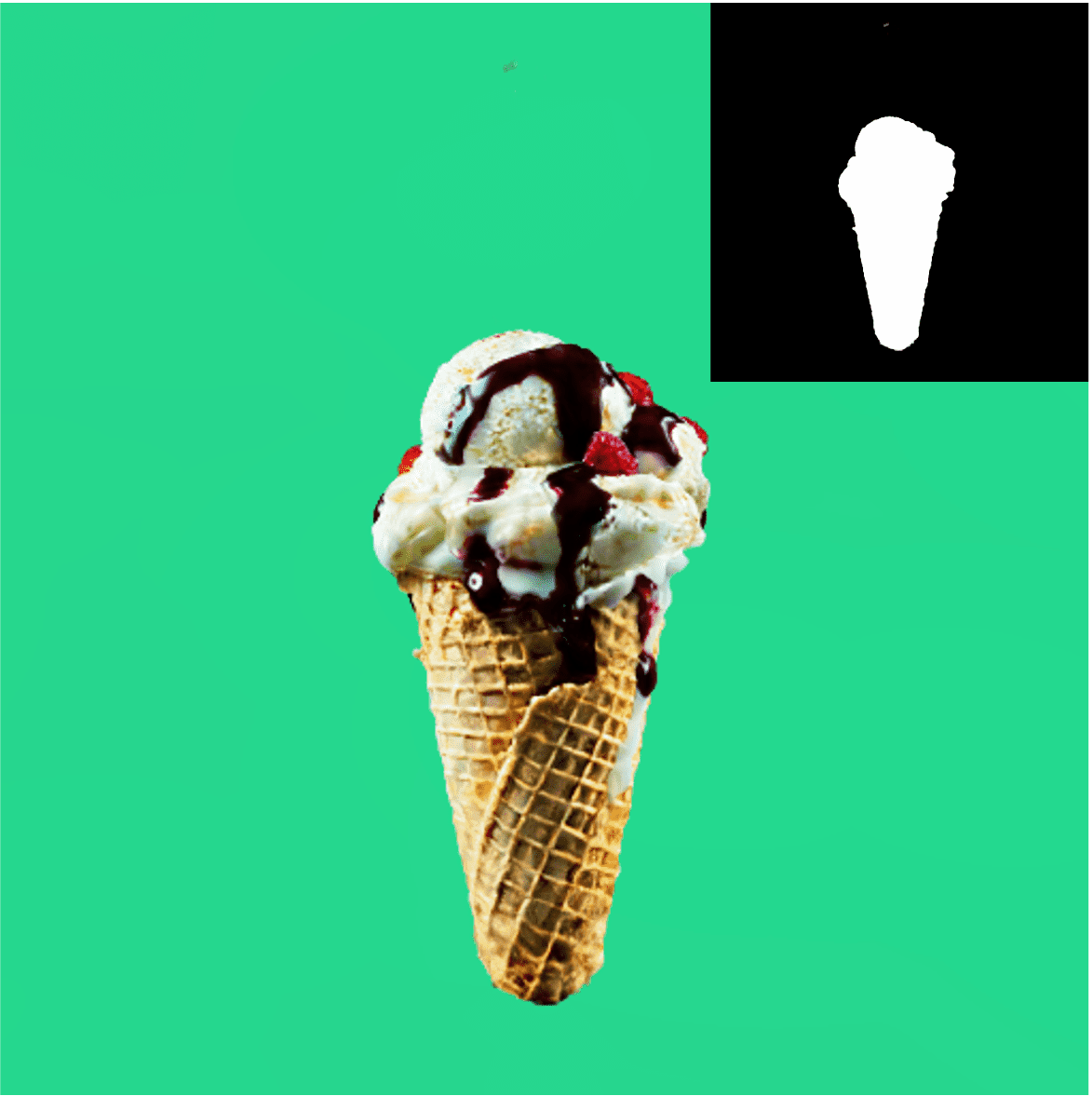}&
    \includegraphics[width=.245\linewidth]{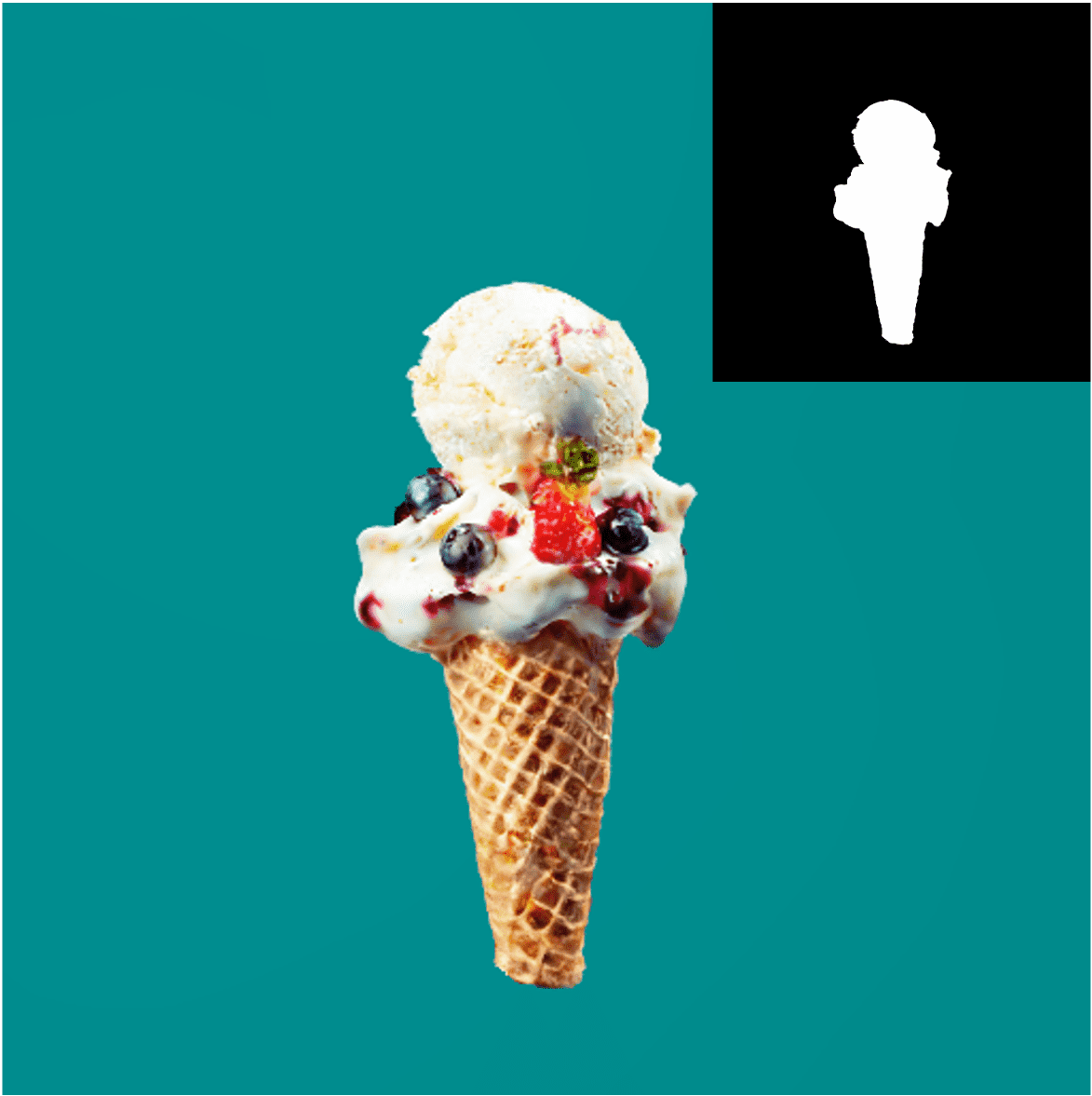}\\
    \multicolumn{4}{c}{\textit{an ice cream sundae}}\\
    \end{tabular}
\caption{Diversity evaluation. ASD can easily distill different 2D and 3D results by switching the random seed.}
\label{fig:diversity}
\vspace{-1mm}
\end{figure}

For image editing, we feed the source image into the second discriminator $D(x_t; z)$ following DDS~\cite{dds}. In such case, we use Eq.~\eqref{eq:loss_d_text} to optimize the first discriminator $D(x_t; y)$ and employ Eq.~\eqref{eq:grad_d} to optimize the second discriminator $D(x_t; z)$, where $\gamma$ in Eq.~\eqref{eq:loss_d_text} is set to -1, $\eta$ and $\gamma$ in Eq.~\eqref{eq:grad_d} are set to 0 and 1, respectively. The source image of image editing is generated from Stable Diffusion v2-1~\cite{sd_2_1}.

\begin{figure*}[t]
    \small
    \centering
    \addtolength{\tabcolsep}{-5.5pt}
    \begin{tabular}{ccccccccc}
    \raisebox{0.22in}{\rotatebox{90}{DDS}}
    \includegraphics[width=.103\linewidth]{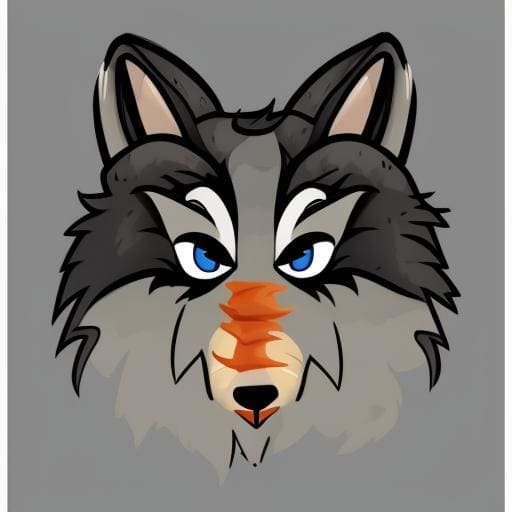}&
    \includegraphics[width=.103\linewidth]{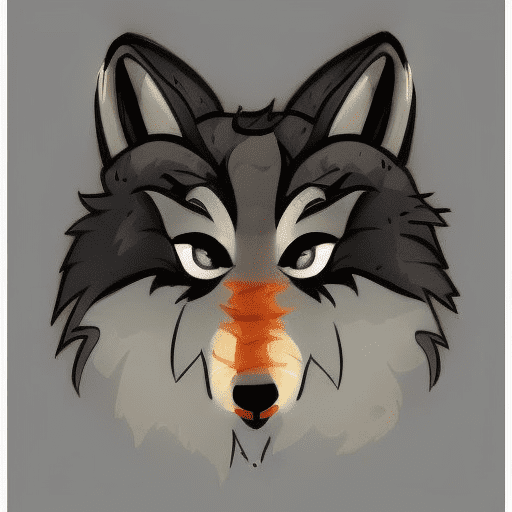}&
    \includegraphics[width=.103\linewidth]{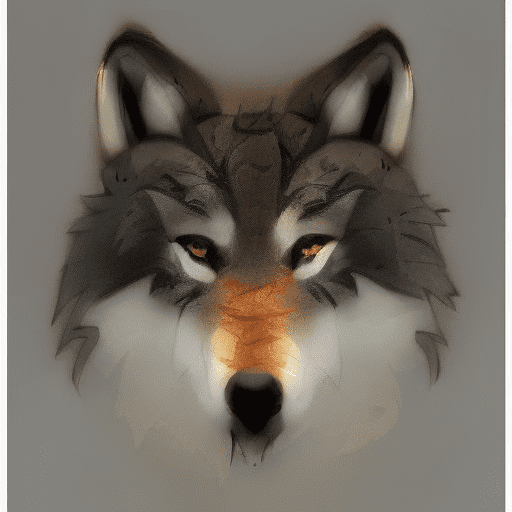}&

    \includegraphics[width=.103\linewidth]{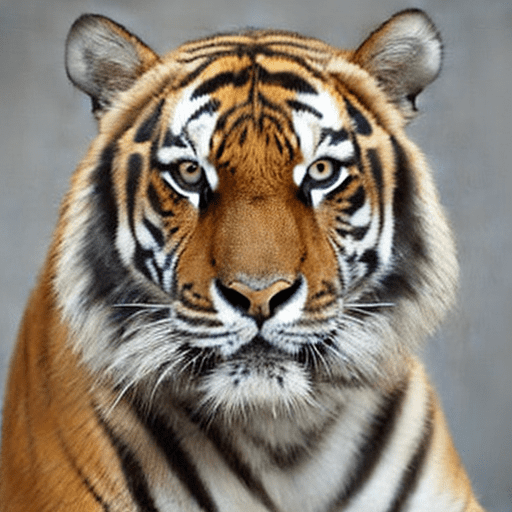}&
    \includegraphics[width=.103\linewidth]{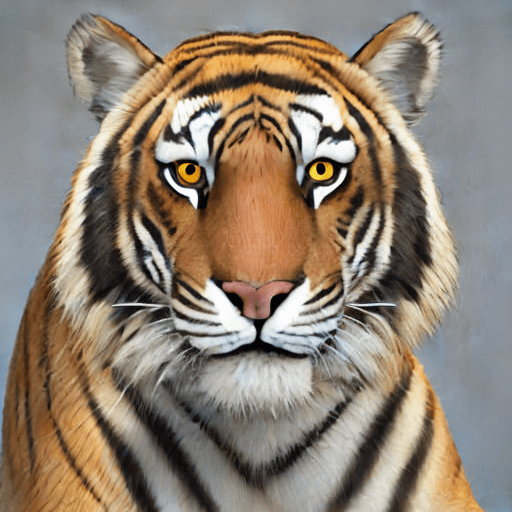}&
    \includegraphics[width=.103\linewidth]{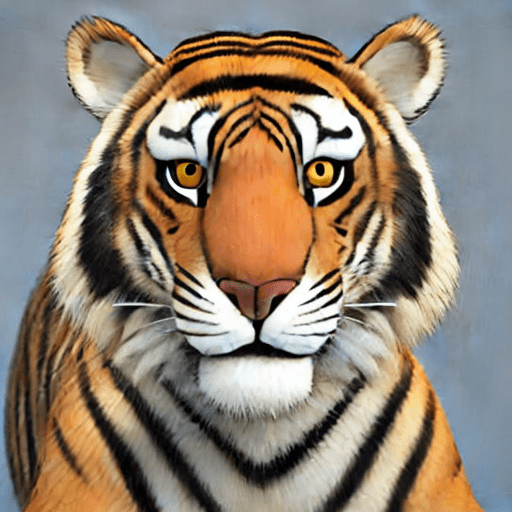}&

    \includegraphics[width=.103\linewidth]{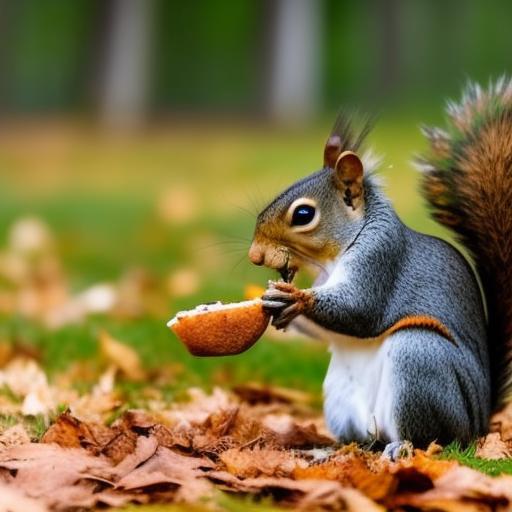}&
    \includegraphics[width=.103\linewidth]{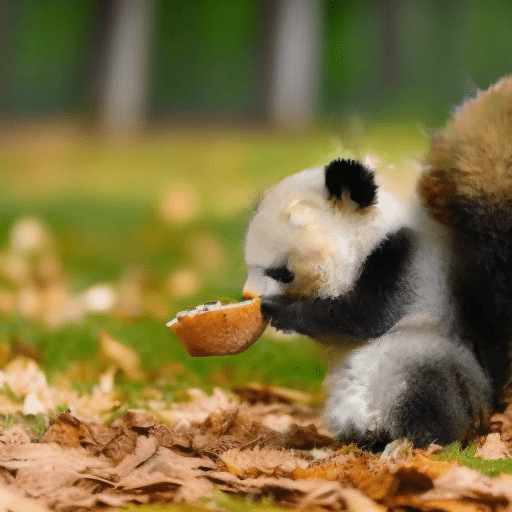}&
    \includegraphics[width=.103\linewidth]{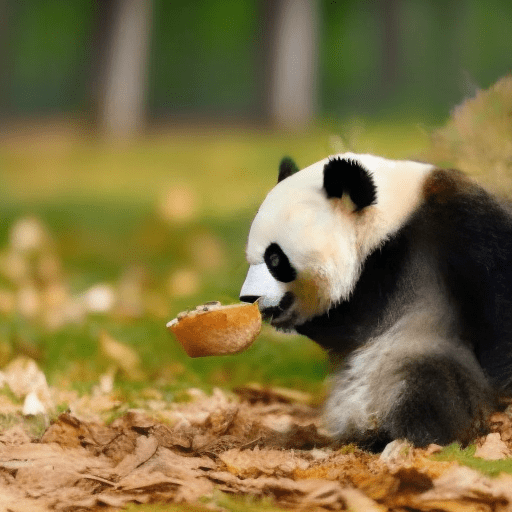}\\

    \raisebox{0.22in}{\rotatebox{90}{ASD}}
    \includegraphics[width=.103\linewidth]{figures/wolf.jpg}&
    \includegraphics[width=.103\linewidth]{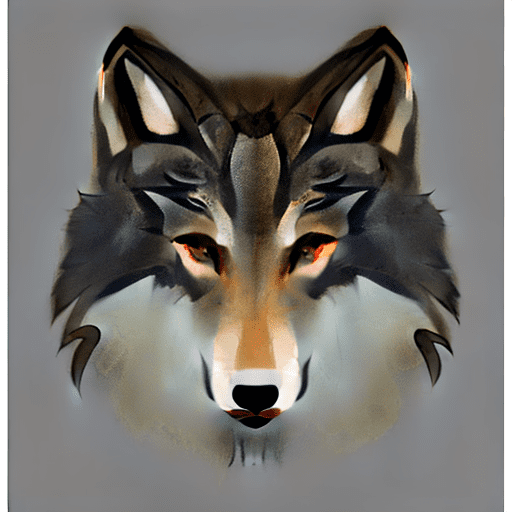}&
    \includegraphics[width=.103\linewidth]{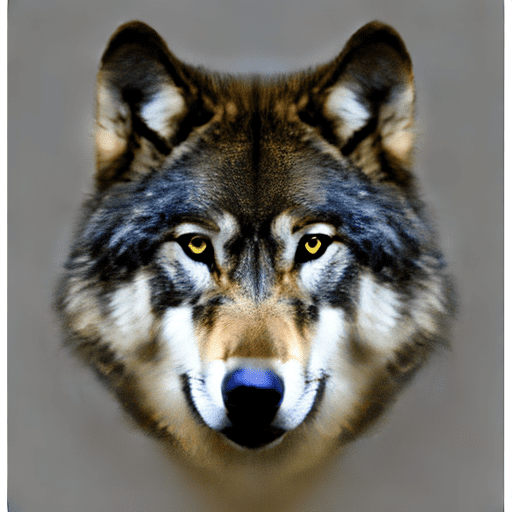}&

    \includegraphics[width=.103\linewidth]{figures/500.png}&
    \includegraphics[width=.103\linewidth]{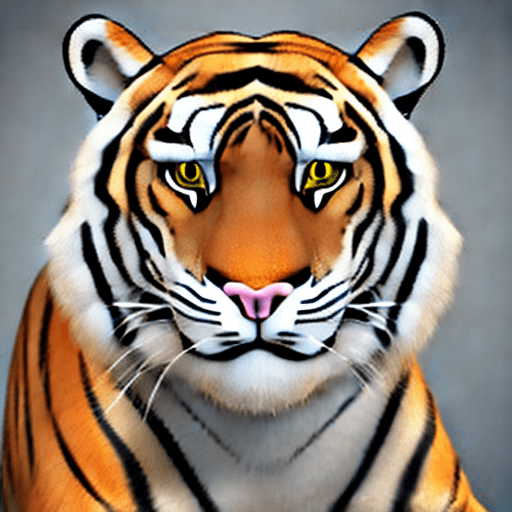}&
    \includegraphics[width=.103\linewidth]{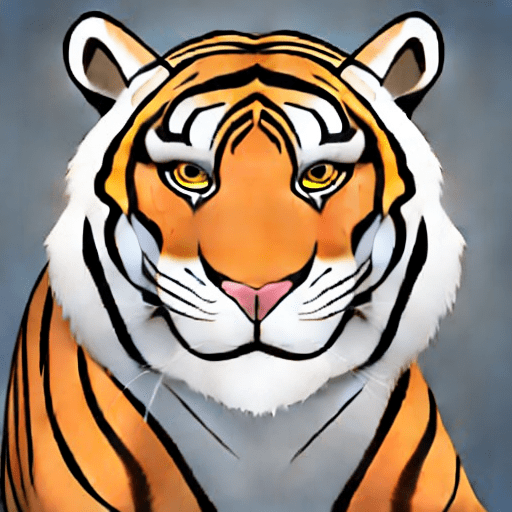}&

    \includegraphics[width=.103\linewidth]{figures/squirrel.jpg}&
    \includegraphics[width=.103\linewidth]{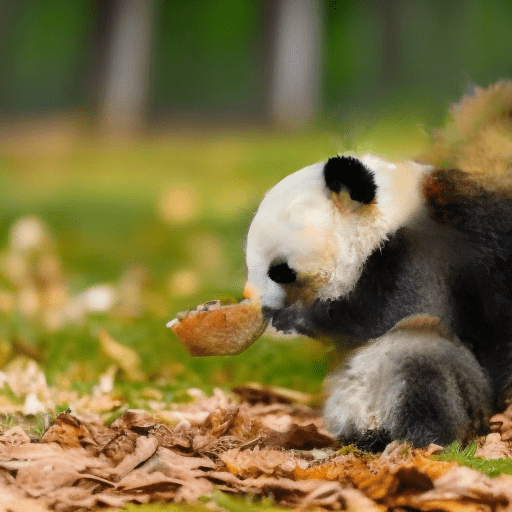}&
    \includegraphics[width=.103\linewidth]{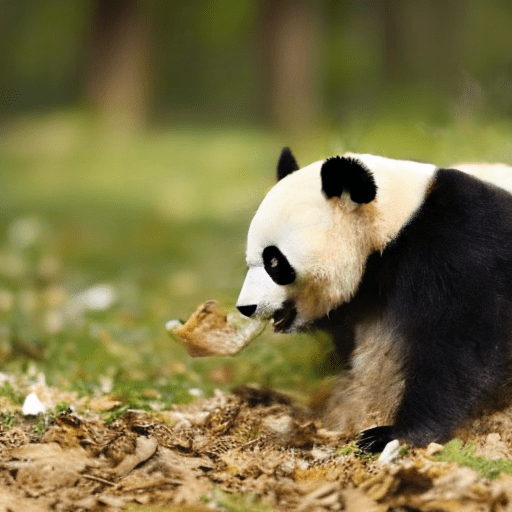}\\

    \multicolumn{3}{c}{\multirow{2}{*}{\textit{a \darkred{cartoon} wolf} $\rightarrow$ \textit{a \darkgreen{\st{cartoon}} wolf}}}&
    \multicolumn{3}{c}{\multirow{2}{*}{\textit{a tiger} $\rightarrow$ \textit{a \darkgreen{cartoon} tiger}}}&
    \multicolumn{3}{c}{\textit{a \darkred{squirrel} is *} $\rightarrow$ \textit{a \darkgreen{panda} is *}} \\
    &&&&&&\multicolumn{3}{c}{* \textit{eating a piece of food on the ground}}\\
    \end{tabular}
    \vspace{-1mm}
    \caption{Comparisons between DDS and ASD on 2D image editing tasks. We test our ASD in three image editing situations: simplification, refinement, and replacement. For each case, we use three images to show the optimization progress.}
    \vspace{-1mm}
    \label{fig:editing}
\end{figure*}

\vspace{1mm}
\noindent \textbf{Quality.} We first compare the distillation quality among SDS~\cite{sds}, VSD~\cite{vsd}, and ASD. Figure~\ref{fig:quality} shows the visual results obtained by different distillation methods in 2D distillation and text-to-3D tasks. In 2D distillation, Figure ~\ref{fig:quality} (a) demonstrates that ASD can consistently achieve better results than SDS and VSD for three types of images: portraits, realistic images, and caricatures. In comparison, VSD results are slightly more saturated (as shown in the bear image) and contain artifacts due to instability in the distillation process (as shown in the flamingo image). Both phenomena are also seen in the text-to-3D task, where the VSD results are slightly more saturated in color and have more structural artifacts. Thanks to the complete discriminator optimization, the proposed ASD can obtain clean 3D NeRF compared to VSD.

\vspace{1mm}
\noindent \textbf{Stability.} We further evaluate the stability of ASD by comparing the distillation process with VSD. Figure~\ref{fig:stability} shows the distillation progress in 2D and 3D tasks. In 2D distillation, ASD always keeps the main objects and image structures unchanged. The gradient of the ASD generator loss is focused on color and details, where the color of the dog gradually changes from black to yellow. However, VSD is not that stable. Both the main object, the image structure, the color of the dog, and other details are constantly changing during the VSD distillation process. We note that this instability has less impact on the text-to-3D task, as occupancy loss and NeRF help maintain structural stability. Despite this, the structures distilled by VSD still change faster than those distilled by ASD, resulting in structural artifacts (see the pineapple leaves). It is worth noting that both ASD and VSD will change objects or structures eventually, due to their adversarial nature. ASD can slow down and stabilize this process, thereby reducing artifacts.

\vspace{1mm}
\noindent \textbf{Diversity.} By switching seeds, we can easily get various distillation results. Figure~\ref{fig:diversity} exhibits several distillation results using the same text prompt but different seeds. It can be seen that ASD can always obtain high-quality results in both 2D and 3D tasks.

\vspace{1mm}
\noindent \textbf{Image Editing.} In addition to typical score distillation tasks, we further extend ASD to zero-short image editing. Figure~\ref{fig:editing} shows how ASD simplifies the source image, refines the source image, and replaces objects in the source image. For the simplification tasks, our ASD can remove the most unnecessary high-frequency details and retain the main content based on the target prompt. ASD can also add realistic details back to cartoon-style images, which is significantly better than the recent zero-shot image editing method DDS~\cite{dds}. We also challenged the object replacement task. Our ASD method produces more reasonable, detailed, and realistic results based on semantic information instead of the rigid fashion of copying fixed objects and then pasting them at fixed locations. See Figure~\ref{fig:editing}, the squirrel's tail still appears in the results of DDS. Besides reasonable object replacement, ASD can maintain the details of the source image as well.

\vspace{1mm}
\noindent \textbf{Quantitative Results.} Table~\ref{tbl:quantitative_results} presents quantitative evaluations of CLIP scores and user studies for 3D generation tasks, demonstrating that ASD performs favorably against SDS and VSD. See the supplementary for more discussion about CLIP scores.

\begin{table}
\centering
\scalebox{0.77}{
    \begin{tabular}{lccc}
    \hline
    \multirow{2}{*}{Method} &\multicolumn{3}{c}{CLIP Score} \\
    \cline{2-4}
    &CLIP B/32$\uparrow$ & CLIP B/16$\uparrow$ & CLIP L/14$\uparrow$\\
    \hline
    DreamFusion (SDS) & 0.3282 & 0.3290 & \textbf{0.2889}\\
    ProlificDreamer (VSD) & 0.3203 & 0.3258 & 0.2789\\
    Ours (ASD) & \textbf{0.3314} &\textbf{ 0.3376} & 0.2854\\
    \hline
    
    \multirow{3}{*}{Method}&\multicolumn{3}{c}{User Study} \\
    \cline{2-4}
    &Details(\%)& {\makecell[c]{Semantic \\Alignment}}{(\%)} & {\makecell[c]{Overall \\Quality}}{(\%)}\\
    \hline
    ProlificDreamer (VSD) & 35.3 & 45.7 & 34.5\\
    Ours (ASD) & \textbf{64.7} & \textbf{54.3} & \textbf{65.5}\\
    \hline
    \end{tabular}}
    \vspace{-3pt}
    \caption{Quantitative Results. CLIP scores are the cosine similarity between multiple randomly rendered views of the 3D model and the given text prompt. User evaluation is conducted between VSD and ASD from three perspectives.}
    \label{tbl:quantitative_results}
    \vspace{-3mm}
\end{table}
\section{Limitation}
Despite showing favorable capabilities in terms of distillation quality, stability, and diversity, the speed of ASD is similar to VSD. We find that sharing the same added noise between generator optimization and discriminator optimization can speed up distillation by 30 $\sim$ 50\%, with a little quality sacrifice. However, distilling a 3D NeRF still takes more than 5 hours on a single A800 GPU.  Using faster 3D representation~\cite{3dgs}, or faster PEFT~\cite{lora,liu2023gpt,liu2022few,edalati2022krona,li2021prefix} methods to update the discriminator, or designing forward-only methods to approximate $\epsilon_{x_t; \phi, t}$ may be promising solutions to this limitation.

\section{Conclusion}
In this paper, we proposed the Adversarial Score Distillation (ASD) under the WGAN paradigm. Compared to existing methods, our ASD exploits a complete discriminator loss with a principled discussion, which alleviates the CFG-sensitive issue and leads to impressive performances. 

More importantly, we elucidated the relationship between score distillation and GAN, which not only explains methodological issues of existing score distillation, but also enables the pretrained model to extend to downstream tasks by various GAN paradigms.

\section{Acknowledgements}
We would like to express our gratitude to all the individuals who provided assistance in this work: Chaohui Yu, Fan Wang. We also thank the authors of all the open-source codes and models that we used.
{
    \small
    \bibliographystyle{ieeenat_fullname}
    \bibliography{main}
}

\appendix
\onecolumn


\setcounter{page}{1}
\section{Quantitative Evaluations and User Studies for 3D Generation}

We quantitatively compare our ASD with DreamFusion~\cite{sds} (SDS) and ProlificDreamer~\cite{vsd} (VSD) using CLIP score, following previous works~\cite{sds,dreamfield}. We evaluate the similarities between text prompts and randomly rendered views of the 3D NeRFs using three variants of CLIP~\cite{clip} (\ie, CLIP B/32, CLIP B/16, and CLIP B/14). The CLIP score may not honestly reflect the quality of 3D results. For example, the CLIP score of Figure~\ref{fig:clip_example}(a) is higher than Figure~\ref{fig:clip_example}(b) even though its quality is significantly worse, which does not align with human perceptual judgments. Several existing methods~\cite{vsd,fantasia3d} choose to only conduct user studies for performance evaluation.

\begin{table}[h]
\centering
    \begin{tabular}{cccc}
    \hline
    Method & CLIP B/32$\uparrow$ & CLIP B/16$\uparrow$ & CLIP L/14$\uparrow$\\
    \hline
    DreamFusion (SDS) & 0.3282 & 0.3290 & 0.2889\\
    ProlificDreamer (VSD) & 0.3203 & 0.3258 & 0.2789\\
    Ours (ASD) & 0.3314 & 0.3376 & 0.2854\\
    \hline
    \end{tabular}
    \caption{CLIP Score by computing the similarity between multiple randomly rendered views of the 3D model and the given text prompt.}
    \vspace{-5mm}
\end{table}

\begin{multicols}{2}
\begin{center}
    \begin{tabular}{cc}
    \includegraphics[width=.47\linewidth]{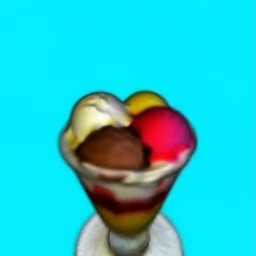}&
    \includegraphics[width=.47\linewidth]{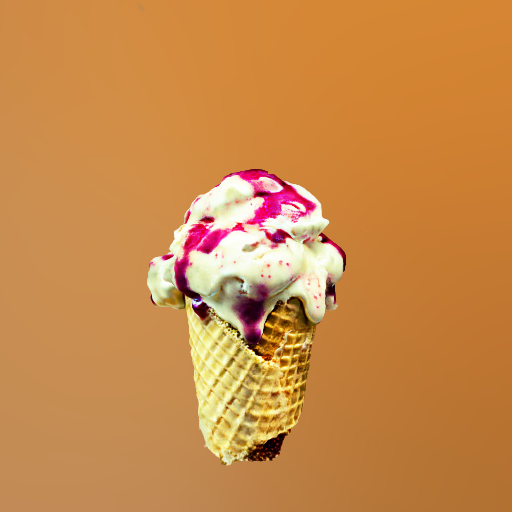}\\
    (a) CLIP score: 0.3401 & (b) CLIP score: 0.3069
    \end{tabular}
\caption{Illustration of CLIP scores that conflict with human perceptual judgments. We use CLIP B/16 to compute scores here.}
\label{fig:clip_example}
\end{center}

\begin{center}
    \centerline{\vspace{-2mm}\includegraphics[width=.9\linewidth, trim=0 0 0 20, clip]{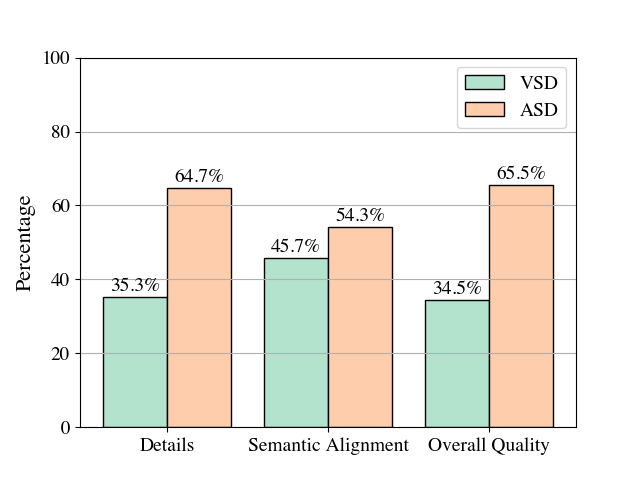}}
\caption{User evaluation between VSD and ASD.}
\vspace{-5mm}
\label{fig:user_study}
\end{center}
\end{multicols}

Therefore, we also present user studies in Figure~\ref{fig:user_study}. We compare 3D NeRFs generated by VSD and ASD with different prompts in terms of details, semantic alignment, and overall quality. Figure~\ref{fig:user_study} demonstrates that ASD leads to better details, which is consistent with our observation that VSD's results are slightly more saturated and therefore have worse details. The semantic alignment scores of VSD and ASD are similar. In general, more users prefer the results generated by ASD.

\section{Derivation of Adversarial Score Distillation Loss and Gradients}
We define a discriminator in the form as
\begin{equation}
    D(x_t;y) = \log\frac{p(y|x_t)}{p(\phi|x_t)}
\end{equation}
Based on WGAN~\cite{wgan,wgan_div}, we have the generator loss as
\begin{equation}
    \mathcal{L}_G = \mathbb{E}_{t,\epsilon}[-D(x_t^g;y)] = \mathbb{E}_{t,\epsilon}[-\log\frac{p(y|x_t)}{p(\phi|x_t)}] \leq \mathbb{E}_{t,\epsilon}[-\log\frac{p(y|x_t)^\lambda}{p(\phi|x_t)}]:=\mathcal{L}_G'
\end{equation}
where $x_t^g = \alpha_t g(\theta, c) + \sigma_t\epsilon$ denotes the generated sample with noise.
According to the Bayesian rule, we can do the transformation $\nabla_{x_t} \text{log}p(y|x_t) = -1/\sigma_t(\epsilon_{x_t; y, t} - \epsilon_{x_t, t})$ similar to CFG~\cite{cfg}. Thus, we have
\begin{align}
    &\nabla_\theta \mathcal{L}_G' = \nabla_\theta \mathbb{E}_{t,\epsilon}[\log p(\phi|x_t^g) - \lambda\log p(y|x_t^g)] \nonumber\\
    &= \mathbb{E}_{t,\epsilon}[\omega(t)( - \Big(\epsilon_{x_t^g,\phi,t} - \epsilon_{x_t^g,t}\Big) + \lambda\Big(\epsilon_{x_t^g,y,t} - \epsilon_{x_t^g,t}\Big))\frac{\partial x_0^g}{\partial \theta}] \label{eq:g_loss_supp}\\
    &= \mathbb{E}_{t,\epsilon}[\omega(t)(  \epsilon_{x_t^g,t}+ \lambda(\epsilon_{x_t^g,y,t} - \epsilon_{x_t^g,t})- \epsilon_{x_t^g,\phi,t} )\frac{\partial x_0^g}{\partial \theta}]\nonumber 
\end{align}
where $\omega(t)$ is a weight based on the time $t$.

Then we have the discriminator loss as
\begin{align}
    &\mathcal{L}_D = \mathbb{E}_{t,\epsilon}[D(x_t^g;y) - D(x_t^r;y) + \tau\lVert \nabla_{\tilde{x}_t} D(\tilde{x}_t;y) \rVert_2^2] \nonumber \\
    &= \mathbb{E}_{t,\epsilon}[\log\frac{p(y|x_t^g)}{p(\phi|x_t^g)} - \log\frac{p(y|x_t^r)}{p(\phi|x_t^r)} + \tau\lVert \nabla_{\tilde{x}_t} \log\frac{p(y|\tilde{x}_t)}{p(\phi|\tilde{x}_t)} \rVert_2^2] \label{eq:d_loss_supp}
\end{align}
where $x_t^r = \alpha_t x_0^r + \sigma_t \epsilon$ is the noisy real sample based on the real sample $x_0^r$ and the time $t$. In Eq.~\eqref{eq:d_loss_supp}, $p(y|x_t^r)$ and
$p(y|x_t^g)$ are based on the given pretrained diffusion model, so 
they are not optimizable. $\phi$ in $p(\phi|x_t^g)$ and $p(\phi|x_t^r)$ is optimizable, which can be implemented using an textual-inversion embedding~\cite{text_inervsion} or LoRA~\cite{lora}. Thus, $\mathcal{L}_D$ is a loss function only about parameters $\phi$, its gradient can be derived as
\begin{align}
    &\nabla_\phi \mathcal{L}_D = \nabla_\phi \mathbb{E}_{t,\epsilon}[\log p(\phi|x_t^r) - \log p(\phi|x_t^g) + \tau\lVert \nabla_{\tilde{x}_t} \log\frac{p(y|\tilde{x}_t)}{p(\phi|\tilde{x}_t)} \rVert_2^2] \nonumber \\
    &= \nabla_\phi \mathbb{E}_{t,\epsilon}[\Big(\log p(x_t^r|\phi) + \log p(\phi) -\log p(x_t^r)\Big) - \Big(\log p(x_t^g|\phi)  + \log p(\phi) - \log p(x_t^g)\Big) + \tau\lVert \nabla_{\tilde{x}_t} \log\frac{p(y|\tilde{x}_t)}{p(\phi|\tilde{x}_t)} \rVert_2^2] \nonumber \\
    &= \nabla_\phi \mathbb{E}_{t,\epsilon}[\log \frac{p(x_t^r|\phi)}{p(x_t^g|\phi)} + \tau\lVert \nabla_{\tilde{x}_t} (\Big(\log p(\tilde{x}_t|y) + \log p(y) - \log p(\tilde{x}_t)\Big) - \Big(\log p(\tilde{x}_t|\phi) + \log p(\phi) - \log p(\tilde{x}_t)\Big))\rVert_2^2] \nonumber \\
    &= \nabla_\phi \mathbb{E}_{t,\epsilon}[\log p(x_t^r|\phi) - \log p(x_t^g|\phi) + \tau\lVert \nabla_{\tilde{x}_t}\log p(\tilde{x}_t|y) - \nabla_{\tilde{x}_t}\log p(\tilde{x}_t|\phi)\rVert_2^2]\nonumber\\
    &\approx  \nabla_\phi \mathbb{E}_{t,\epsilon}[\lVert\epsilon_{x_t^g,\phi,t} - \epsilon \rVert_2^2 - \lVert\epsilon_{x_t^r,\phi,t} - \epsilon \rVert_2^2 + \eta \lVert\epsilon_{x_t^r,y,t} - \epsilon_{x_t^r,\phi,t} \rVert_2^2 + \gamma \lVert\epsilon_{x_t^g,y,t} - \epsilon_{x_t^g,\phi,t}\rVert_2^2] \label{eq:loss_supp}
\end{align}
where we use $-\log p(x_0) \approx \mathbb{E}_{t,\epsilon}[\lVert\epsilon_{x_t,\phi,t} - \epsilon \rVert_2^2]$, according to~\cite{ddpm, vae}. $\tilde{x}$ comes from a mixture distribution $\tilde{\mu}$, so we can sampling both real or fake data points~\cite{wgan_div} which means $\tilde{x}_t$ can be replaced by a combination of $x_t^r$ and $x_t^g$. $\eta$ and $\gamma$ are adjustable hyperparameters, to keep the penalty term positive, they are subject to $\gamma \geq - \eta \lVert \epsilon_{x_t^r; y, t} - \epsilon_{x_t^r;\phi, t}\rVert_2^2 /  \lVert \epsilon_{x_t^g; y, t} - \epsilon_{x_t^g;\phi, t}\rVert_2^2$. 
However, in Eq.~\eqref{eq:loss_supp}, the noisy real sample $x_t^r$ is unavailable for text-conditioned cases. For these cases, we can use the upper bound of $\mathcal{L}_D$, (\textit{i.e.} $\mathcal{L}'_D$), which does not contain $x_t^r$ terms. When $\eta = 1/2$, based on triangle and Cauchy–Schwarz inequality, we have:
\begin{align}
     &\nabla_\phi \mathcal{L}'_D = \nabla_\phi \mathbb{E}_{t,\epsilon}[\lVert\epsilon_{x_t^g,\phi,t} - \epsilon \rVert_2^2 + \gamma \lVert\epsilon_{x_t^g,y,t} - \epsilon_{x_t^g,\phi,t}\rVert_2^2] \nonumber \\
     &= \nabla_\phi \mathbb{E}_{t,\epsilon}[\lVert\epsilon_{x_t^g,\phi,t} - \epsilon \rVert_2^2 - \lVert\epsilon_{x_t^r,\phi,t} - \epsilon \rVert_2^2 + (\lVert\epsilon_{x_t^r,\phi,t} - \epsilon \rVert_2^2 + \lVert\epsilon_{x_t^r,y,t} - \epsilon \rVert_2^2) + \gamma \lVert\epsilon_{x_t^g,y,t} - \epsilon_{x_t^g,\phi,t}\rVert_2^2] \nonumber \\
     &= \nabla_\phi \mathbb{E}_{t,\epsilon}[\lVert\epsilon_{x_t^g,\phi,t} - \epsilon \rVert_2^2 - \lVert\epsilon_{x_t^r,\phi,t} - \epsilon \rVert_2^2 + \frac{1}{2}(1^2 + 1^2)(\lVert\epsilon_{x_t^r,\phi,t} - \epsilon \rVert_2^2 + \lVert\epsilon_{x_t^r,y,t} - \epsilon \rVert_2^2) + \gamma \lVert\epsilon_{x_t^g,y,t} - \epsilon_{x_t^g,\phi,t}\rVert_2^2] \nonumber \\ 
     &\geq \nabla_\phi \mathbb{E}_{t,\epsilon}[\lVert\epsilon_{x_t^g,\phi,t} - \epsilon \rVert_2^2 - \lVert\epsilon_{x_t^r,\phi,t} - \epsilon \rVert_2^2 + \frac{1}{2}(\lVert\epsilon_{x_t^r,\phi,t} - \epsilon \rVert_2 + \lVert\epsilon_{x_t^r,y,t} - \epsilon \rVert_2)^2 + \gamma \lVert\epsilon_{x_t^g,y,t} - \epsilon_{x_t^g,\phi,t}\rVert_2^2] \nonumber \\
      &\geq \nabla_\phi \mathbb{E}_{t,\epsilon}[\lVert\epsilon_{x_t^g,\phi,t} - \epsilon \rVert_2^2 - \lVert\epsilon_{x_t^r,\phi,t} - \epsilon \rVert_2^2 + \frac{1}{2}(\lVert\epsilon_{x_t^r,\phi,t} - \epsilon + \epsilon - \epsilon_{x_t^r,y,t}\rVert_2)^2 + \gamma \lVert\epsilon_{x_t^g,y,t} - \epsilon_{x_t^g,\phi,t}\rVert_2^2] \nonumber \\
      &= \nabla_\phi \mathbb{E}_{t,\epsilon}[\lVert\epsilon_{x_t^g,\phi,t} - \epsilon \rVert_2^2 - \lVert\epsilon_{x_t^r,\phi,t} - \epsilon \rVert_2^2 + \frac{1}{2} \lVert\epsilon_{x_t^r,y,t} - \epsilon_{x_t^r,\phi,t} \rVert_2^2 + \gamma \lVert\epsilon_{x_t^g,y,t} - \epsilon_{x_t^g,\phi,t}\rVert_2^2] = \nabla_\phi \mathcal{L}_D
\end{align}
In practice, we find $\eta = 1/2$, $\gamma \in [-1, 0)$ works for most cases.

\section{Explanation of the workflow}
To get a clearer picture of the optimization workflow, we further provide an algorithm description in Algorithm~\ref{algorithm:1}.

\begin{algorithm}
\caption{Adversarial Score Distillation for 3D tasks}
\textbf{Input:}\text{ Prompt $y$, optimizable textual-inversion embedding or LoRA $\phi$, and pretrained diffusion model.}
  \vspace{-5mm}
  \begin{algorithmic}[1]
  \REPEAT
  \STATE Sample a camera pose $c$.
  \STATE Differentiable render the 3D structure $\theta$ at pose $c$ to get a 2D image $x_0 = g(\theta,c)$. 
  \STATE Random select a timestep $t$ and add random noise $\epsilon$ to get $x_t^g = \alpha_t x_0 + \sigma_t\epsilon$.
  \STATE Predict noise $\epsilon_{x_t^g; \phi, t}$, $\epsilon_{x_t^g; y, t}$ and $\epsilon_{x_t^g; t}$ using the pretrained diffusion model.
  \STATE (Update G) Update $\theta$ with the gradient $\nabla_\theta\mathcal{L}_G': \mathbb{E}_{t, \epsilon}[\omega(t)(\epsilon_{x_t^g; t} + \lambda (\epsilon_{x_t^g; y, t} -\epsilon_{x_t^g; t}) - \epsilon_{x_t^g; \phi, t})\frac{\partial x_0^g}{\partial \theta}]$
  \STATE (Update D) Update $\phi$ with the gradient $\nabla_\phi \mathcal{L}_{D}': \nabla_\phi \mathbb{E}_{t,\epsilon}[\lVert\epsilon_{x_t^g; \phi, t} - \epsilon\rVert_2^2 + \gamma\lVert \epsilon_{x_t^g; y, t} - \epsilon_{x_t^g;\phi, t}\rVert_2^2] $
  \UNTIL converged
\end{algorithmic}
\textbf{Output:} 3D structure $\theta$.
\label{algorithm:1}
\end{algorithm}

\begin{figure}
    \centering 
    \begin{tabular}{cc}
    \includegraphics[width=.45\linewidth]{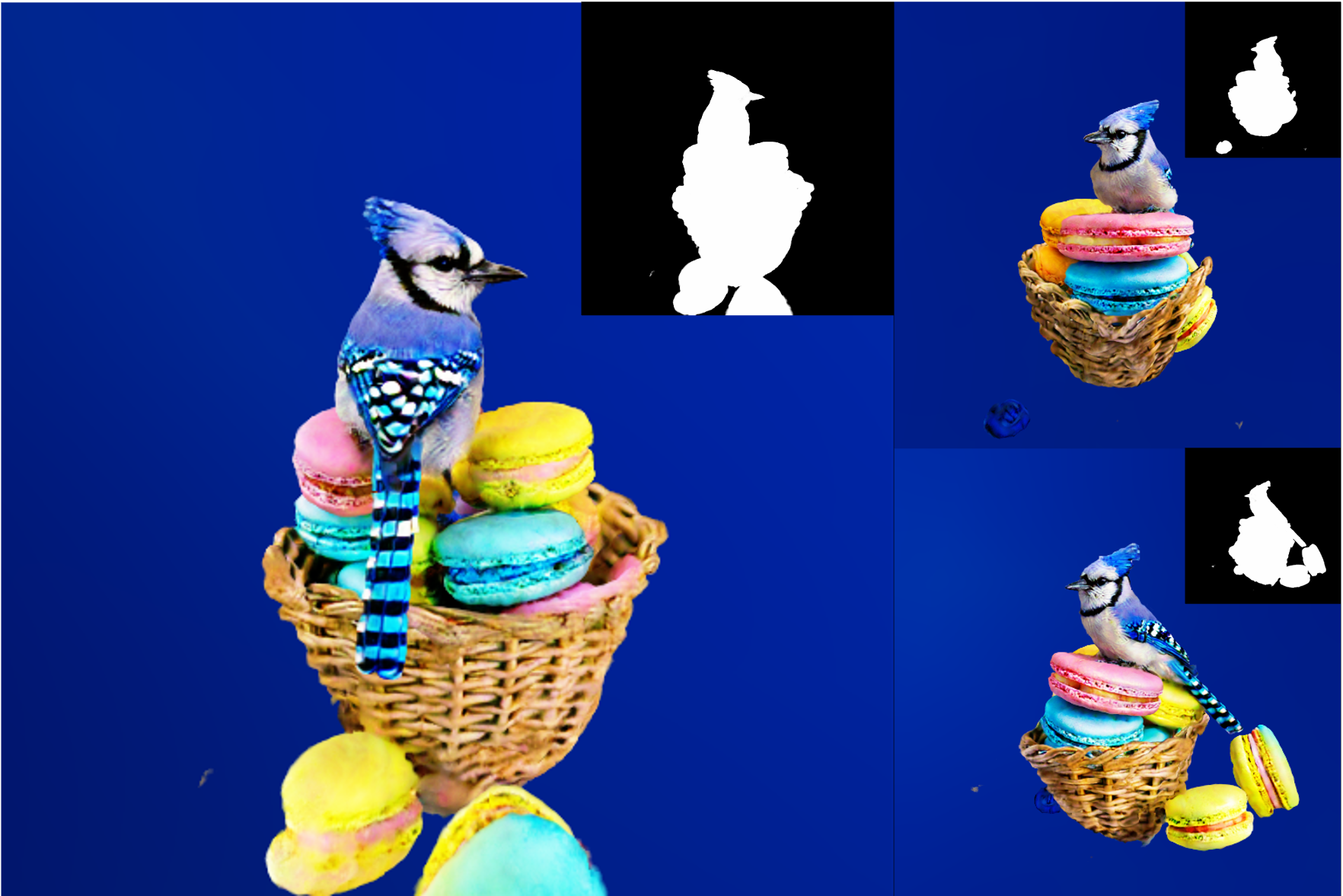}&
    \includegraphics[width=.45\linewidth]{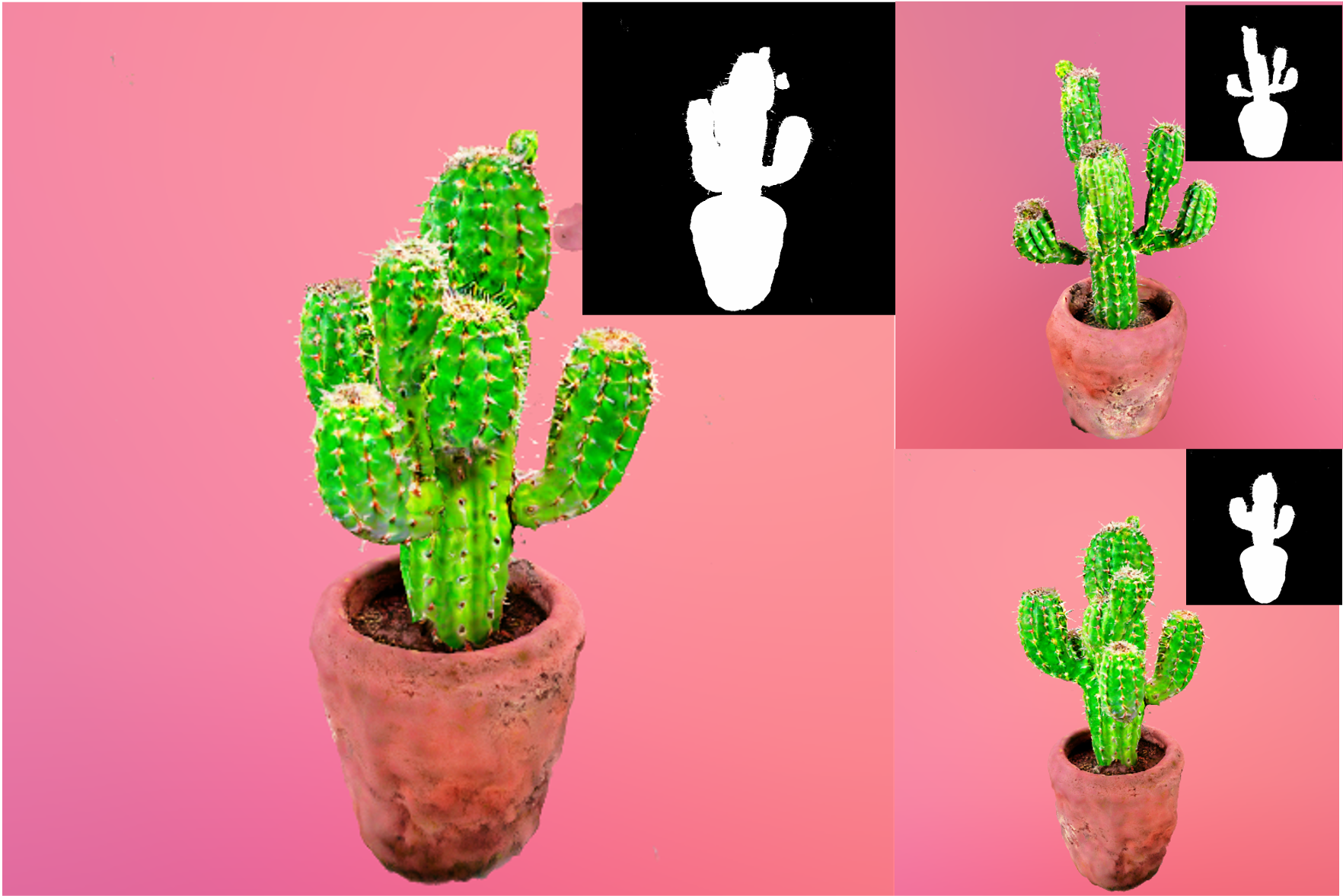}\\
    \includegraphics[width=.45\linewidth]{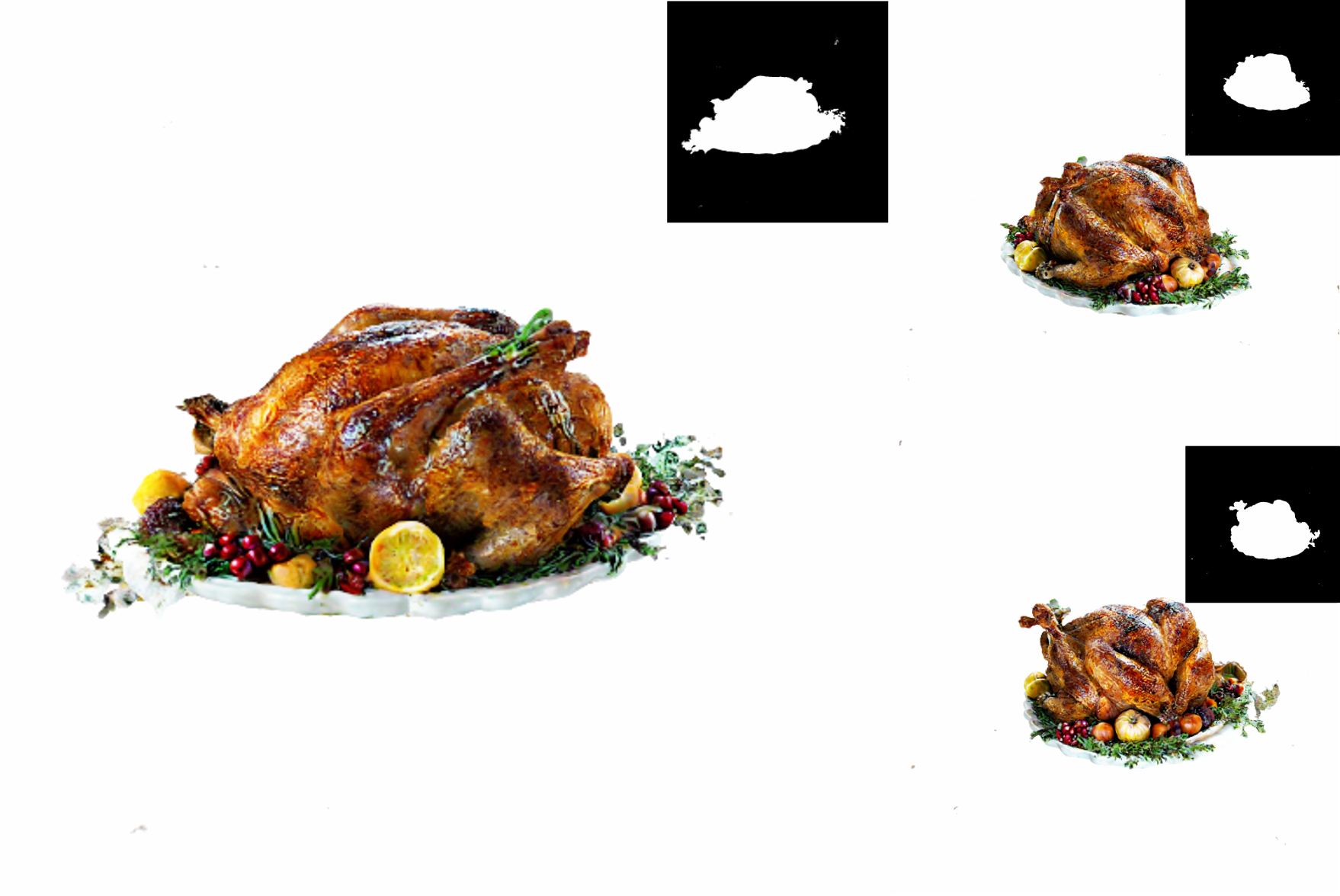}&
    \includegraphics[width=.45\linewidth]{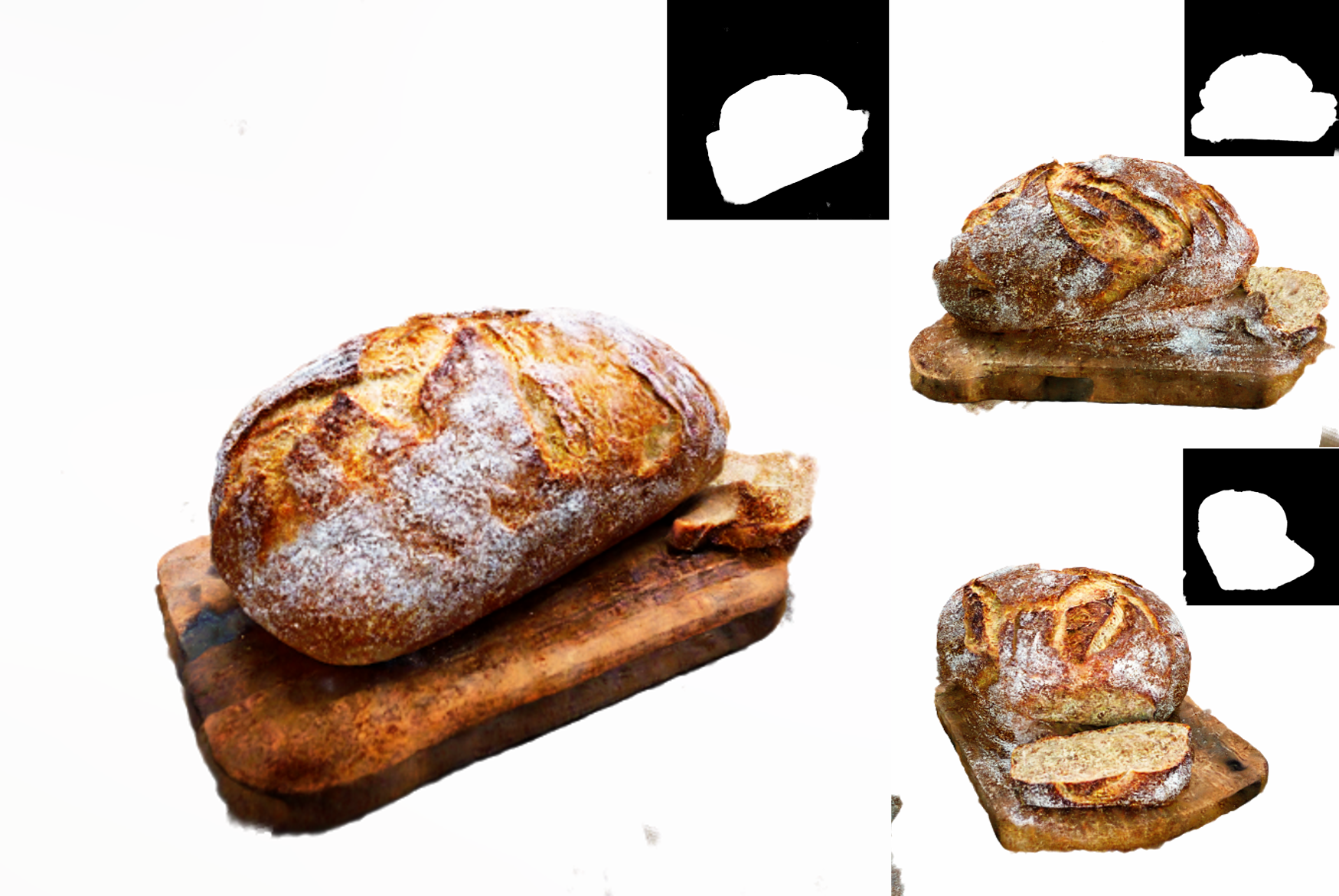}\\
    \includegraphics[width=.45\linewidth]{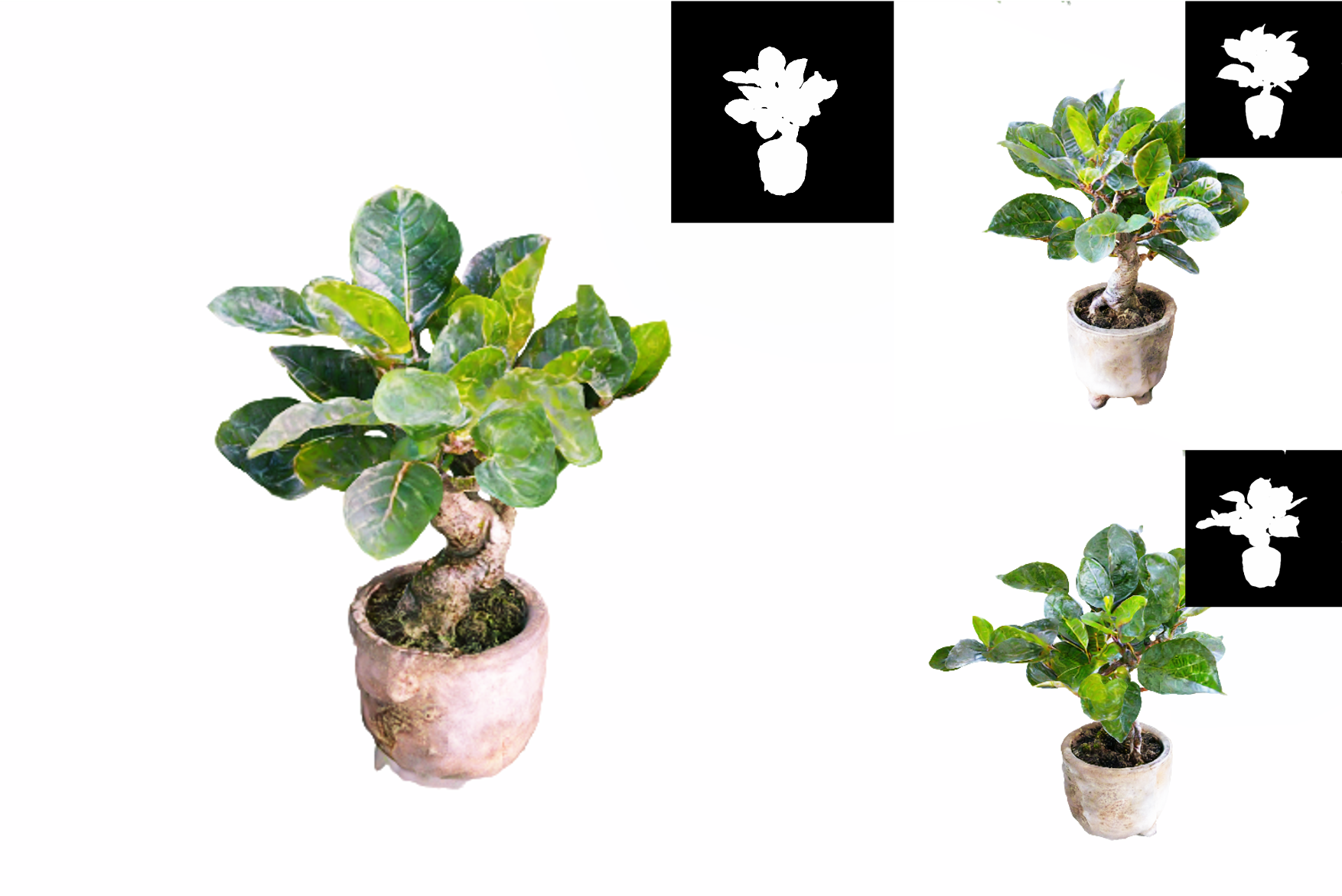}&
    \includegraphics[width=.45\linewidth]{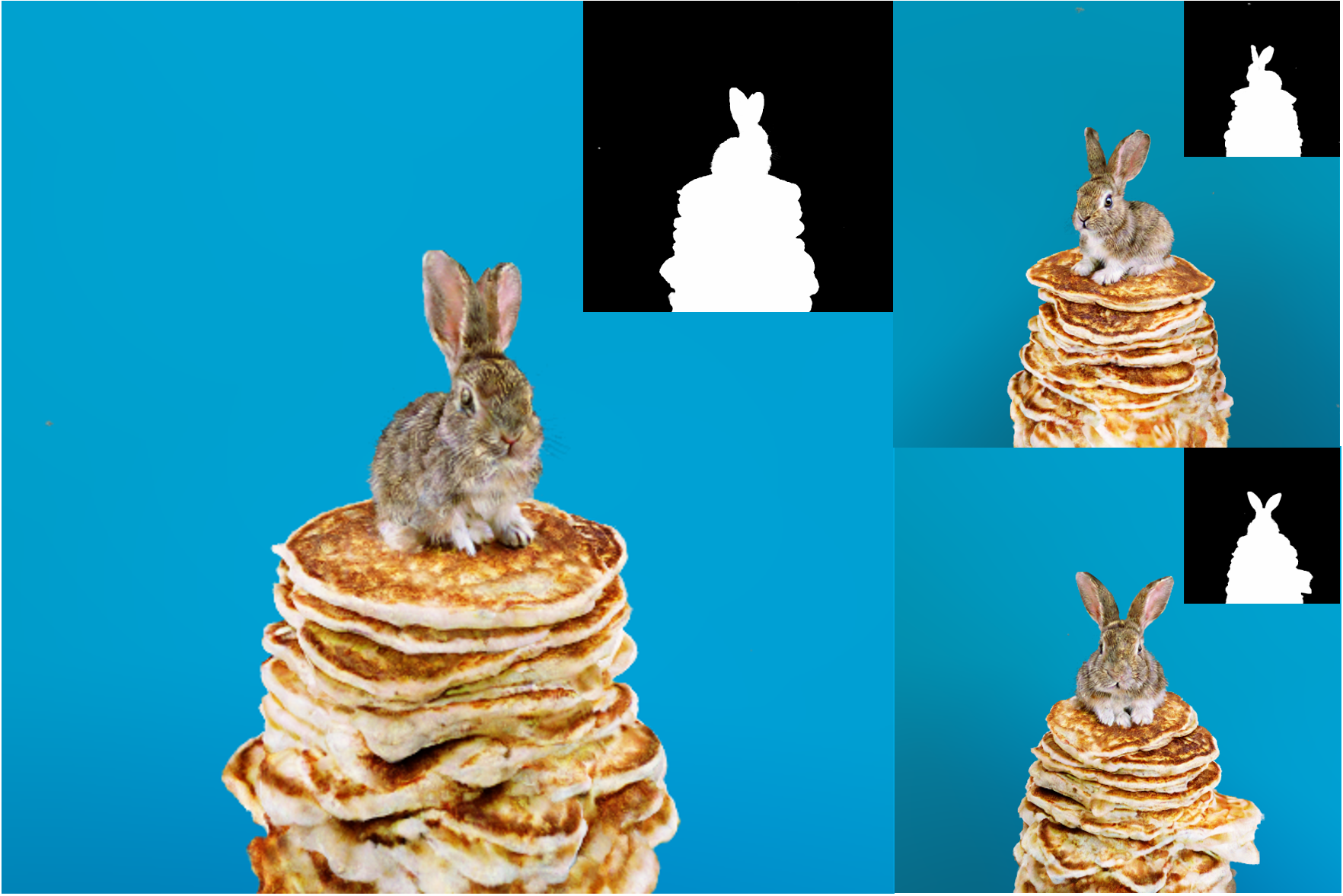}\\
    \end{tabular}
\caption{Multi-view visualizations of generated 3D results.}
\label{fig:more}
\end{figure}

\begin{figure}[p]
    \centering
    \small
    \addtolength{\tabcolsep}{-6pt}
    \begin{tabular}{cccc}
    \includegraphics[width=.249\linewidth]{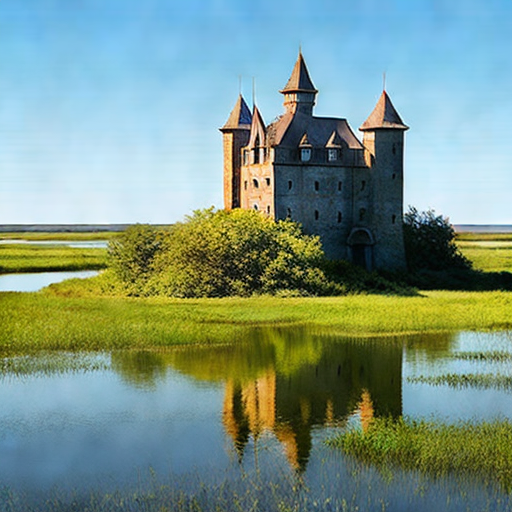}&
    \includegraphics[width=.249\linewidth]{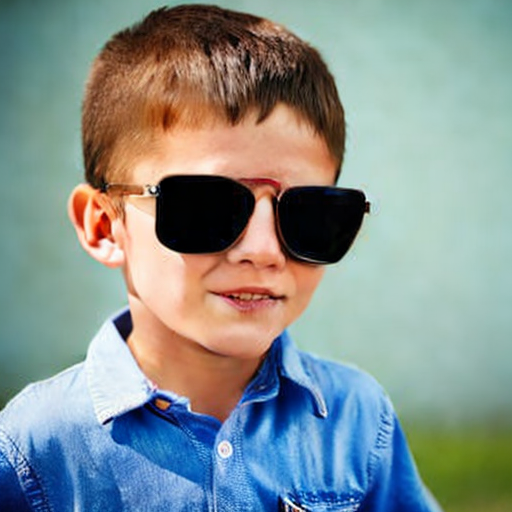}&
    \includegraphics[width=.249\linewidth]{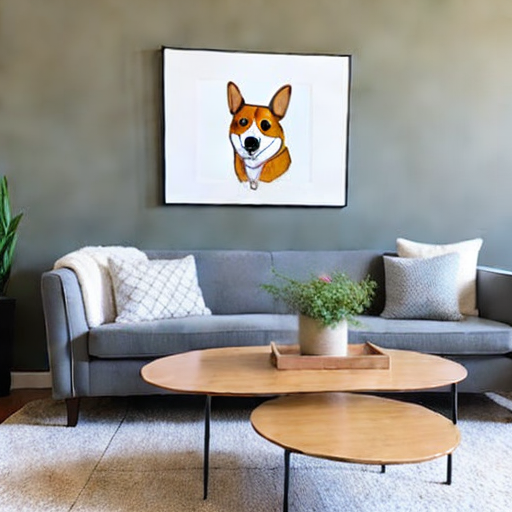}&
    \includegraphics[width=.249\linewidth]{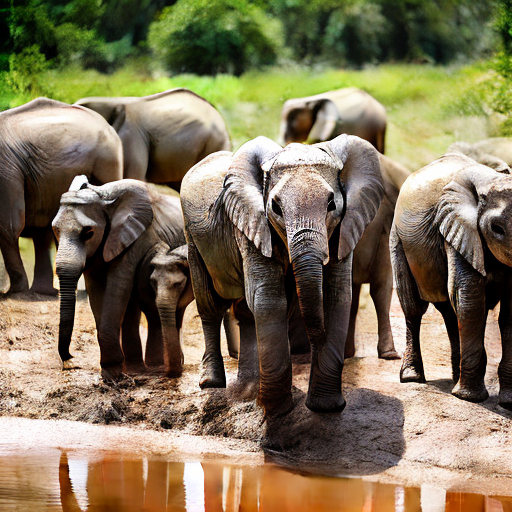}\\
    \tabincell{c}{\textit{a castle in the middle of}\\\textit{a marsh}} &
    \tabincell{c}{\textit{a boy portrait with}\\\textit{sunglasses}} &
    \tabincell{c}{\textit{a cozy living room with a}\\\textit{painting of a corgi} [...]} &
    \tabincell{c}{\textit{a group of elephants walking}\\\textit{in muddy wate}}\vspace{1mm}\\
    \includegraphics[width=.249\linewidth]{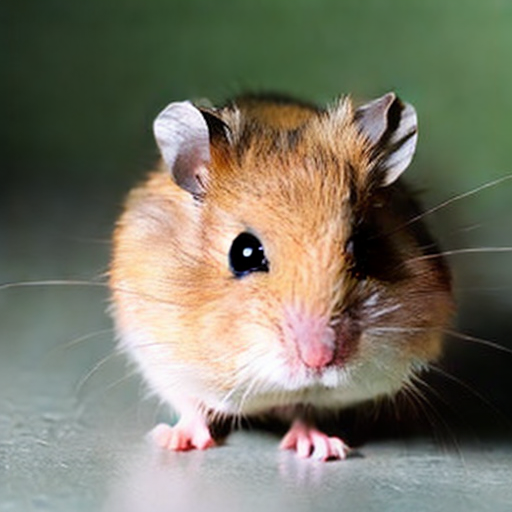}&
    \includegraphics[width=.249\linewidth]{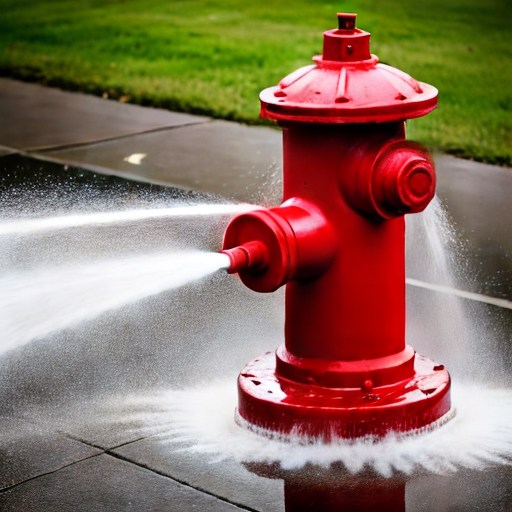}&
    \includegraphics[width=.249\linewidth]{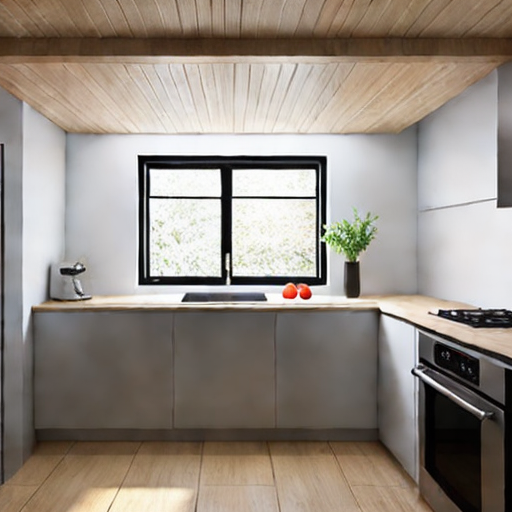}&
    \includegraphics[width=.249\linewidth]{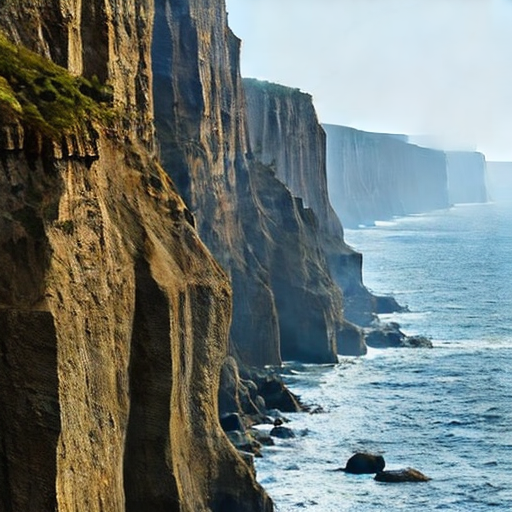}\\
    \tabincell{c}{\textit{a photograph of a hamster}} &
    \tabincell{c}{\textit{a red fire hydrant spraying}\\\textit{water}} &
    \tabincell{c}{\textit{a small kitchen with a low}\\\textit{ceiling}} &
    \tabincell{c}{\textit{cliffs at day time}}\vspace{1mm}\\
    \includegraphics[width=.249\linewidth]{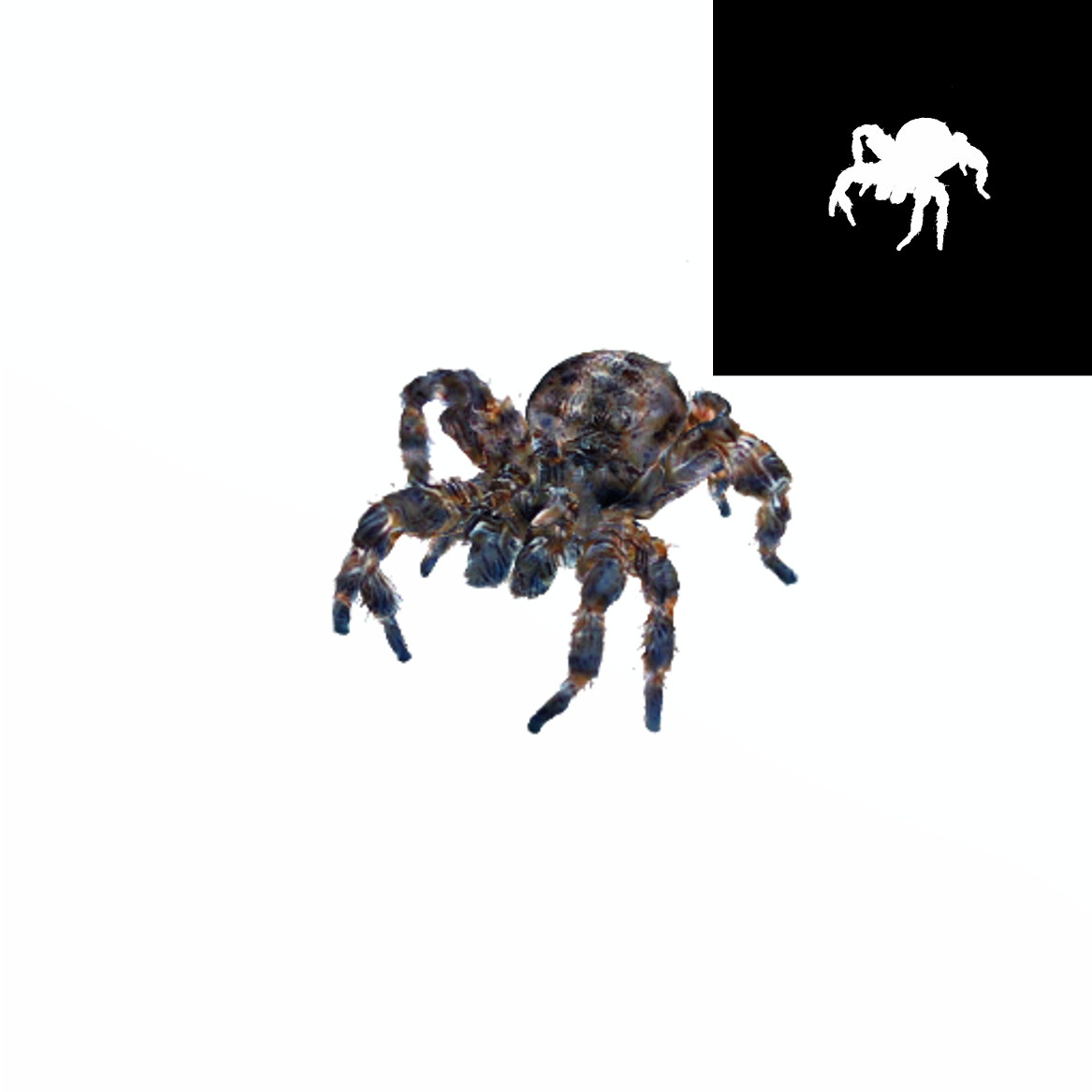}&
    \includegraphics[width=.249\linewidth]{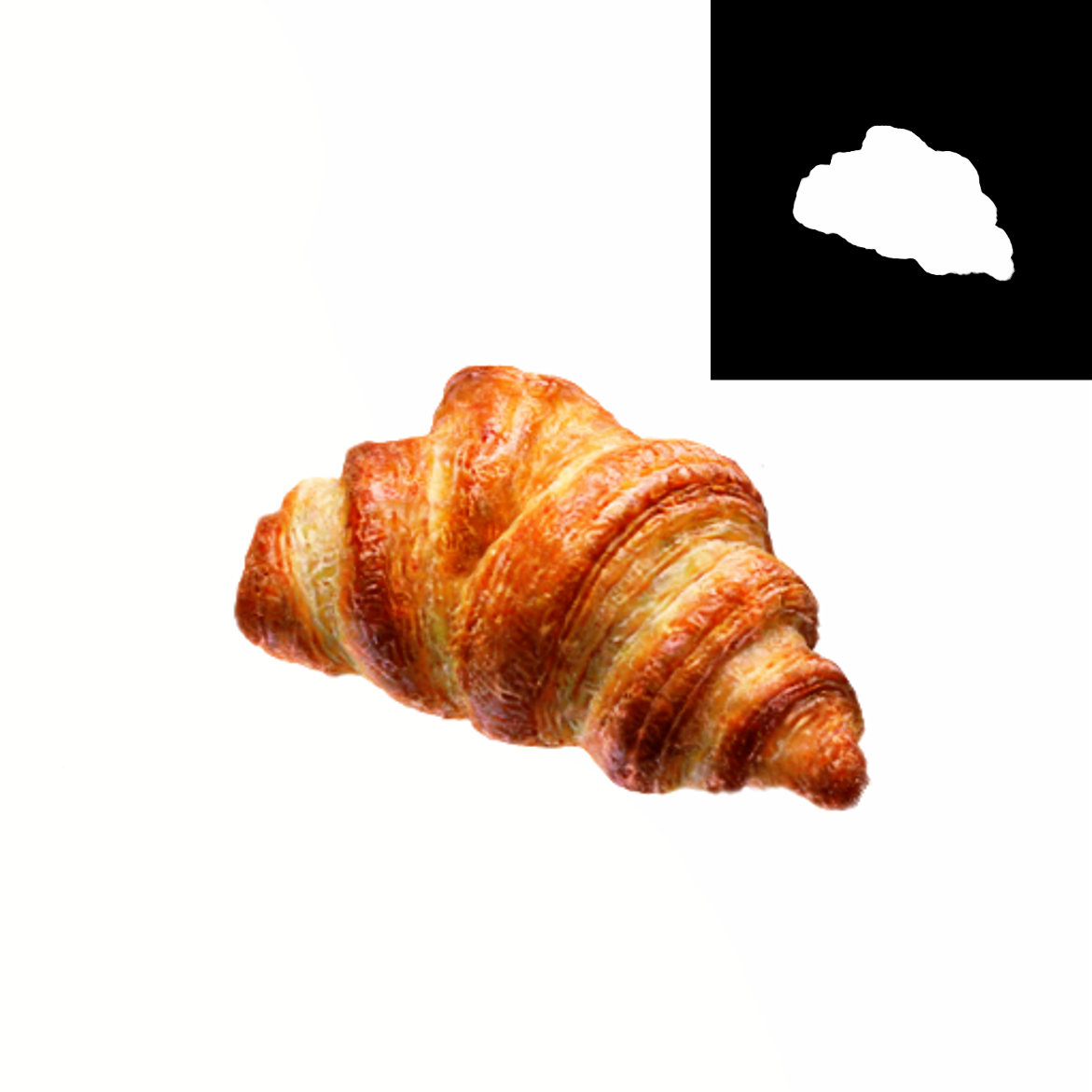}&
    \includegraphics[width=.249\linewidth]{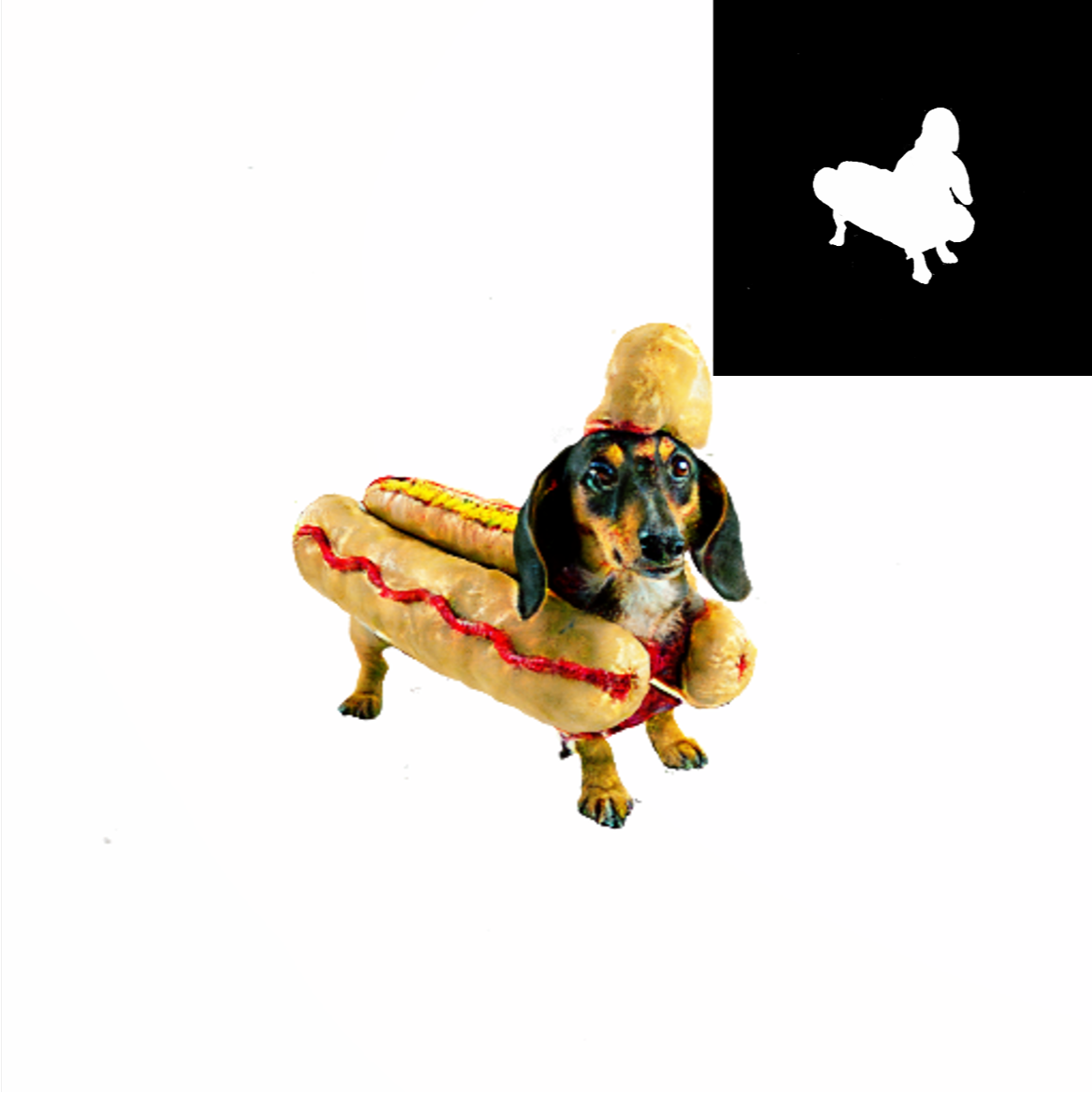}&
    \includegraphics[width=.249\linewidth]{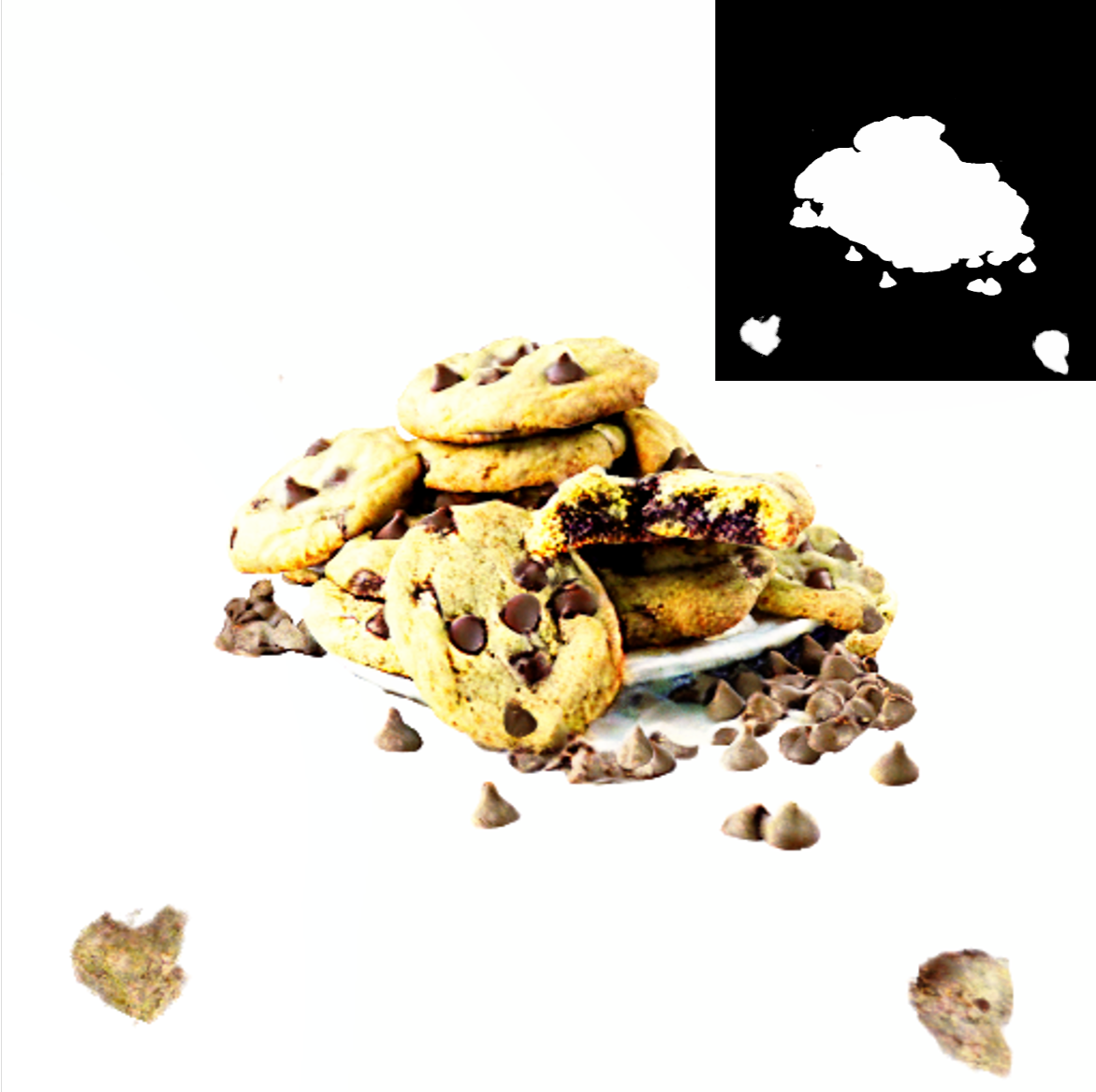}\\
    \tabincell{c}{\textit{a tarantula, highly detailed}} &
    \tabincell{c}{\textit{a delicious croissant}} &
    \tabincell{c}{\textit{a dachsund dressed up} [...]} &
    \tabincell{c}{[...] \textit{chocolate chip cookies}}\vspace{1mm}\\
    \includegraphics[width=.249\linewidth]{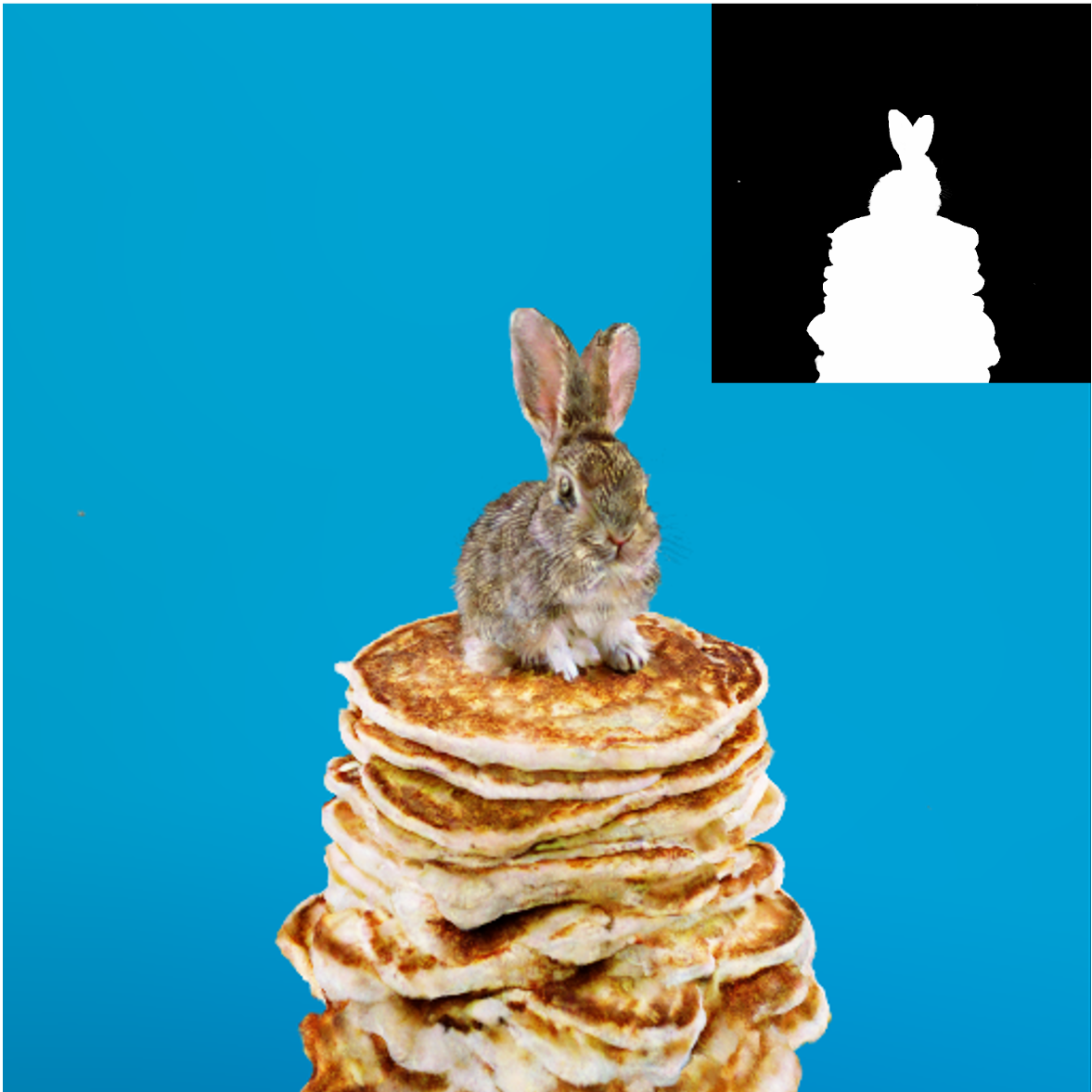}&
    \includegraphics[width=.249\linewidth]{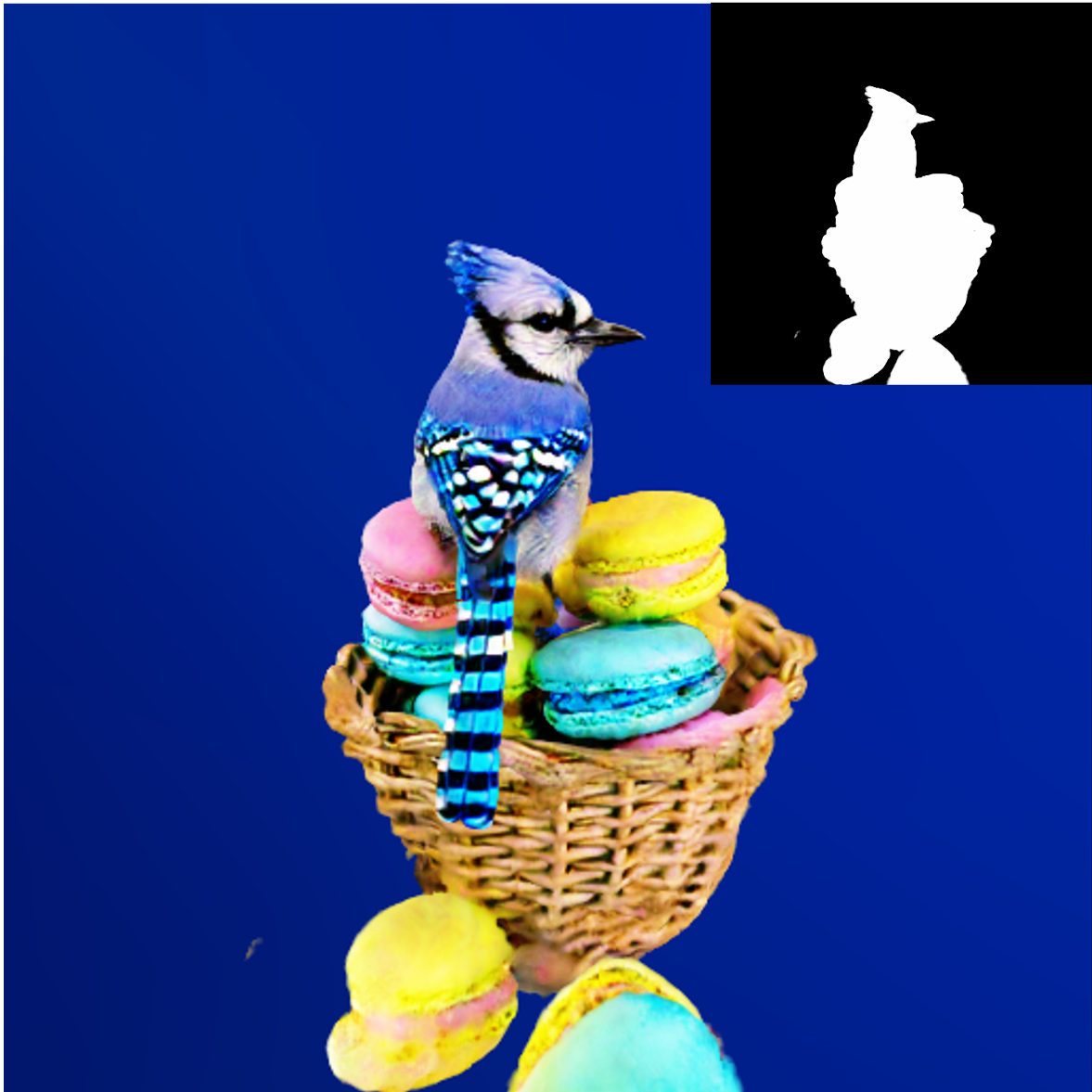}&
    \includegraphics[width=.249\linewidth]{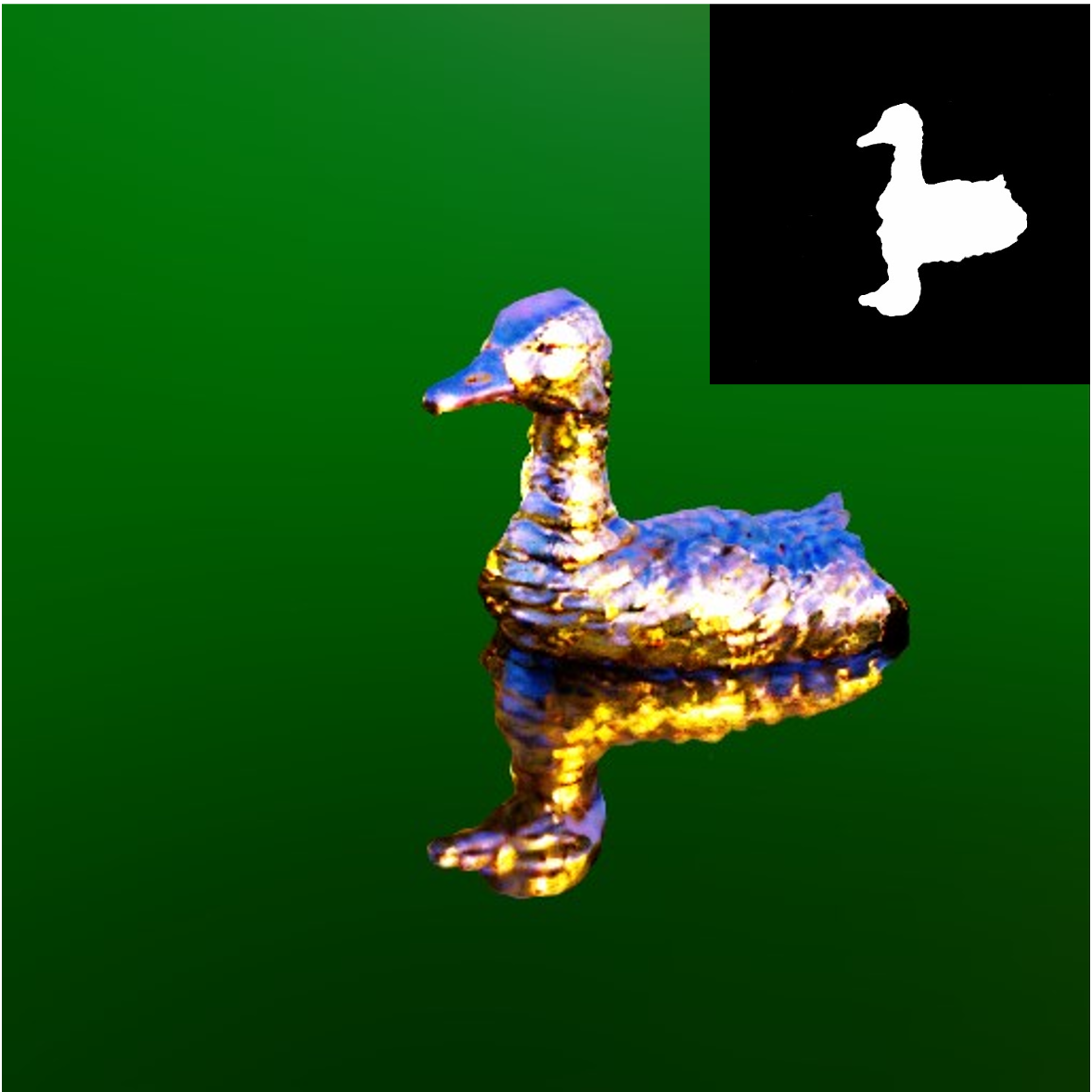}&
    \includegraphics[width=.249\linewidth]{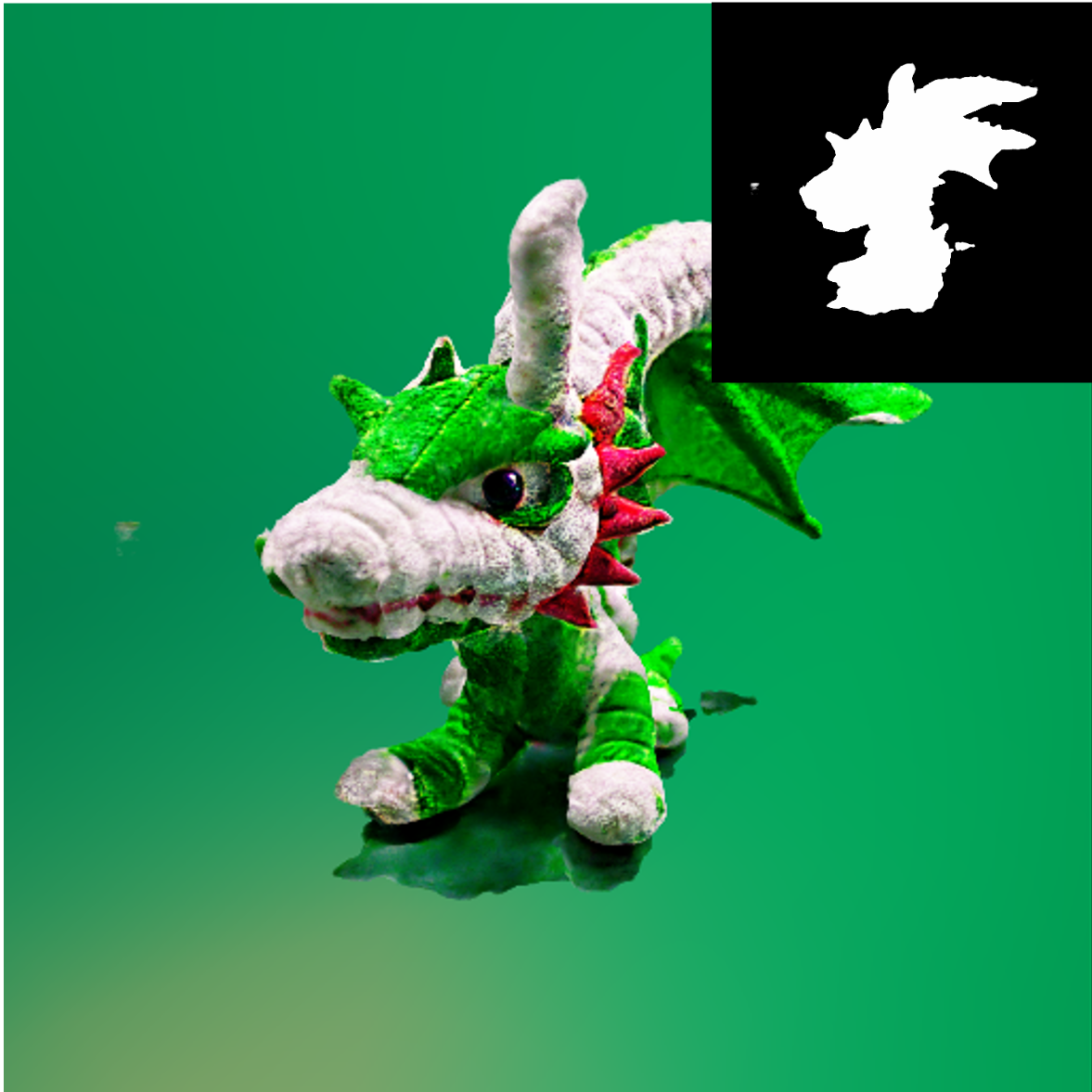}\\
    \tabincell{c}{\textit{a baby bunny sitting on} [...]} &
    \tabincell{c}{\textit{a blue jay standing on} [...]} &
    \tabincell{c}{[...] \textit{a goose made out of gold}} &
    \tabincell{c}{\textit{a plush dragon toy}}\vspace{1mm}\\
    \end{tabular}
\caption{More examples generated by ASD in both 2D and 3D.}
\label{fig:more_examples}
\end{figure}

\section{Multi-view visualization}
We exhibit some examples in Figure~\ref{fig:more}. The code and project page will be released which will contain rotating videos of the generated 3D models.

\section{Additional Experimental Results}
More examples in both 2D distillation and text-to-3D tasks are shown in Figure~\ref{fig:more_examples}. In the text-to-3D task, we only generate 3D NeRFs from scratch for simplicity, which is the first stage of VSD.

\end{document}